\documentclass{article}

\PassOptionsToPackage{numbers, compress}{natbib}
\usepackage[main, final]{NeurIPS_2026}

\usepackage[utf8]{inputenc} % allow utf-8 input
\usepackage[T1]{fontenc}    % use 8-bit T1 fonts
\usepackage{hyperref}       % hyperlinks
\usepackage{url}            % simple URL typesetting
\usepackage{booktabs}       % professional-quality tables
\usepackage{amsfonts}       % blackboard math symbols
\usepackage{nicefrac}       % compact symbols for 1/2, etc.
\usepackage{microtype}      % microtypography
\usepackage{xcolor}         % colors
\usepackage{graphicx}
\usepackage{amsmath}
\usepackage{amsmath,amsfonts,bm}

\def\eqref#1{equation~\ref{#1}}
\def\1{\bm{1}}

\DeclareMathAlphabet{\mathsfit}{\encodingdefault}{\sfdefault}{m}{sl}
\SetMathAlphabet{\mathsfit}{bold}{\encodingdefault}{\sfdefault}{bx}{n}

\usepackage{booktabs} 
\usepackage{multirow}
\usepackage{subcaption}
\usepackage{tikz}
\usepackage{tabularx}
\usepackage{fancyhdr}
\usepackage{tocloft}
\usepackage{xurl}
\usepackage{minitoc}
\usepackage{wrapfig}
\usepackage{caption} % NeurIPS usually has this, needed for \captionof

\title{Phaedra: Learning High-Fidelity Discrete Tokenization for the Physical Sciences}

\author{%
  Levi Lingsch$^{1,2,3}$\\
  \texttt{levi.lingsch@ai.ethz.ch} \\
  \And
  Georgios Kissas$^{4}$ \\
  \AND
  Johannes Jakubik$^{3}$ \\
  \And
  Siddhartha Mishra$^{1,2}$ \\
  \And 
  \\
  \begin{tabular}{cc}
  $^{1}$ETH AI Center  &  $^2$Seminar for Applied Mathematics, ETH Zurich\\
  $^3$IBM Research Europe  &  $^4$Swiss Data Science Center\\
  \end{tabular}\\
  Zurich, Switzerland
}

\begin{document}
\doparttoc             % Tells LaTeX to prepare part-level TOCs
\faketableofcontents   % Generates the .toc file silently without printing it

\maketitle

\begin{abstract}
Tokens are discrete representations that allow modern deep learning to scale by transforming high-dimensional data into sequences that can be efficiently learned, generated, and generalized to new tasks. While foundational for image and video generation, the application of tokens to physical simulation remains nascent. Because existing tokenizers are designed for the perceptual requirements of natural images, they struggle with scientific data, which exhibits large dynamic ranges and requires exact preservation of physical and spectral properties. In this work, we investigate the performance of a suite of image tokenizers across metrics designed to measure PDE fidelity. Observing that these baselines struggle to simultaneously capture fine geometric details and precise physical magnitudes, we propose \textbf{Phaedra}, a novel tokenizer inspired by classical shape-gain quantization and the paradigm of basis functions coupled with continuous coefficients. Phaedra acts as a highly effective nonlinear compression algorithm, massively reducing dataset footprints while maintaining physical fidelity. We demonstrate that Phaedra consistently improves reconstruction across diverse 2D gridded PDE solutions, generalizes robustly to unseen PDE types and real-world Earth observation data, and is competitive with continuous models in downstream proof-of-concept operator learning and masked autoencoding tasks.
\end{abstract}

\section{Introduction}

\begin{figure*}
    \centering
    \includegraphics[width=\linewidth]{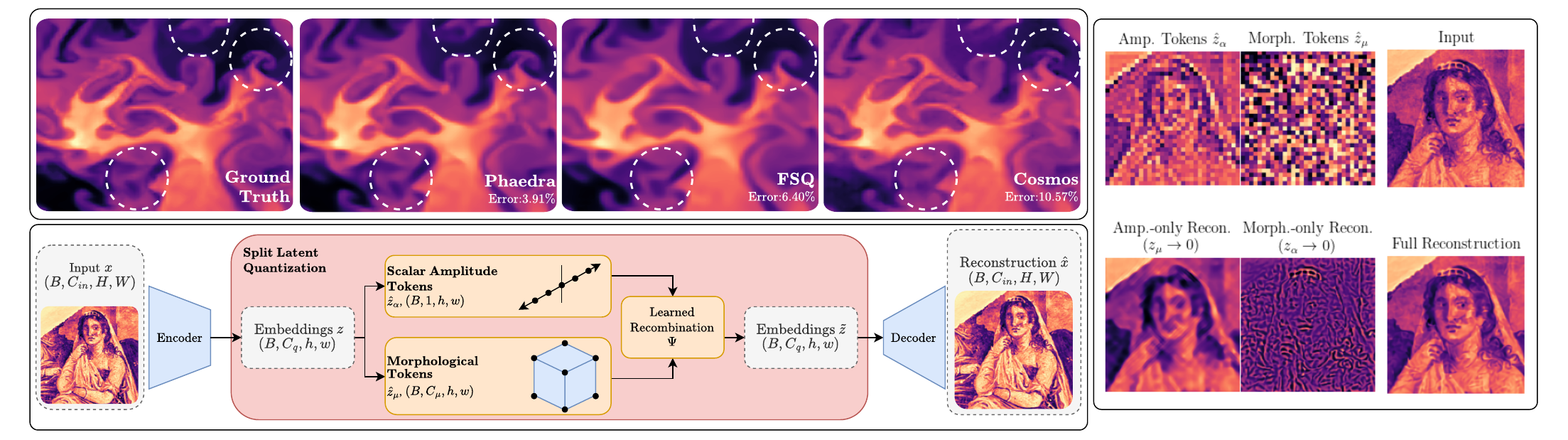}
    \caption{\textbf{Top Left:} Density field of compressible Euler equations zoomed in to show fine details, ground truth and reconstructions with relative $L_1$ errors. FSQ fails to capture high frequency information, resulting in smoothing of structures. Cosmos is designed to focus on fine details, but fails to capture precise amplitudes. Phaedra is able to model both phenomena accurately, minimizing reconstruction errors. \textbf{Bottom Left:} Phaedra tokenization pipeline. Embeddings are split and encoded in two streams: 1-dimensional \emph{amplitude tokens} (finely quantized with 1024 levels) and $C_\mu$-dimensional \emph{morphological tokens} (quantized via multi-dimensional FSQ).
    \textbf{Right:} Visualization of the disentangled representation. The morphological component, generated here as a reconstruction with amplitude tokens set to zero, encodes local structure and high frequency features. The amplitude tokens encode a globally coherent, smoothed representation of the signal. Learning global and local features separately improves the final quality of the reconstruction.}
    \label{fig:main figure}
\end{figure*}

Foundation models form the very basis of large-scale AI models such as LLMs, large vision models, multi-modal and world models \cite{FM1}. They have revolutionized the application of AI in a wide variety of fields ranging from language and image processing to robotics and genomics. The dominant structural paradigm for foundation models---especially multimodal ones---consists of two ingredients. First, a \emph{tokenizer} transforms different input modalities (text, images, video etc.) into a sequence of \emph{discrete, quantized} tokens, providing both significant levels of information compression as well as a unified space of latent representations across modalities. Second, a \emph{processor} (typically a large transformer) either predicts the next token autoregressively or unmasks sets of masked tokens. The output is then \emph{detokenized} via the tokenizer to the desired modality. This fundamental paradigm has enabled the extreme scalability and outstanding representation learning capacity of foundation models, while equipping them for accurate fine-tuning on very diverse downstream tasks.  

Foundation models for the physical sciences and engineering are rapidly evolving \cite{aurora, herde2024poseidonefficientfoundationmodels, dpot, mccabe2025walruscrossdomainfoundationmodel, serrano2025zebraincontextgenerativepretraining, koupai2025enma}. These models are critical; pretraining a large model to learn fundamental physical relationships equips it to quickly adapt to new systems of equations, drastically decreasing data requirements \cite{herde2024poseidonefficientfoundationmodels}. However, in contrast to the discrete paradigm of LLMs, many current physics models (such as neural operators \cite{NO,Reno,FNO,raonic2023convolutionalneuraloperatorsrobust,GAOT,RIGNO,UPT, kissas2022learning}) employ a \emph{continuous latent space}. Consequently, these models are trained to regress to the labeled output data by regressive losses (MSE/MAE). The current continuous paradigm has been shown to suffer from smoothing during long rollouts, but this can be alleviated by generative modeling \cite{gencfd}. Discrete tokenization allows for the use of cross-entropy losses, enabling models to represent complex uncertainty distributions within a generative framework. More importantly, PDE foundation models such as \cite{herde2024poseidonefficientfoundationmodels,dpot,mccabe2025walruscrossdomainfoundationmodel}, are limited in their capacity to model multi-modal data. Domains in different dimensions (2D vs. 3D), domain input points that vary between Cartesian and unstructured grids, or symbolic information about PDEs as an input are not straightforward to combine, on account of the incompatibility of the continuous latent representations over individual modalities. Although tokenizers for 3D, irregular meshes, symbolic forms are not studied explicitly in this work, the ability of a transformer to process multimodal data in the same format serves as a primary motivation for this research. By mapping these heterogeneous inputs to a unified sequence of discrete tokens, a transformer can be leveraged as a universal processor for multi- and cross-modal learning. This shift to discrete tokens not only provides significant data compression, but also increases the \emph{information density} of the latent representations, focusing the transformer-processor's capacity on learning the underlying physics.

\begin{wrapfigure}{r}{0.5\textwidth}
    \includegraphics[width=\linewidth]{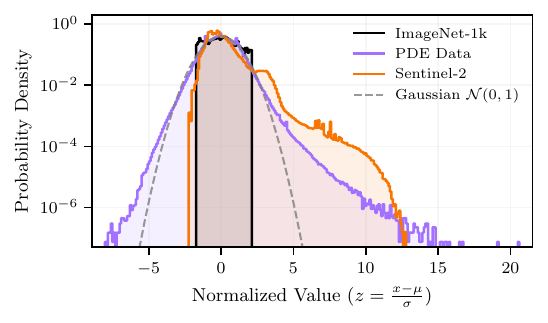}
    \captionof{figure}{Even after normalization, natural images have a much narrower range than scientific data.}
    \label{fig:data distribution imagenet vs pde}
\end{wrapfigure}

Given the advantages of tokenization, namely its alignment with scalable transformer backbones and robust generative objectives and the fact that discrete token are the de facto representation in modern AI, these discrete representations are increasingly being utilized in emerging SciML models (e.g., PhysiX \cite{nguyen2025physixfoundationmodelphysics}, Zebra \cite{serrano2025zebraincontextgenerativepretraining}, Terramind \cite{jakubik2025terramindlargescalegenerativemultimodality}, and QST \cite{basile2026quantizingspacetimefusing}). However, this shift introduces a new challenge: finding a tokenizer suitable for physical data. An obvious starting point is to repurpose off-the-shelf image tokenizers, such as Nvidia Cosmos \cite{nvidia2025cosmosworldfoundationmodel}, to tokenize continuous spatio-temporally varying fields. While these tokenizers can be adapted by interpreting physical fields as images, their performance is observed to be poor (see Fig.~\ref{fig:main figure} and Tab.~\ref{tab: cosmos comparison}). Even after retraining a variety of image tokenizers directly on scientific datasets \cite{nguyen2025physixfoundationmodelphysics, jakubik2025terramindlargescalegenerativemultimodality}, the reconstruction errors remain high. 

Our main aim in this paper is to answer the critical open question of \emph{how} to construct these discrete tokens to best connect with the precise mathematics of PDEs. To motivate our design, we argue that the shortcomings of image tokenizers such as VQ-VAE \cite{van2017neural} or FSQ \cite{mentzer2023finite} in this context can be attributed to three factors: \\
 i) \textbf{Perceptual vs.\ Physical Fidelity:} Image tokenizers are optimized for perceptual similarity (e.g., LPIPS), hallucinating high-frequency information that mimics natural textures. Optimizing for visual fidelity often lowers the pixel-wise accuracy of the reconstruction, leading to the \emph{perception-distortion tradeoff} \cite{Blau_2018}. In physics, such errors in small-scale gradients can violate conservation laws and symmetries or lead to divergent spectral behavior during time-stepping. \\
 ii) \textbf{Unbounded Dynamic Range:} Natural images are strictly bounded (e.g., pixel values in $[0,255]$). Physical fields, on the contrary, exhibit heavy-tailed distributions with a large dynamic range (Figure~\ref{fig:data distribution imagenet vs pde}). A standard fixed codebook struggles to capture rare, high-energy events without sacrificing resolution for the bulk of the low-energy flow. \\
 iii) \textbf{The Amplitude-Morphology Conflict:} A discrete codebook forces a trade-off. To capture subtle variations in magnitude, the codebook must be exponentially large. However, to capture diverse geometric shapes, it must be semantically rich. Attempting to do both with a single integer code creates a bottleneck where the model generates the correct ``shape'' of a feature but fails to reconstruct its exact intensity, or vice versa.  

To address these shortcomings, we propose \emph{Phaedra}, a novel tokenizer designed to represent spatially varying fields using two complementary discrete representations: morphology (the local pattern) and amplitude (an orthogonal element capturing absolute magnitude), conceptually echoing the classical mathematical paradigm of combining basis functions with continuous coefficients. As depicted in Figure~\ref{fig:main figure}, the morphology is discretized with vector quantization to form a codebook of reusable local patterns, while the amplitude stream is discretized with scalar quantization to preserve magnitude and dynamic range. Physical fields are reconstructed by recombining these two factorized tokens.

In summary, the main contributions of this work are:
\begin{itemize}
    \item We introduce \textbf{Phaedra}, a discrete tokenizer that resolves the amplitude-morphology conflict by learning patterns and physical scales independently, presented in Section \ref{sec: disentanglement}.
    \item We demonstrate that Phaedra acts as a highly effective nonlinear compression algorithm, reducing scientific data footprints while preserving physical fidelity.
    \item We show that Phaedra outperforms state-of-the-art image tokenizers (including Cosmos \cite{nvidia2025cosmosworldfoundationmodel} and VAR \cite{tian2024visual}) on complex physics datasets, exhibiting robust zero-shot performance on unseen PDE families, real-world Earth observation, and weather data.
    \item We provide initial downstream evaluations as a proof of principle study that Phaedra's superior tokenization directly translates to enhanced performance in operator learning and masked autoencoding tasks compared to continuous baselines and standard tokenizers.
\end{itemize}

\section{Methods}
\label{sec: methods}
\subsection{Preliminaries: Discrete Tokenization}

\begin{wraptable}{r}{0.5\textwidth}
    \centering
    \caption{Structural comparison of tokenization schemes.}
    \label{tab:tokenizers}
    \footnotesize % Shrinks the text slightly
    \resizebox{\linewidth}{!}{ % Scales the table to fit exactly inside the wrap box
    \setlength{\tabcolsep}{3pt} % Reduces padding between columns
    \begin{tabular}{lccccc}
    \toprule
    Method & $K$ & $\Phi$ & $Q$ & $\Psi$ & $\#$(Tokens) \\
    \midrule
    VQ-VAE     & 1 & Id.      & VQ  & Id.     & $hw$ \\
    FSQ        & 1 & Id.      & FSQ & Id.      & $hw$ \\
    VQ-VAE-2   & 2 & Scale Split    & VQ $\times$2  &  Con.+Ups.   & $h_1w_1+h_2w_2$ \\
    VAR        & $S$ & Residual & FSQ    & Add  & $\sum_s h_sw_s$ \\
    Phaedra    & 2 & Channel Split  & FSQ$\times$2 & Learn & $2hw$ \\
    \bottomrule
    \end{tabular}}
\end{wraptable}

To better explain Phaedra and where it stands in the literature, we first mathematically formalize discrete tokenization systems. A \emph{discrete tokenization system} transforms continuous inputs $x \in \mathbb{R}^{C \times H \times W}$ into discrete codes and reconstructs the input from those codes. We unify such systems under the following framework. Let $z \in \mathbb{R}^{C_q \times h \times w}$ be a continuous latent, $z^k \in \mathbb{R}^{d_k \times h_k \times w_k}$ a factorized branch latent, $\hat{z}^k \in \mathbb{R}^{d_k \times h_k \times w_k}$ its quantized counterpart (a continuous vector restricted to a finite codebook), and $c^k \in \left(\prod_{j=1}^{m_k} [V_k^j]\right)^{h_k \times w_k}$ the corresponding discrete integer indices, where $[V] = \{0, 1, \ldots, V-1\}$. The full tokenization pipeline is:
\begin{equation}
\label{eq:dis1}
    x \xrightarrow{\mathcal{E}} z \xrightarrow{\Phi} \{ z^{k} \}_{k=1}^K \xrightarrow{\{ Q_k\}} \{ (\hat{z}^{k}, c^k ) \}_{k=1}^K \xrightarrow{\Psi} \tilde{z} \xrightarrow{\mathcal{D}} \hat{x}.
\end{equation}
The maps are defined as follows. The \textbf{encoder} $\mathcal{E}: \mathbb{R}^{C \times H \times W} \to \mathbb{R}^{C_q \times h \times w}$ maps input data to a continuous latent $z$. The \textbf{factorization operator} $\Phi: \mathbb{R}^{C_q \times h \times w} \to \prod_{k=1}^K \mathbb{R}^{d_k \times h_k \times w_k}$ decomposes $z$ into $K$ branch latents $\{z^k\}_{k=1}^K$ (e.g., by channel splitting or residual refinement). Each \textbf{quantization operator} $Q_k: \mathbb{R}^{d_k \times h_k \times w_k} \to \mathbb{R}^{d_k \times h_k \times w_k} \times \prod_{j=1}^{m_k} [V^j_k]^{h_k \times w_k}$ maps $z^k$ to a pair $(\hat{z}^k, c^k)$, where $\hat{z}^k$ is the quantized continuous latent (the codebook vector) and $c^k$ is the corresponding index (the token). The \textbf{recombination operator} $\Psi: \prod_{k=1}^K \mathbb{R}^{d_k \times h_k \times w_k} \to \mathbb{R}^{C'_q \times h' \times w'}$ merges the quantized latents $\{\hat{z}^k\}_{k=1}^K$ into a single latent $\tilde{z}$. Finally, the \textbf{decoder} $\mathcal{D}: \mathbb{R}^{C'_q \times h' \times w'} \to \mathbb{R}^{C \times H \times W}$ reconstructs $\hat{x}$ from $\tilde{z}$. This framework accommodates different tokenization schemes via choices of $\Phi$, $Q$, and $\Psi$; see Table~\ref{tab:tokenizers} for a taxonomy.

\textbf{Finite Scalar Quantization (FSQ).} To mitigate the bottlenecks of a learned codebook, \citet{mentzer2023finite} proposed FSQ, which simplifies quantization by projecting the continuous latent $z$ onto a fixed, predefined rectangular lattice. To properly accommodate both even and odd numbers of levels $L\in \mathbb{N}^d$, the canonical FSQ implementation uses an asymmetric shift. For each dimension, let $w=\lfloor L/2 \rfloor$, $o=0.5$ if $L$ is even and $o=0$ if odd, and $l_{1/2}=\frac{L-1}{2}(1-\epsilon)$, where $0<\epsilon<<1$. We define a pre-activation shift $s=\tanh^{-1}(o/l_{1/2})$. The continuous bounded representation is $b(z) = l_{1/2} \tanh(z+s) - o$.
The continuous values are rounded to discrete intermediate states $q= \text{round}(b(z))$. The final non-negative integer indices $c$ (for codebook lookups) and the normalized quantized latents $\hat{z}$ (passed to the decoder) are obtained via $c = q + w,$ $\hat{z} = {q}/{w}$.

\subsection{The Phaedra Architecture}
We propose \textbf{Phaedra}, a tokenizer designed to address the unique statistical properties of physical fields via a \textbf{Dual-Latent Factorization} operator $\Phi$ and a \textbf{Learned Recombination} operator $\Psi$, motivated by signal analysis and reduced-order modeling.

\paragraph{Theoretical Motivation: Neural Shape-Gain Quantization.} The factorization in Phaedra is motivated by the observation that physical fields exhibit independent variations in \emph{structure} (morphology) and \emph{energy} (amplitude). For example, the shape of a vortex or shock is often independent of the magnitude of the underlying velocity field. We draw inspiration from Shape-Gain Quantization (SGQ) \citep{gersho1992vector}, which approximates a vector as the product of a normalized shape vector and a scalar gain. Analogously, $z_\mu$ encodes geometric ``basis functions'' of the field, while $z_\alpha$ is a scalar coefficient projecting those bases onto the correct dynamic range. We note here, no explicit basis functions are chosen in this model -- we use this term merely as an analogy to shape-gain quantization.

This offers two key advantages: it enables precise amplitude reconstruction across the domain, and focuses the representational capacity of $z_\mu$ on local features, mitigating the smoothing artifacts common to other quantization approaches. Although implemented as a generic nonlinear map, the role of $\Psi$ is to modulate structural and amplitude information in a learned—rather than enforced—manner, allowing the decoder to discover interactions that best fit the data. The factorized latent provides structured coordinates on a nonlinear solution manifold, while $\mathcal{D} \circ \Psi$ acts as a nonlinear manifold decoder, akin to recent ROM and operator learning approaches \citep{seidman2022nomad, lingsch2024fuse}.

\paragraph{Encoder and Factorization.} Given a physical field $x\in \mathbb{R}^{C\times H \times W}$, the encoder $\mathcal{E}$ produces a latent tensor $z$. The factorization operator $\Phi$ then performs a channel-wise split to segregate semantic roles:
\begin{equation}
    z = \mathcal{E}(x), \qquad \Phi(z) = [z_\mu,\, z_\alpha] \quad \text{for} \quad z\in \mathbb{R}^{C_q\times h \times w},
\end{equation}
where $z_\mu \in \mathbb{R}^{C_\mu \times h \times w}$ captures local structural invariants (morphology), $z_\alpha \in \mathbb{R}^{1 \times h \times w}$ captures local energy density (amplitude), and $C_q = C_\mu + 1$. Importantly, these roles are not constrained by a loss. Instead, they are disentangled naturally as a result of how the channels are quantized. 

\paragraph{Split-FSQ Quantization.} We apply distinct FSQ configurations to each branch. For \textbf{morphological quantization}, we define levels $L_\mu \in \mathbb{N}^{C_\mu}$ to produce a semantically rich implicit codebook $\mathcal{C}_\mu$, with quantized latent $(\hat{z}_\mu, c_\mu) = Q_{\mathrm{FSQ}}(z_\mu;\, L_\mu)$. For \textbf{amplitude quantization}, high numerical precision is essential, as amplitudes are ``strongly continuous''—there are no learnable discrete patterns to exploit. We therefore treat $z_\alpha$ as a 1-D manifold quantized at high density with $L_\alpha = [1024]$. 

To ensure stability and avoid a domain mismatch between pre- and post-quantization values, we wrap the core FSQ operation in a domain-preserving scaling module. For a given branch latent $z$, the full quantization operator $Q_{FSQ}$ scales the input by a hyperparameter $\tilde{s}$ to align the latent distribution predominantly with the approximately linear region of the tanh activation, applies the bounded quantization, and then divides by $\tilde{s}$. This scaling prevents the bulk of the latents from saturating the tanh function, which would artificially cap the denormalized outputs to $\pm1/\tilde{s}$ and inflate the commitment loss, yet still allows the tanh to softly bound extreme outliers.

\paragraph{Latent Recombination and Decoding.} The quantized morphology and amplitude latents are concatenated and passed through a learnable channel-mixing convolution,
\begin{equation}
    \tilde{z} = \Psi\!\left(\operatorname{concat}[\hat{z}_\mu,\, \hat{z}_\alpha]\right), \qquad \hat{x} = \mathcal{D}(\tilde{z}).
\end{equation}

\paragraph{Loss Function.} We train with a combination of an $L_1$ reconstruction loss and commitment losses \citep{van2017neural},
\begin{equation}
    \mathcal{L}_{\mathrm{Phaedra}} = |x-\hat{x}|_1 + \beta\|z_\mu - \operatorname{sg}[\hat{z}_\mu]\|_2^2 + \|z_\alpha - \operatorname{sg}[\hat{z}_\alpha]\|_2^2,
\end{equation}
where $\operatorname{sg}[\cdot]$ is the stop-gradient operator. We use $\beta=0.25$ for the morphological commitment term; no scaling factor is applied to the amplitude term, as its loss is already small due to the density of the token space. While the commitment losses are not strictly necessary for FSQ, they can help to prevent the encoder from producing embeddings far outside the nominal range of the codebook, potentially mitigating any instability during training.

\section{Experiments \& Results}
We evaluate Phaedra against a suite of SOTA tokenizers on three axes: (i) reconstruction fidelity on in-domain physics; (ii) high-compression efficiency versus large-scale foundation tokenizers; and (iii) generalization across unseen PDE realizations, unseen PDE types, and non-PDE scientific datasets. We evaluate using normalized MAE (nMAE), normalized RMSE (nRMSE), minimal spectral coherence $\gamma_{\min}\in[0,1]$, and local variance error $\Delta \sigma^2_{\mathrm{loc}}$. The metrics $\gamma_{\min}\in[0,1]$ and $\Delta \sigma^2_{\mathrm{loc}}$ are used specifically to quantify the preservation of essential structures such as shocks and vortices. Calculation of these metrics and additional metrics for outlier errors, spectra, and codebook utilization are presented in full detail in \ref{sec: results evaluation in physical space}, \ref{sec:results evaluation in spectral space}, and \ref{sec: results evaluation of token usage}.

\paragraph{Amplitude--Morphology Disentanglement.}
Before reporting reconstruction metrics, we provide quantitative evidence that the dual-latent factorization achieves a meaningful separation of amplitude and structure (motivated by SGQ in Section~\ref{sec: methods}). Because the nonlinear decoder prevents perfect linear separation, this is a \emph{soft} disentanglement. We validate this through three independent experiments: token swapping, spectral ablation, and filter baseline comparisons (full details in Appendix~\ref{sec: disentanglement}). 

In the \textbf{token swap} experiment, we independently swap the amplitude or morphological tokens between two uncorrelated fields, $A$ and $B$ (the Phaedra image and density field from Fig.~\ref{fig:main figure}, respectively). Swapping only $B$'s amplitude tokens into $A$ drops the energy correlation with $A$ to $-0.09$ while correlation to $B$ reaches $0.99$. Conversely, swapping only morphological tokens leaves the energy anchored to $A$ ($r=0.97$). This confirms that macroscopic energy follows the 1D amplitude tokens precisely.

In the \textbf{filtering} experiment, we compare the amplitude-only and morphology-only reconstructions with FFT based low-pass and high-pass filtering, where the filtering frequency is set equal to the bottleneck resolution of the tokenizer, 32. Using only amplitude tokens, reconstructions correlate strongly ($r=0.93$) with the FFT low-pass filtered field, with near-zero correlation to the high-pass filtered field. Likewise, the morphology-only reconstruction correlates moderately ($r=0.50$) with the high-pass filtered field and has no correlation with the low-pass field. These correlations support the claim that large-scale structure, represented by low-frequency modes of the input, are captured almost entirely by $z_\alpha$, while $z_\mu$ represents local, high-frequency structure. 

\subsection{Core Benchmarks: In-Distribution PDEs}
We assess performance on a pretraining fluid dynamics corpus (Compressible Euler \& Incompressible Navier-Stokes; \citep{herde2024poseidonefficientfoundationmodels}), at $(128^2)$ input resolution with $32^2$ latent resolution. Baselines include FSQ, VQ-VAE-2, IBQ, and VAR (two scales), described in greater detail in Section \ref{sec: Model Architectures}. Figure~\ref{fig:taxonomy} situates each baseline within the common framework of
Eq.~\ref{eq:dis1}, making explicit that all models share an encoder and decoder and
differ only in $\Phi$, $Q$, and $\Psi$.. In this setting, FSQ serves both as a standalone baseline, and as an ablation of Phaedra without the amplitude channel. Full training and dataset details are in \ref{sec: Experimental Setup}, including model and codebook sizes (Table \ref{tab:model_variants}).

\textbf{Performance Analysis.} The in-distribution (ID) test set comprises 240 trajectories across 20 variables, each with 21 timesteps (${\approx}10^5$ samples). As shown in Table~\ref{tab: results 4x4}, Phaedra achieves a new state-of-the-art, reducing nMAE by over $40\%$ versus the strongest baselines (FSQ, VAR$_{\mathrm{large}}$) at a comparable token budget. The significantly lower $\Delta\sigma^2_{\mathrm{loc}}$ and $\gamma_{\min}$ near $100\%$ confirm the dual-latent strategy effectively eliminates spectral cutoff artifacts that afflict standard quantizers. Phaedra also approaches the theoretical performance ceiling of discrete tokenization: while a continuous autoencoder achieves the lowest in-distribution error, Phaedra closes most of that gap (SM Table~\ref{tab: results continuous vs. phaedra}), and—as discussed in Section~\ref{sec:generalization}—discrete methods yield better OOD robustness.

\textbf{Ablation Studies.} Two ablations validate architectural choices (Table~\ref{tab: results 4x4}, bottom rows). Replacing the 1-D amplitude codebook ($L_\alpha=[1024]$) with a 5-D FSQ codebook $[4,4,4,4,4]$ (same cardinality, matching VQ-VAE-2 style) confirms the advantage of learning scalar embeddings for physical intensity. A residual-learning ablation modifying Phaedra to operate sequentially (coarse 1-D tokens then 8-D residuals, akin to RQ-VAE \citep{leeRQVAE2022}) outperforms baselines but remains inferior to the parallel dual-latent design.

\begin{table*}[htbp] % Change to \begin{table}[htbp] if you are in a single-column document
    \centering
    \begin{minipage}[t]{0.48\textwidth}
        \centering
        \caption{\textbf{Tokenizer Reconstruction Tasks.} Summary metrics are averaged across variables. ID: in-distribution test split. OD$_1$: same PDEs, different conditions. OD$_2$: entirely new PDEs. Best results in bold.}
        \label{tab: results 4x4}
        \vspace{0.1cm} % Optional: adds a tiny bit of breathing room below the caption
        \resizebox{\linewidth}{!}{
            \begin{tabular}{llcccc} % Corrected to 6 columns
            \toprule
            Model & Dataset & nMAE$\downarrow$ & nRMSE$\downarrow$ & $\Delta \sigma_{loc}^2$$\downarrow$ & $\gamma_{min}$$\uparrow$ \\
            \midrule
            \multirow{3}{*}{VQ-VAE-2} & ID  & 3.024 & 5.069 & 15.02 & 79.1\% \\
            & OD$_1$ & 7.195 & 13.139 & 50.14 & 47.1\% \\
            & OD$_2$ & 7.901 & 11.092 & 17.06 & 51.4\% \\
            \midrule
            \multirow{3}{*}{FSQ} & ID  & 2.603 & 4.292 & 11.29 & 85.3\% \\
            & OD$_1$ & 1.878 & 4.331 & 20.55 & 93.9\% \\
            & OD$_2$ & 4.865 & 6.675 & 11.66 & 64.3\% \\
            \midrule
            \multirow{3}{*}{IBQ} & ID  & 8.492 & 13.186 & 44.65 & 37.9\% \\
            & OD$_1$ & 22.355 & 28.144 & 50.14 & 45.7\% \\
            & OD$_2$ & 18.574 & 24.918 & 55.02 & 18.5\% \\
            \midrule
            \multirow{3}{*}{VAR$_{large}$} & ID  & 3.229 & 5.112 & 13.33 & 81.7\% \\
            & OD$_1$ & 2.179 & 4.486 & 16.08 & 92.0\% \\
            & OD$_2$ & 6.711 & 9.145 & 14.78 & 58.8\% \\
            \midrule
            \multirow{3}{*}{VAR$_{small}$} & ID  & 4.056 & 6.266 & 16.04 & 75.6\% \\
            & OD$_1$ & 2.630 & 5.361 & 19.03 & 88.1\% \\
            & OD$_2$ & 10.143 & 14.218 & 21.65 & 38.3\% \\
            \midrule
            \multirow{3}{*}{Phaedra} & ID & \textbf{1.522} & \textbf{2.489} & \textbf{5.96} & \textbf{93.6}\% \\
            & OD$_1$ & \textbf{1.224} & \textbf{2.435} & \textbf{6.47} & \textbf{98.1}\% \\
            & OD$_2$ & \textbf{3.147} & \textbf{4.237} & \textbf{5.83} & \textbf{77.6}\% \\
            \bottomrule
            \end{tabular}
        }
    \end{minipage}
    \hfill
    \begin{minipage}[t]{0.48\textwidth}
        \centering
        \caption{\textbf{Downstream Tasks.} Relative $L1$ errors (\%), averaged across variables. KH, RC: in-distribution; RKH: out-of-distribution (unseen during tokenizer training). Full details in \ref{sec: downstream task results}.}
        \label{tab:downstream}
        \vspace{0.1cm} % Optional: adds a tiny bit of breathing room below the caption
        \resizebox{\linewidth}{!}{
            \begin{tabular}{llccc} % Corrected to 8 columns
            \toprule
            Task & Model & KH & RC & RKH \\
            \midrule
            \multirow{4}{*}{Op.\ Learning} 
            & FNO     & 12.55 & 35.63 &  16.66  \\
            & CNO     & 10.15 & 29.92 & 11.82 \\
            & ViT     & \textbf{9.18} & 39.68 & 13.20 \\
            & Cont.      & 9.28 & 119.0 & 11.56 \\
            & VQ-VAE-2 & 15.14 & 42.40 & 24.52 \\
            & FSQ    & 11.28 & 38.05 & 17.99 \\
            & Phaedra & 9.50 & \textbf{27.21} & \textbf{10.23} \\
            \midrule
            \multirow{2}{*}{MAE (75\%)}
            & FSQ & 5.84 & 29.16 & 9.45 \\
            & Phaedra & \textbf{4.75} & \textbf{22.17} & \textbf{7.27} \\
            \bottomrule
            \end{tabular}
        }
        \includegraphics[width=0.9\linewidth]{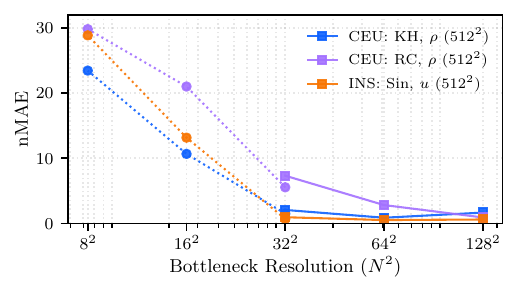}
        \caption{\textbf{Token Scaling Analysis.} Performance scales with total token count, independent of input resolution or downsampling factor. Dotted lines: low-resolution inputs.}
        \label{fig:nmae scaling}
    \end{minipage}
\end{table*}

\subsection{Scaling Laws \& Vision Foundation Model Comparison}

\paragraph{Token Budget vs.\ Resolution.} We scale latent resolution from $8^2$ to $128^2$ across low-resolution ($128^2$) and high-resolution ($512^2$) inputs (Figure~\ref{fig:nmae scaling}). Performance is primarily determined by the \emph{total token count} $N^2$, independent of the specific downsampling factor, with saturation beginning at $32^2$—our Pareto-optimal operating point. Additional metrics are in \ref{sec: scaling results with respect to bottleneck resolution}. We also find that even with amplitude codebook size $L_\alpha=32$, Phaedra outperforms FSQ, with gains leveling off near $L_\alpha=[512]$ (Figure~\ref{fig:Scaling wrt amplitude codebook size}).

\textbf{Compression Efficiency.} Space savings may be calculated as $\left(1-\frac{\text{Compressed Size}}{\text{Uncompressed Size}}\right)\times100\%$. Compressing a 32-bit, $128\times128$ resolution input to a latent resolution of $32\times32$ with 1024 amplitude tokens (10 bits) and 8640 morphology tokens (14 bits) would result in 95.3\% savings. We validate this by compressing a 65.0 GB NetCDF training dataset to 3.44 GB, totaling 94.7\% savings. In benchmark comparisons, Phaedra achieves this reduction with an ID nMAE of 1.522; this nearly doubles the compression of a continuous AE (50.0\% savings, 0.672 nMAE) while preserving markedly higher fidelity than FSQ (97.2\% savings, 2.603 nMAE). On \textbf{high-resolution} ($512\times512$) 32-bit 2D, the theoretical space savings are 98.9\% and 99.7\% when using $8\times8$ and $16\times16$ downsampling, respectively.

\paragraph{Comparison with Nvidia Cosmos.} We benchmark against the Nvidia Cosmos DI Tokenizer ($8{\times}8$ and $16{\times}16$) on $512^2$ PDE data. Since Cosmos expects pixel values $\in [0,1]$, we evaluate it in a best-case scenario where each sample is individually normalized to this range (Cosmos$^{0:1}$). As shown in Table~\ref{tab: cosmos comparison}, general-purpose image tokenizers struggle with the high dynamic range of PDE data. At $16^2$ compression, Cosmos exhibits severe spectral smoothing ($\gamma_{\min}\approx60\%$), while Phaedra reduces nMAE by nearly $6\times$ and maintains topological correctness ($\gamma_{\min}\approx86\%$). Phaedra also outperforms the image-native Cosmos on grayscale ImageNet (Table~\ref{tab: cosmos comparison}, bottom), confirming it is a robust general-purpose compressor.

\begin{figure}[htbp]
    \centering
    \begin{minipage}[c]{0.44\textwidth}
    
    \begin{subfigure}[b]{0.3\linewidth}
        \centering
        \includegraphics[trim={15pt 15pt 90pt 15pt},clip, width=\textwidth, height=\textwidth]{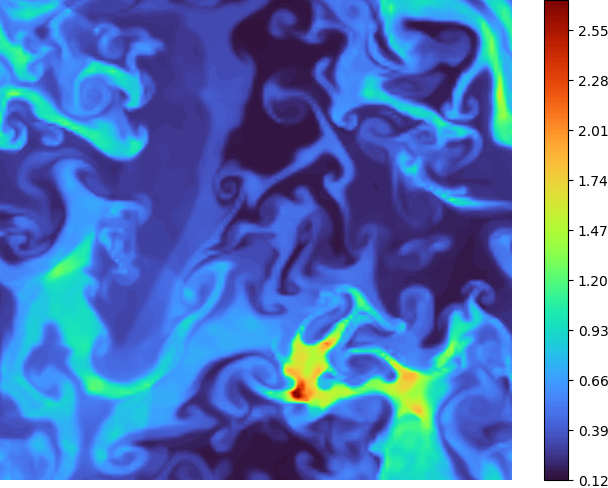}
    \end{subfigure}
    \hspace{0.01cm}
    \begin{subfigure}[b]{0.3\linewidth}
        \centering
        \includegraphics[trim={15pt 15pt 90pt 15pt},clip, width=\textwidth, height=\textwidth]{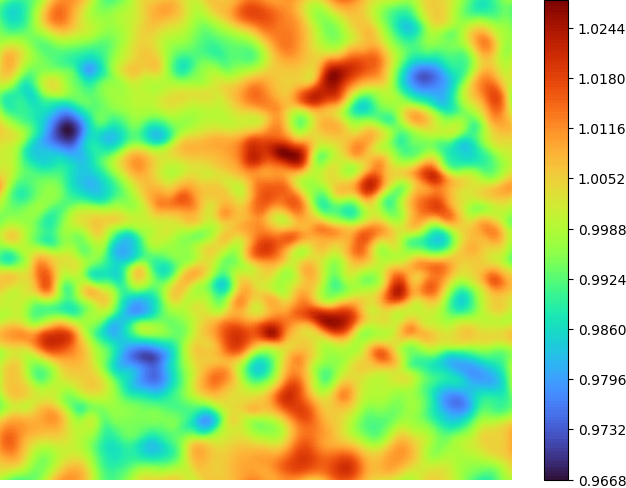}
    \end{subfigure}
    \hspace{0.01cm}
    \begin{subfigure}[b]{0.3\linewidth}
        \centering
        \includegraphics[trim={15pt 15pt 90pt 15pt},clip, width=\textwidth, height=\textwidth]{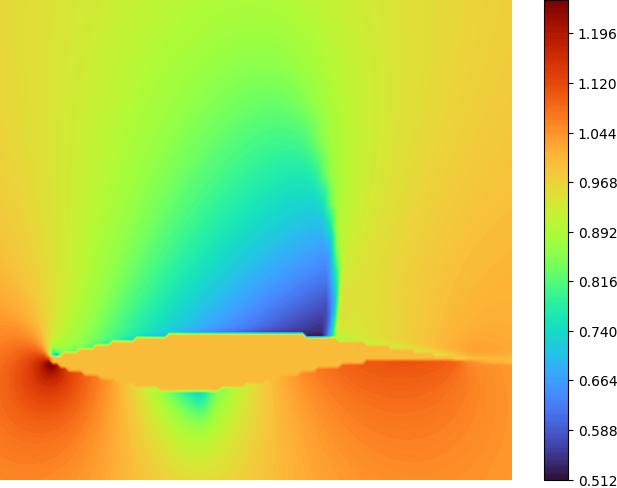}
    \end{subfigure}
    \hfill

    \vspace{0.05cm} % Add small vertical space between rows

    \begin{subfigure}[b]{0.3\linewidth}
        \centering
        \includegraphics[trim={15pt 15pt 90pt 15pt},clip, width=\textwidth, height=\textwidth]{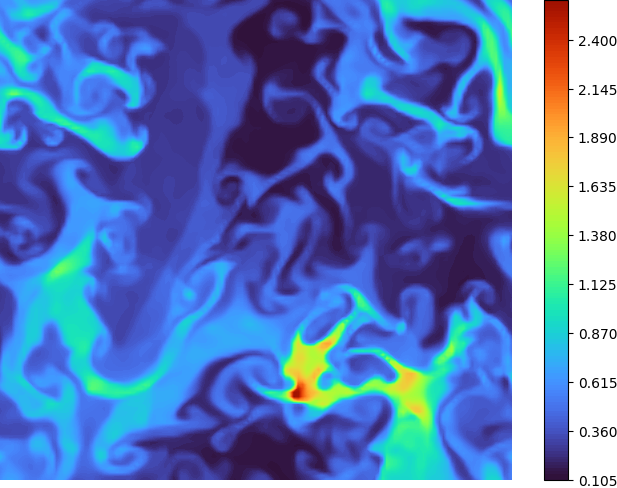}
    \end{subfigure}
    \hspace{0.01cm}
    \begin{subfigure}[b]{0.3\linewidth}
        \centering
        \includegraphics[trim={15pt 15pt 90pt 15pt},clip, width=\textwidth, height=\textwidth]{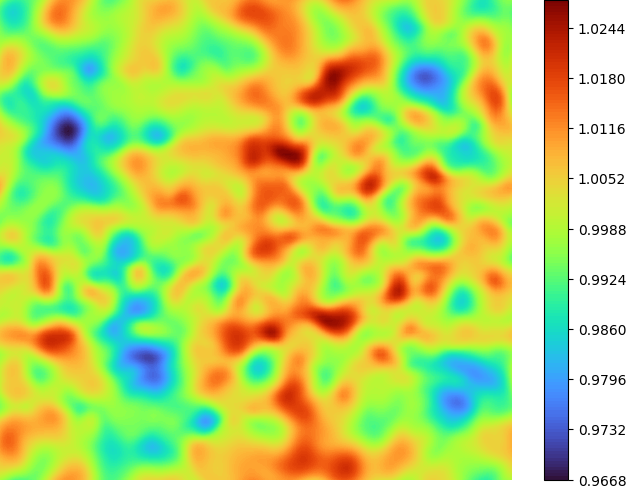}
    \end{subfigure}
    \hspace{0.01cm}
    \begin{subfigure}[b]{0.3\linewidth}
        \centering
        \includegraphics[trim={15pt 15pt 90pt 15pt},clip, width=\textwidth,  height=\textwidth]{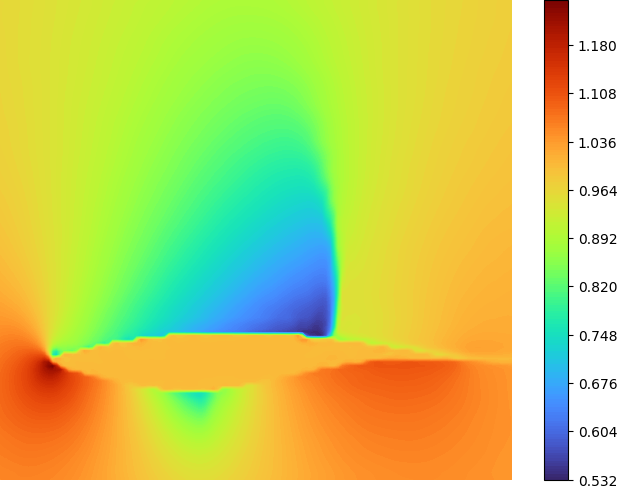}
    \end{subfigure}
    
    \caption{\textbf{PDE Reconstructions.} Inputs (top) and reconstructions (bottom). Columns left to right: CEU RC $\rho$, CEU KH $p$, Airfoil.}
    \label{fig:CEU_reconstructions}
    \end{minipage}
    \hfill
    \begin{minipage}[c]{0.44\textwidth}
        \centering
        \includegraphics[trim={30pt 200pt 10pt 30pt},clip,width=\linewidth]{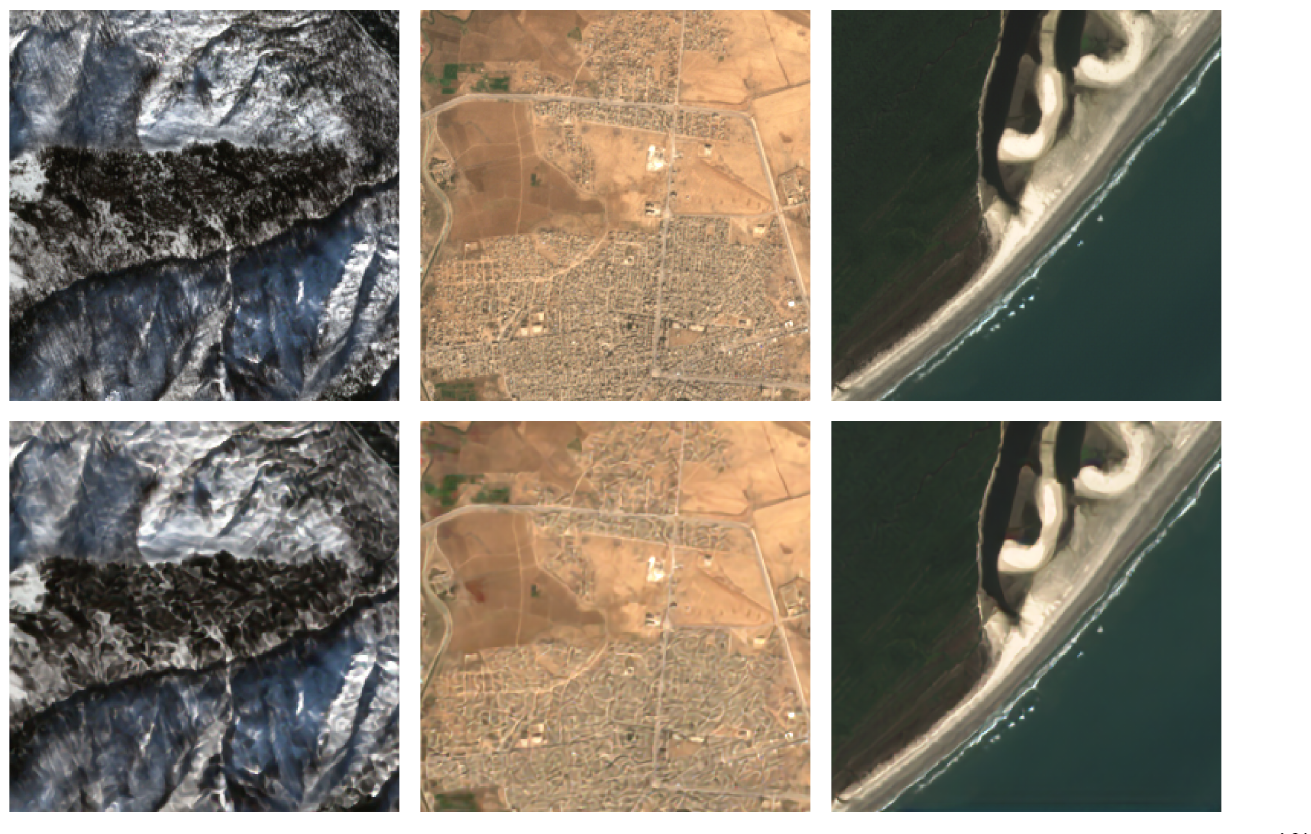}
        \caption{\textbf{Sentinel-2 L1C Reconstructions.} Inputs (top) and reconstructions (bottom) for various locations (full details: \ref{sec: earth observation eo data figures samples}).}
        \label{fig:eo_reconstructions}
    \end{minipage}
\end{figure}

\subsection{Zero-Shot Cross-Domain Generalization}
\label{sec:generalization}
We evaluate generalization without any further training across four settings. Reconstruction examples for all tasks are provided in \ref{sec: reconstruction figures}.

\textbf{OD$_1$ / OD$_2$.} OD$_1$ contains Euler/NS problems with unseen initial and boundary conditions (curved Riemann problems, airfoil flows, opposing-velocity NS). OD$_2$ contains entirely new PDEs: Poisson, Darcy flow, Allen-Cahn, and the Acoustic Wave equation. Results are reported in Table~\ref{tab: results 4x4}. Critically, while the continuous autoencoder degrades by $70\%$ on OD$_1$ (nMAE), Phaedra's error \emph{improves}, suggesting that discrete tokenization acts as an implicit regularizer, forcing the model to learn transferable physical primitives (see \ref{sec: Error Analysis of quantization} for an empirical error-decomposition analysis).

\textbf{Earth Observation.} On Sentinel-2 L1C (13 bands, $10$--$60$m resolution; Table~\ref{tab:sentinel 2 l1c results}, Figure~\ref{fig:eo_reconstructions}), Phaedra$_4$ achieves $\gamma_{\min}\approx93\%$ and outperforms FSQ$_4$ across all metrics. At $8^2$ downsampling, Cosmos exhibits catastrophic failure in local variance preservation ($\Delta\sigma^2_{\mathrm{loc}}\approx1.9{\times}10^4$), while Phaedra remains stable. A global evaluation over 64,000 samples across diverse biomes (details in \ref{sec: earth observation numerical results}) confirms consistent rankings. We extend zero-shot evaluation to Sentinel-1 RTC radar, DEMs, and vegetation indices.

\textbf{Weather.} On ERA5 reanalysis (temperature, zonal/meridional winds), Phaedra$_4$ reduces $rL_2$ error from $14.003\pm0.482$ (FSQ) to $9.117\pm0.362$, closing $55.85\%$ of the gap to the continuous autoencoder—demonstrating that the learned morphological basis captures fundamental atmospheric primitives, not synthetic-data artifacts. 

\begin{figure}[h!]
    \centering
    \begin{minipage}[c]{0.48\textwidth}
        \centering
        \captionof{table}{High-resolution ($512^2$) PDE benchmarks and 200 ImageNet samples. Cosmos$^{0:1}$: per-sample $[0,1]$ normalization for a best-case comparison.}
        \label{tab: cosmos comparison}
        \resizebox{\linewidth}{!}{
        \begin{tabular}{llcccc}
        \toprule
        Model & Data & nMAE$\downarrow$ & nRMSE$\downarrow$ & $\Delta \sigma_{loc}^2$$\downarrow$ & $\gamma_{min}$$\uparrow$ \\
        \midrule
        Phaedra$_{8}$       & PDE & 1.20 & 2.35 & 19.81 & 95.3\% \\
        Cosmos$^{0:1}_{8}$  & PDE & 5.73 & 7.12 & 30.44 & 80.4\% \\
        Phaedra$_{16}$      & PDE & 2.33 & 4.29 & 44.42 & 86.2\% \\
        Cosmos$^{0:1}_{16}$ & PDE & 14.53 & 17.10 & 64.08 & 60.2\% \\
        \midrule
        Phaedra$_{8}$  & ImageNet & 2.64 & 4.33 & 72.04 & 90.3\% \\
        Cosmos$_{8}$   & ImageNet & 2.70 & 4.16 & 37.23 & 88.6\% \\
        Phaedra$_{16}$ & ImageNet & 3.65 & 5.75 & 107.8 & 70.2\% \\
        Cosmos$_{16}$  & ImageNet & 4.83 & 7.34 & 72.13 & 44.1\% \\
        \bottomrule
        \end{tabular}}
    \end{minipage}
    \hfill
    \begin{minipage}[c]{0.48\textwidth}
        \centering
        \captionof{table}{\textbf{Sentinel-2 L1C.} 13 bands at 60m/20m/10m resolution; metrics on native resolution.}
        \label{tab:sentinel 2 l1c results}
        \resizebox{\linewidth}{!}{
        \begin{tabular}{lcccc}
        \toprule
            Model & r$L_1$ $\downarrow$ & r$L_2$ $\downarrow$ & $\Delta \sigma^2_{loc}$ $\downarrow$ & $\gamma_{min}$ $\uparrow$\\
        \midrule
            Continuous   & 7.426 & 8.100 & 62.61 & 90.02\% \\
        \midrule 
            Phaedra$_4$  & 8.895 & 9.749 & 128.5 & 93.17\%  \\ 
            FSQ$_4$      & 11.053 & 12.405 & 215.8 & 77.62\% \\
        \midrule
            Phaedra$_8$  & 9.900 & 11.475 & 163.9 & 62.72\% \\
            Cosmos$_8$   & 16.717 & 19.245 & 19{,}566 & 79.70\%  \\
        \bottomrule
        \end{tabular}}
    \end{minipage}
\end{figure}

\subsection{Downstream Task Evaluation}
To confirm that higher tokenization fidelity translates to improved task performance, we provide an initial proof-of-concept evaluation on downstream operator learning using two ID (KH, RC) and one OD$_1$ (RKH) compressible Euler datasets. We emphasize that this section is not intended as an expansive study on optimal transformer architectures for operator learning, but rather a demonstration that Phaedra's discrete tokens provide a viable, high-fidelity substrate for standard discrete architectures. In the following experiments, we use the frozen encoders to tokenize the continuous input fields, saving these results to train the downstream transformer. For testing, we use the frozen decoders to reconstruct the original fields from the downstream model predictions.

\paragraph{Operator Learning.} We frame operator learning as a non-autoregressive sequence-to-sequence (seq2seq) task. We train a 38M parameter transformer encoder to process this sequence, using self-attention to build a global representation of the initial physical state. A transformer decoder then uses cross-attention to condition on this representation and generates the latent sequence for a future state. To accommodate Phaedra's dual-channel design, we construct a bifurcated decoder: one head outputs a categorical distribution (via cross-entropy loss) over the discrete morphological codebook, and the other predicts a distribution over the amplitude codebook. We follow the all2all training scheme from \cite{herde2024poseidonefficientfoundationmodels}, where all input timesteps are mapped to all future output timesteps. For 7 timesteps, this results in 28 input-output pairs. For the transformer, we use 2D rotary positional embeddings \cite{heo2024rotarypositionembeddingvision}, as well as an embedding to designate the time difference of the input-output pair.

\begin{figure}[h!]
    \centering
    \includegraphics[width=0.98\linewidth, trim={0 395 0 0}, clip]{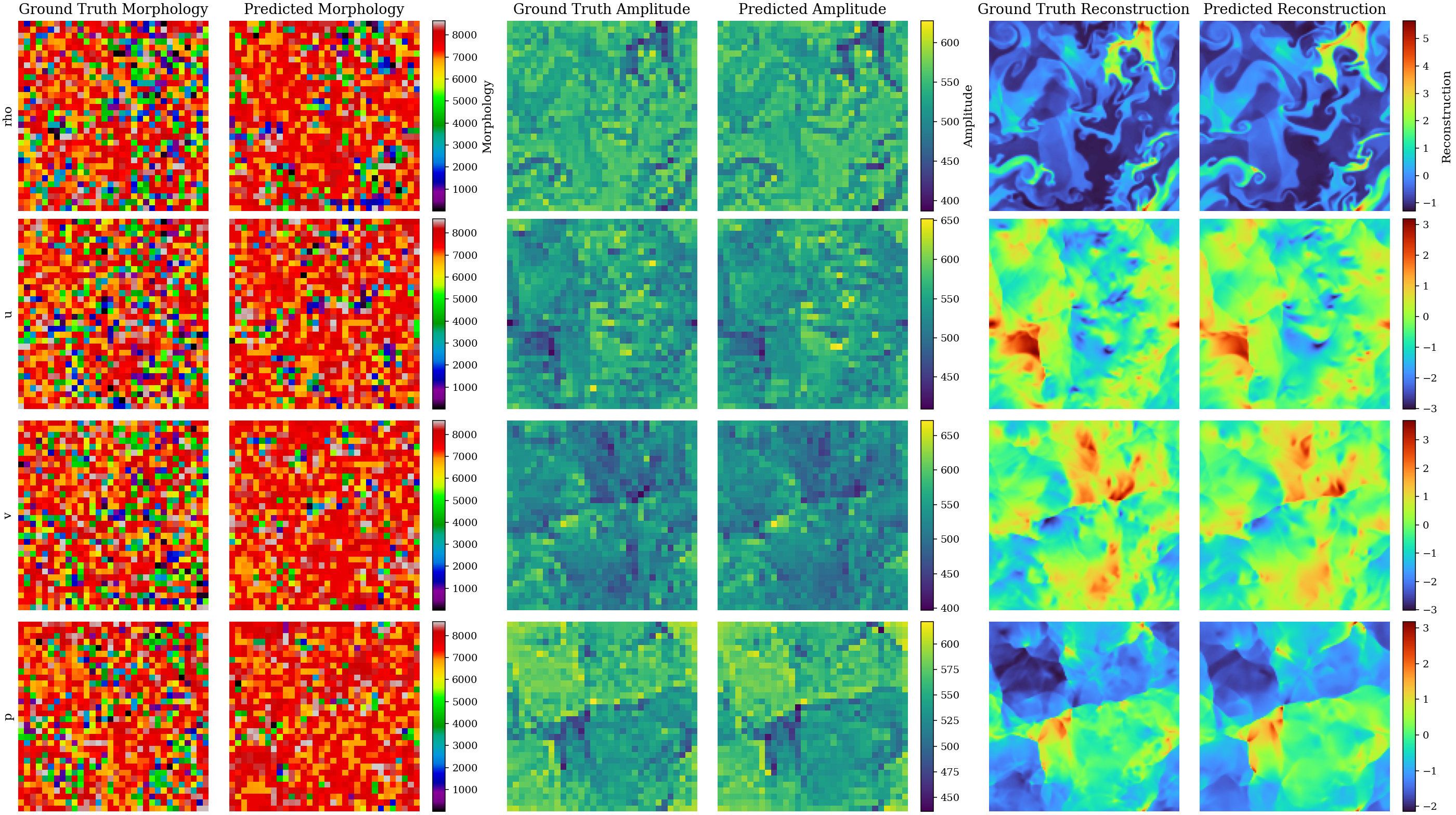}
    \caption{Downstream Operator Learning. A downstream transformer is trained using cross-entropy loss to predict the dynamics of the compressible Euler equations with the Riemann-curved (RC) initial conditions. The results show the prediction at the final timestep, using only the initial conditions as inputs. The amplitude tokens clearly capture much of the overall dynamics in density (top) and $u-$velocity (bottom), while fine structure is added via morphology token prediction. Additional figures for all downstream tasks are available in \ref{sec: downstream task results}.}
    \label{fig: downstream operator learning main text}
\end{figure}

We predict future timesteps from initial conditions, comparing against a continuous Fourier Neural Operator (FNO), Convolutional Neural Operator (CNO), Vision Transformer (ViT), and transformers trained on the FSQ, VQ-VAE-2, and Continuous tokens. We report the relative $L_1$ errors on the last timestep given the initial conditions, averaged across all 4 physical variables ($\rho, u, v, p$) for 240 test trajectories in each dataset. Results are available in Table~\ref{tab:downstream}, with an example of the predictions displayed in Fig.~\ref{fig: downstream operator learning main text}. The Phaedra transformer consistently matches or outperforms the baselines. While FNO evaluates faster at inference ($0.013$s vs.\ $0.040$s per sample), Phaedra provides significantly higher predictive accuracy. The continuous AE embeddings provide marginally more accurate predictions for KH, but blow-up on RC. This suggests that discreteness alone is not the mechanism which provides more accurate results, but it may additionally provide regularization which stabilize training to focus on learning dynamics. In Section \ref{sec: downstream task results}, we also present a scaling study across 5/38/200M parameters, demonstrating monotonic improvements in prediction accuracy across all timesteps with model size.

\paragraph{Masked Autoencoding (MAE).} To test spatial correlation and interpolation, we train a seq2seq transformer on a masked token reconstruction task. Here, $75\%$ of the input tokens are randomly replaced with a learned mask embedding, and the model must reconstruct the missing tokens from the visible context. We compare Phaedra against standard FSQ tokens. In this setting, we use only one decoder head for both models, training identical architectures--only the tokenized data itself differs. Additionally, we pretrain the models on all datasets and subsequently fine-tune for each dataset, exhibiting the practical use of such transformers as pretrained foundation models. Phaedra tokens yield consistent improvements across all datasets and metrics (Table~\ref{tab:downstream}), with the largest gains on RC $rL_1$ ($22.17$\% vs.\ $29.16$\%), confirming that Phaedra's richer discrete representation directly benefits spatial pretraining objectives. 

Full experimental details, per-variable results, and visualizations of generated tokens and reconstructions are provided in \ref{sec: downstream task results}.

\section{Discussion.}

\textbf{Related Work.} Discrete latent spaces, popularized by VQ-VAE \citep{van2017neural} and refined by VQ-GAN \citep{esser2021taming}, are foundational to modern generative modeling. However, their reliance on perceptual losses (e.g., LPIPS) is ill-suited for scientific data, as such metrics are invariant to high-frequency spectral details and precise physical magnitudes. Recent advances improve representational capacity via hierarchical or hybrid structures to preserve fine details (VQ-VAE-2 \citep{razavi2019generating}, HQ-VAE \citep{takida2024hqvaehierarchicaldiscreterepresentation}, HART \citep{tang2024hart}). Parallel efforts enhance codebook efficiency and stability through techniques like Lookup-Free Quantization (MAGVIT-v2 \citep{yu2024languagemodelbeatsdiffusion}), Index Backpropagation (IBQ \citep{shi2025scalableimagetokenizationindex}), and Finite Scalar Quantization (FSQ \citep{mentzer2023finite}). Despite these innovations, these architectures remain optimized for the bounded, uniform statistics of natural images, struggling with the heavy-tailed, unbounded distributions characteristic of physical fields. A strong parallel line of work forgoes discretization altogether and models dynamics generatively in a \emph{continuous} latent space \citep{gencfd}; we do not claim that tokenization dominates this paradigm, and we note that token-based multi-modal modeling carries unresolved alignment problems of its own between modality-specific vocabularies.
 
\textbf{Why Discrete Tokenization?} Our case for a discrete representation rests on three arguments. \emph{Empirically}, discrete tokenizers degrade more gracefully out of distribution than a continuous autoencoder of matched capacity (Section~\ref{sec:generalization}), while Phaedra closes most of the in-distribution reconstruction gap that quantization opens (Table~\ref{tab: results continuous vs. phaedra}); downstream, this buys parity with strong continuous pipelines on our smoothest dataset and a clear advantage wherever discontinuities or distribution shift dominate. \emph{Practically}, tokenization is an effective nonlinear compressor: our $65.0$ GB training corpus is reduced to $3.44$ GB ($94.7\%$) at an nMAE of $1.522$, which keeps data movement from hindering training at the scales foundation models demand. \emph{Methodologically}, discrete tokens are the native representation of the vision and vision--language communities, so methodological advances there transfer directly to models built on discrete scientific tokens. MaskGIT \citep{chang2022maskgit}, for instance, matches or exceeds strong generative baselines while constructing an image by unmasking tokens in roughly an order of magnitude fewer decoding steps. Phaedra is intended as a bridge between these communities.
 
\textbf{Main Results.} In this work, we introduced \textbf{Phaedra}, a novel dual-channel tokenizer designed specifically for scientific datasets. Inspired by Shape-Gain Quantization \citep{Buzo1980ShapeGain, gersho1992vector}, Phaedra learns a nonlinear factorization where a vector codebook captures the manifold of normalized physical structures (morphology), while a continuous scalar channel retains precise, heavy-tailed energy dynamics (amplitude). Evaluating across in-distribution PDEs and diverse out-of-distribution datasets (unseen PDE conditions, novel PDEs, Weather, and Earth Observation data), our results support four conclusions:
\begin{enumerate}
\item \emph{Phaedra consistently outperforms image tokenizers, including Nvidia Cosmos}, on continuous physical data, particularly in preserving the energy spectra and local variance that standard tokenizers smooth out.
\item Discrete tokenizers---with Phaedra leading---\emph{degrade less under zero-shot transfer} than a continuous autoencoder of matched capacity. This suggests that Phaedra's learned ``morphological dictionary'' captures transferable physical primitives rather than overfitting to dataset-specific statistics.
\item We demonstrate that \emph{2D gridded scientific data can be compressed at rates comparable to natural images} ($16^2\times$) without sacrificing high-frequency fidelity, enabled directly by isolating macroscopic energy from structural details.
\item \emph{Tokenization quality transfers to downstream tasks}: on operator learning and masked autoencoding, a standard transformer trained on Phaedra tokens matches strong continuous pipelines on the smoothest problem and outperforms every baseline where dynamic range, discontinuities, or distribution shift dominate.
\end{enumerate}
 
\textbf{What the Downstream Results Show.} To probe the fourth conclusion, we compare five representations within one identical pipeline (the same seq2seq transformer at ${\approx}38$M parameters, the same all2all scheme, and the same training budget), so that the representation is the only variable: Phaedra tokens, FSQ tokens, VQ-VAE-2 tokens, frozen continuous-autoencoder latents with a regression head, and raw fields patched by a ViT with no tokenizer at all. FNO and CNO serve as parameter-matched operator baselines. Phaedra versus FSQ isolates the factorization alone, Phaedra versus VQ-VAE-2 the choice of discrete family, Phaedra versus the continuous transformer discreteness itself, and Phaedra versus the ViT the value of a learned representation at all. On KH, our smoothest problem, Phaedra, the ViT, and the continuous transformer are statistically indistinguishable at the final timestep. The advantage of a tokenizer built for physical fields instead concentrates where the physics is hard: on the shock-dominated RC data and under distribution shift on RKH, Phaedra leads every baseline, while the continuous transformer fails to converge on RC entirely ($119.05\%$ versus $27.21\%$ relative $L_1$). 
Continuous embeddings vary rapidly between adjacent tokens, obscuring the large-scale dynamics unless the latent space is regularized (as by quantization) or learned jointly with the dynamics, as in the ViT.
 
Two further results indicate that the representation is a substrate for pretraining and not only for single-task fitting. Under $75\%$ masked-token reconstruction with identical architectures, Phaedra tokens improve every dataset and metric over FSQ tokens. 
Our scaling study shows error decreases monotonically with model capacity at every horizon from $5$M to $200$M parameters with no sign of saturation (Table~\ref{tab:model scaling across timesteps}). We stress that this remains an initial proof of concept: a full study of token-based operator learning, and full multi-dataset foundation-model pretraining, lie beyond the scope of this paper.
 
\textbf{Limitations and Future Work.} We tokenize each field component independently, which ignores the inter-variable coupling inherent to systems such as the compressible Euler equations. We did not train a joint-channel tokenization variant, so we cannot yet quantify what independence sacrifices. On disentanglement, our convolutional decoder modulates structural and amplitude information nonlinearly, buying expressiveness at the cost of exact separation. Finally, Phaedra has been evaluated on 2D PDEs defined on uniform grids. Extension to three dimensions is conceptually direct via a 3D convolutional autoencoder, and our pilot (Table~\ref{tab:3d pilot experiment}) reproduces the 2D pattern, but a complete 3D evaluation across PDE families remains future work; point clouds and irregular meshes instead require replacing the convolutional encoder and decoder with geometry-aware alternatives, which is non-trivial. Moving forward, we aim to study discrete-transformer operator learning in greater depth, and to pursue the property that motivates tokenization but that this paper does not yet demonstrate: a single shared vocabulary in which fields in different dimensions, on different geometries, and symbolic tokens describing the governing equations are processed by one transformer.
\newpage

\paragraph{Acknowledgments.}

This work was made possible through the support of the ETH AI Center by a PhD fellowship awarded to Levi Lingsch.

\bibliographystyle{plainnat}
\bibliography{bibliography}

\clearpage
\appendix

\renewcommand{\appendixname}{Supplementary Material}
\renewcommand{\thesection}{SM.\arabic{section}}
\renewcommand{\thesubsection}{\thesection.\arabic{subsection}}

\setcounter{part}{0} 

\setcounter{tocdepth}{2} 

\renewcommand{\ptctitle}{Table of Contents}

\part{Supplementary Material}

\begin{minipage}{\textwidth}
    \parttoc
\end{minipage}

\begin{figure}[b]
    \centering
    \includegraphics[trim={0 20 0 0}, clip, width=0.98\linewidth]{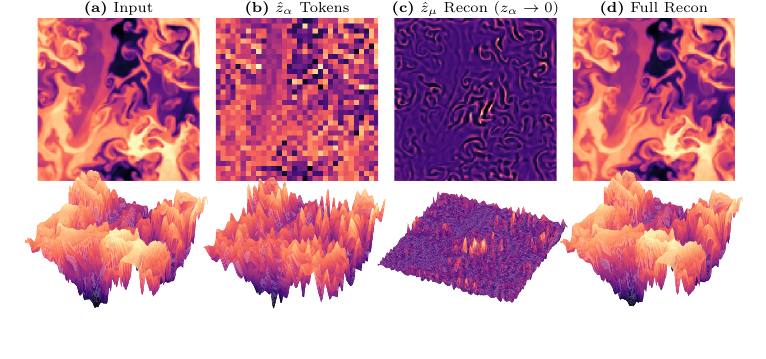}
    \caption{An example of an input, its amplitude tokens, and reconstructions from the (i) morphological tokens with amplitude tokens set such that their embeddings are equal to zero and (ii) full set of tokens. }
    \label{fig:placeholder}
\end{figure}

\clearpage
\section{Error Analysis of Discrete Tokenizers}
\label{sec: Error Analysis of quantization}
\paragraph{Notation.} Let $x$ denote a sample from $D$, the data distribution. An expectation $\mathbb{E}[\cdot]$ is defined over samples $x \sim D$. The encoder $\mathcal{E}$ produces continuous embeddings at scales $s \in S$. The embeddings are given as $z = \{ z^{(k)} \}, z^{(k)} \in \mathbb{R}^{d_k}$. The quantizer $Q$ provides discrete approximations of the continuous embeddings, $\hat{z}^{(k)}=Q(z^{(k)})$. The decoder $\mathcal{D}$ provides reconstructions of the discrete embeddings $\hat{x}_{quant} = \mathcal{D}(\hat{z}^{(k)})$ or continuous embeddings $\hat{x}_{cont} = \mathcal{D}({z}^{(k)})$. The final error is given as 
\begin{equation}
    L_{Final}=\mathbb{E}[||x-\hat{x}_{quant}||^2].
\end{equation}

\paragraph{General error of reconstruction.} The effective difference between data and reconstruction may be decomposed into contributions arising from the autoencoder and quantizer,
\begin{equation}
    x - \hat{x}_{quant} = (x - \hat{x}_{cont}) + (\hat{x}_{cont} - \hat{x}_{quant}),
\end{equation}
where the final error becomes
\begin{equation}
    L_{Final}=\mathbb{E}[||x-\hat{x}_{cont}||^2]+\mathbb{E}[||\hat{x}_{cont}-\hat{x}_{quant}||^2] + 2\mathbb{E}[<x-\hat{x}_{cont}, \hat{x}_{cont}-\hat{x}_{quant}>].
\end{equation}
The correlation between the errors is bounded by the Cauchy-Schwarz inequality, 
\begin{equation}
    \mathbb{E}[<x-\hat{x}_{cont}, \hat{x}_{cont}-\hat{x}_{quant}>] \leq \sqrt{\mathbb{E}[||x-\hat{x}_{cont}||^2]} \cdot\sqrt{\mathbb{E}[||\hat{x}_{cont}-\hat{x}_{quant}||^2]}.
\end{equation}
The final loss is then 

\begin{equation}
    L_{Final} \leq \left( \sqrt{\mathbb{E}[||x-\hat{x}_{cont}||^2]} + \sqrt{\mathbb{E}[||\hat{x}_{cont}-\hat{x}_{quant}||^2]} \right)^2 = \left( \sqrt{L_{emb}} + \sqrt{L_{quant}} \right)^2.
\end{equation}

This result, while simple, clearly illustrates the fact that a quantization perturbation is introduced in discrete tokenizers through the quantization process. Results from numerical experiments confirm this bound: the performance of a continuous autoencoder represents an upper bound on the performance of a discrete quantizer. This finding is consistent with the empirical results of Table \ref{tab: results continuous vs. phaedra}.

\begin{table}[h]
\centering
\caption{Comparison of results for the continuous autoencoder compared to the discrete tokenizer \emph{Phaedra}. ID denotes the test split of the training dataset. OD$_1$ denotes out-of-distribution datasets which are still defined by the Euler equations. OD$_2$ denotes datasets which are governed by PDEs not present in the training data. For the same encoder-decoder family and comparable training, the continuous autoencoder is expected to be at least as expressive because it does not have the quantization constraint; in our experiments, it gives the best reconstruction accuracy. }
\resizebox{0.6\textwidth}{!}{
\begin{tabular}{llccccc}
\toprule
Model & Dataset & nMAE$\downarrow$ & nRMSE$\downarrow$ & $\Delta \sigma_{loc}^2$$\downarrow$ & $\gamma_{min}$$\uparrow$ \\
\midrule
\multirow{3}{*}{Continuous} & ID & 0.672 & 1.122 & 2.98 & 98.4\% \\ 
& OD$_1$ & 1.154 & 2.079 & 3.40 & 99.2\% \\ 
& OD$_2$ & 1.967 & 2.821 & 3.02 & 97.0\% \\ 
\midrule
\multirow{3}{*}{Phaedra} & ID & {1.522} & {2.489} & {5.96} & {93.6}\%  \\
& OD$_1$ & {1.224} & {2.435} & {6.47} & {98.1}\% \\ 
& OD$_2$ & {3.147} & {4.237} & {5.83} & {77.6}\% \\ 
\bottomrule
\end{tabular}}
\label{tab: results continuous vs. phaedra}
\end{table}

\subsection{Embedding Visualizations.}
\begin{figure}
    \centering
    \begin{subfigure}[b]{0.2\textwidth}
        \centering
        \includegraphics[width=\textwidth]{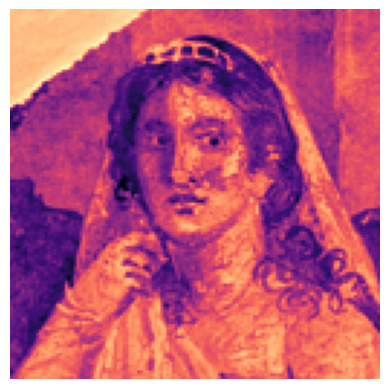}
        \caption{Input image.}
    \end{subfigure}
    \hfill % Adds spacing between figures
    
    \begin{subfigure}[b]{0.7\textwidth}
        \centering
        \includegraphics[width=\textwidth]{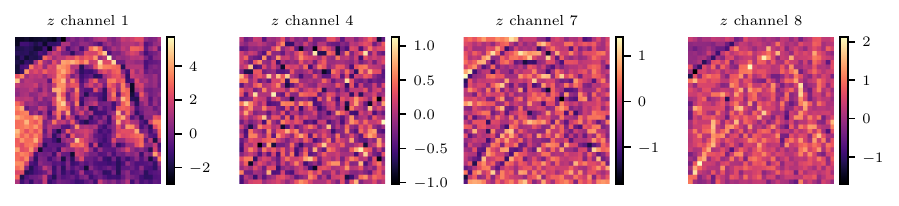}
        \caption{Encoder outputs $z$.}
    \end{subfigure}

    \begin{subfigure}[b]{0.7\textwidth}
        \centering
        \includegraphics[width=\textwidth]{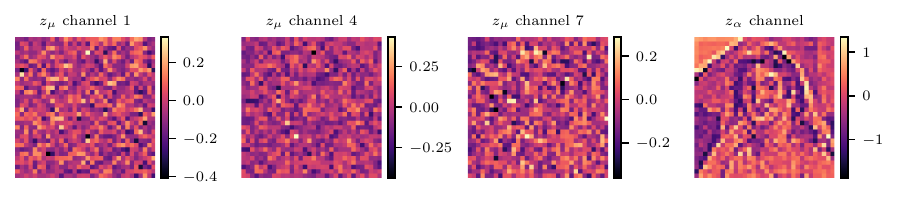}
        \caption{Encoder outputs, following convolution and split into morphological and amplitude embeddings $z_\mu$ and $z_\alpha$.}
    \end{subfigure}
    
    \begin{subfigure}[b]{0.7\textwidth}
        \centering
        \includegraphics[width=\textwidth]{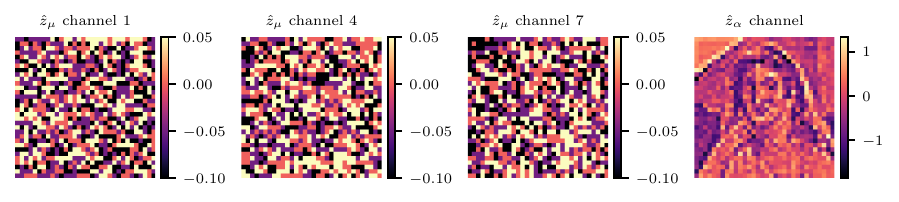}
        \caption{Discrete tokens $\hat{z}_\mu$ and $\hat{z}_\alpha$.}
    \end{subfigure}
    
    \begin{subfigure}[b]{0.7\textwidth}
        \centering
        \includegraphics[width=\textwidth]{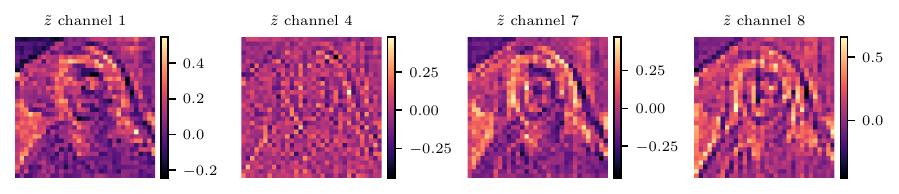}
        \caption{$\tilde{z}$, following a concatenation and recombination of the discrete tokens.}
    \end{subfigure}
    
    \caption{Plots of the encoder outputs, following along from the embeddings, through the segregation of morphological and amplitude components, the quantization step, and their eventual recombination before decoding.}
    \label{fig:phaedra embeddings visualizations}
\end{figure}

\begin{figure}
    \centering
    \begin{subfigure}[b]{0.2\textwidth}
        \centering
        \includegraphics[width=\textwidth]{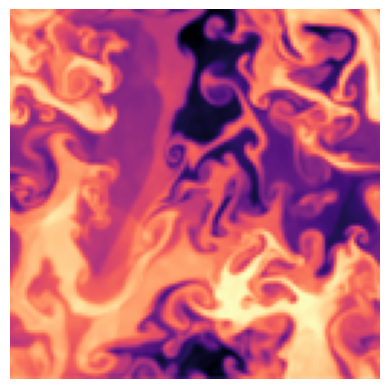}
        \caption{Input field.}
    \end{subfigure}
    \hfill % Adds spacing between figures
    
    \begin{subfigure}[b]{0.7\textwidth}
        \centering
        \includegraphics[width=\textwidth]{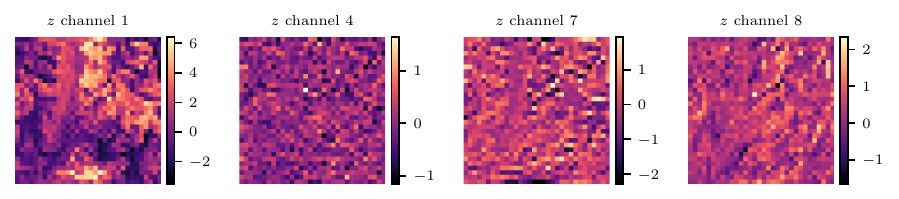}
        \caption{Encoder outputs $z$.}
    \end{subfigure}

    \begin{subfigure}[b]{0.7\textwidth}
        \centering
        \includegraphics[width=\textwidth]{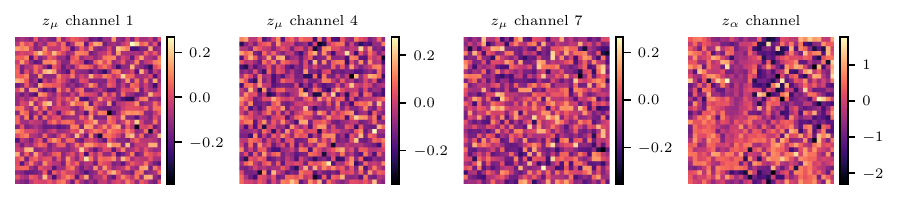}
        \caption{Encoder outputs, following convolution and split into morphological and amplitude embeddings $z_\mu$ and $z_\alpha$.}
    \end{subfigure}
    
    \begin{subfigure}[b]{0.7\textwidth}
        \centering
        \includegraphics[width=\textwidth]{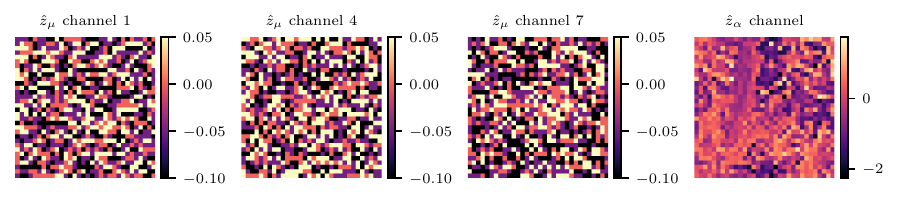}
        \caption{Discrete tokens $\hat{z}_\mu$ and $\hat{z}_\alpha$.}
    \end{subfigure}
    
    \begin{subfigure}[b]{0.7\textwidth}
        \centering
        \includegraphics[width=\textwidth]{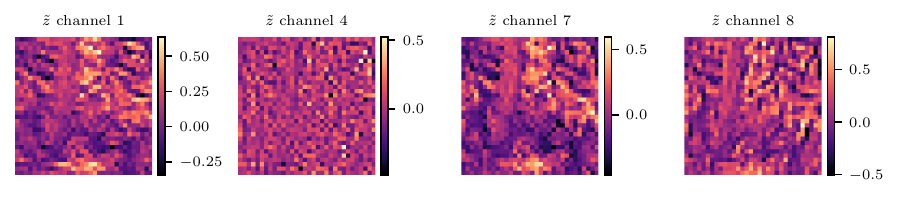}
        \caption{$\tilde{z}$, following a concatenation and recombination of the discrete tokens.}
    \end{subfigure}
    
    \caption{Plots of the encoder outputs, following along from the embeddings, through the segregation of morphological and amplitude components, the quantization step, and their eventual recombination before decoding.}
    \label{fig:ceu crp rho embeddings visualizations}
\end{figure}

The error induced by quantization primarily results from the loss of information in the latent space. Continuous embedding vectors are forced into bins, resulting in a loss of the granular details that contain additional information for the decoder to use during reconstruction. We present two examples visualizing the embeddings vis a vis tokens before and after quantization in Figures \ref{fig:phaedra embeddings visualizations} and \ref{fig:ceu crp rho embeddings visualizations}. There is a clear distinction before and after the quantization process, as a continuous spectrum of colors becomes a similar pattern containing very few distinct colors. 

\clearpage
\section{Extended Analysis of Amplitude--Morphology Disentanglement}
\label{sec: disentanglement}

In Section~\ref{sec: methods}, we introduced the dual-latent factorization designed to disentangle the local amplitude (1D tokens) from the field morphology (8D tokens). To rigorously validate that the network learns this intended physical factorization, and to address the \emph{soft} nature of this disentanglement caused by the nonlinear decoder, we conducted three distinct experiments.

\subsection{Token Swapping Between Uncorrelated Fields}
\begin{figure}[h]
    \centering
    \includegraphics[width=0.8\linewidth]{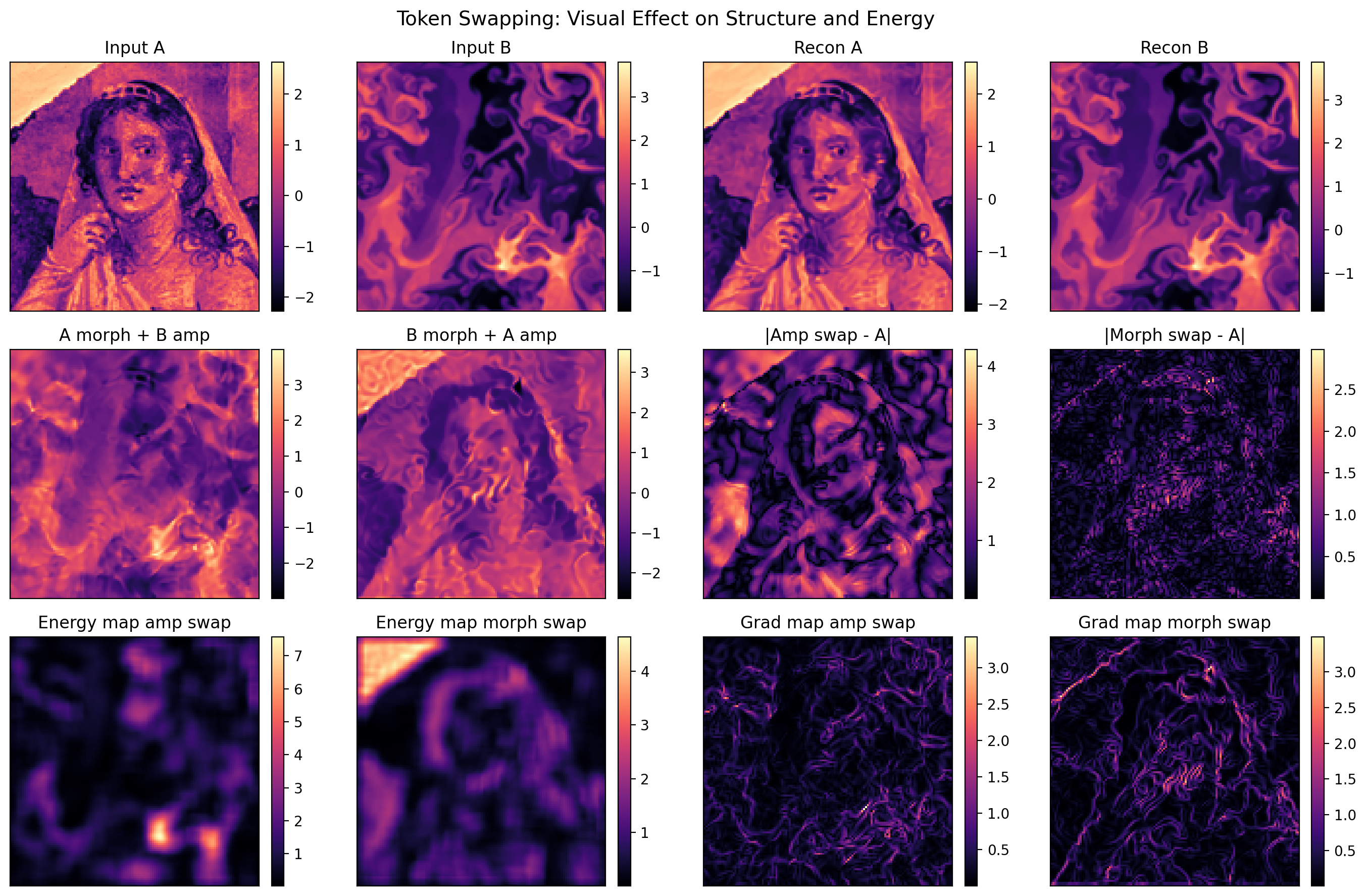}
    \caption{Token swapping of the amplitude and morphology for the Phaedra image (A) and a density field from the CEU RC dataset (B). The reconstruction using amplitude tokens from A with morphology tokens from B maintains overall pixel values consistent with image A, but introduces small-scale vortices and flow patterns from B.}
    \label{fig: token swap visualization}
\end{figure}

To test if energy and structure are truly isolated in their respective latent spaces, we performed a token swap between two completely uncorrelated fields: field $A$ (Phaedra image) and field $B$ (density field), as visualized in Figure~\ref{fig:main figure} and explored further in Figure~\ref{fig: token swap visualization}.
\begin{itemize}
    \item \textbf{Amplitude Swap:} When we inject field $B$'s amplitude tokens into field $A$'s morphological tokens, the resulting reconstruction's energy correlation to the original field $A$ drops to $-0.09$, while its correlation to field $B$ jumps to $0.99$.
    \item \textbf{Morphology Swap:} When we swap the morphological tokens instead, the reconstructed field's energy remains firmly anchored to field $A$, maintaining a strong correlation of $r=0.97$.
\end{itemize}
This demonstrates that the macroscopic energy of the field follows the 1D amplitude tokens almost exactly, leaving the 8D tokens to dictate purely structural features.

\subsection{Spectral Ablation}
We further analyzed the disentanglement in the spectral domain. By independently zeroing out each token embedding group prior to decoding, we computed the radial power spectra of the resulting fields to measure frequency retention compared to the unmodified reconstructions.

\begin{table}[h]
\centering
\caption{Frequency retention after targeted latent space ablation.}
\label{tab:spectral_ablation}
\small
\begin{tabular}{lcc}
\toprule
\textbf{Ablation} & \textbf{Low-freq retention} & \textbf{High-freq retention}\\
\midrule
Zero Amplitude    & $3$--$5\%$    & ${\sim}300\%$ (artifacts) \\
Zero Morphological & $103$--$106\%$ & $47$--$49\%$  \\
\bottomrule
\end{tabular}
\end{table}

As shown in Table~\ref{tab:spectral_ablation}, zeroing the amplitude tokens destroys the large-scale energy of the field. "Low-frequency" refers to the frequencies at or below the bottleneck resolution (32), while "high-frequency" refers to frequencies above this value. Frequency retention is calculated as a ratio of the integrals of the radial power spectra. The massive spike in high-frequency energy represents non-physical artifacts generated by the decoder when deprived of its base amplitude conditioning. Conversely, zeroing the morphological tokens preserves the low-frequency energy ($103$--$106\%$) but roughly halves the high-frequency structure. This matches the predicted SGQ asymmetry, confirming the 8D tokens govern local high-frequency morphology.

\subsection{Fourier Filter Baseline Comparisons}
Finally, we compared our learned disentanglement against classical signal processing baselines. We isolated the low-pass and high-pass components of the true field using standard Fast Fourier Transform (FFT) filters and measured the Pearson correlation ($r$) between these ground truths and our isolated token reconstructions:
\begin{itemize}
    \item \textbf{Amplitude-only Reconstructions:} Correlate strongly ($r=0.93$) with the FFT low-pass filtered true field, and exhibit near-zero correlation with the high-pass filtered field.
    \item \textbf{Morphology-only Reconstructions:} Correlate moderately ($r \approx 0.50$) with the FFT high-pass filtered field, while showing near-zero correlation with the low-pass field.
\end{itemize}
These correlations quantitatively support the visual evidence, proving the 1D tokens act as a learned low-pass amplitude envelope while the 8D tokens capture high-pass geometric details.

\clearpage
\section{Experimental Setup}
\label{sec: Experimental Setup}
\subsection{Model Architectures}
\label{sec: Model Architectures}
All models in this work utilize a shared convolutional autoencoder backbone based on the IBQ architecture \citep{shi2025scalableimagetokenizationindex}. The encoder consists of a series of residual blocks and down-sampling layers determined by the channel multipliers (see Table~\ref{tab:model_variants}). For an input resolution of $128\times128$, we employ a base channel count of 128. We fix the codebook embedding dimension, $d$, across all models to be 8. Initial experiments indicated that larger codebook dimensions degraded performance for both the IBQ and VQ-VAE-2 tokenizers. Every tokenizer in the headline comparison therefore shares an identical encoder, decoder, and training recipe at 97M parameters, and differs \emph{only} in the factorization $\Phi$, the quantizer $Q$, and the recombination $\Psi$ of Eq.~\ref{eq:dis1}; the exceptions are noted in Table~\ref{tab:model_variants} and discussed in \ref{sec:training and optimizer details}.

\begin{figure}[t]
    \centering
    \includegraphics[width=\linewidth]{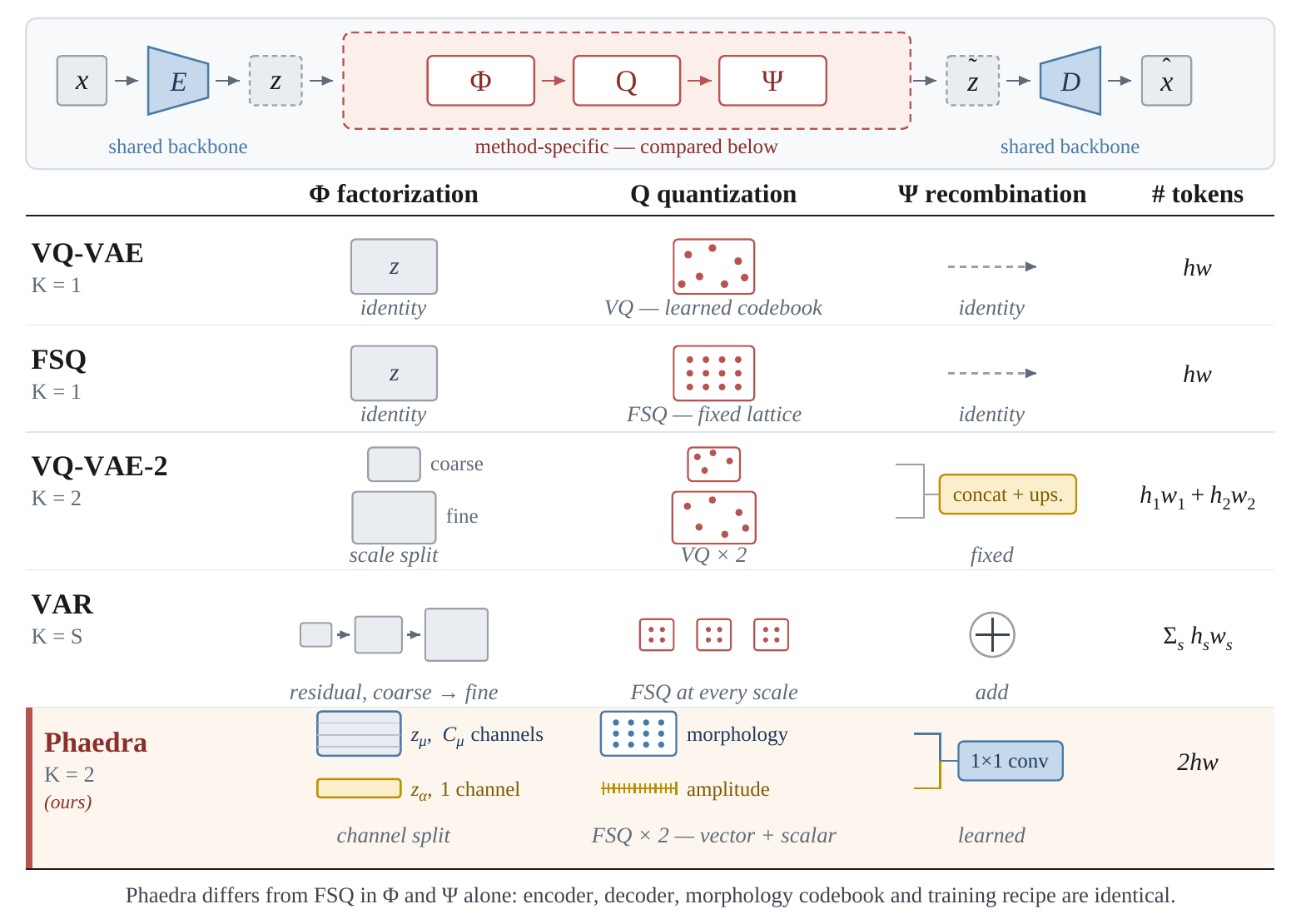}
    \caption{\textbf{A unified view of discrete tokenization.} All methods share the
    encoder $\mathcal{E}$ and decoder $\mathcal{D}$, and differ only in the factorization
    $\Phi$, the quantizers $\{Q_k\}$, and the recombination $\Psi$ of Eq.~\ref{eq:dis1}.
    Prior tokenizers either do not factorize at all (VQ-VAE, FSQ) or factorize across
    \emph{scale} (VQ-VAE-2, VAR); Phaedra factorizes across \emph{channel}, into a
    vector-quantized morphology stream and a densely quantized scalar amplitude stream,
    recombined by a learned channel mixing. Entries correspond to
    Table~\ref{tab:tokenizers}.}
    \label{fig:taxonomy}
\end{figure}

\textbf{Encoder/Decoder Details.} The downsampling factor $f$ is determined by the depth of the encoder. At the bottleneck resolution, we apply a single attention layer if the resolution is $\leq32\times32$ to capture global spatial dependencies. The decoder mirrors the encoder structure, using nearest-neighbor upsampling to recover the original resolution.
 
\paragraph{Phaedra Implementation (Ours).} In our proposed architecture, the latent space is factorized into two distinct semantic channels. For a given patch, the model predicts two independent tokens:
\begin{enumerate}
\item \textbf{Morphology Channel:} This channel captures the structural ``shape'' of the local field. It is quantized using Finite Scalar Quantization (FSQ) \citep{mentzer2023finite} with dimension $d=8$, levels $L_{morph}=[5,4,4,3,3,3,2,2]$, and a scale factor of 10. This yields a vocabulary of 8,640 shapes.
\item \textbf{Amplitude Channel:} This channel captures the intensity or energy of the field. It is a scalar value quantized via a fine-grained 1-dimensional FSQ codebook with 1,024 levels and a scale factor of 0.1.
\end{enumerate}
We stress that the morphology channel is configured \emph{identically} to the FSQ baseline below---same dimension, same levels, same scale factor. The Phaedra--FSQ comparison is consequently an exact ablation of our contribution: the two models share encoder, decoder, codebook, optimizer, schedule, and training data, and differ only in the presence of the amplitude stream and the learned recombination.

\paragraph{Quantization Baselines.} All baselines below reuse the encoder, decoder, and training recipe described above, so that the quantization stage is the only source of variation.
\begin{itemize}
\item \textbf{Continuous AE.} The backbone with the quantizer removed entirely, passing $z$ directly to the decoder. It carries no discrete bottleneck and therefore serves as the unquantized reference against which the cost of quantization is measured (\ref{sec: Error Analysis of quantization}).
\item \textbf{FSQ} \citep{mentzer2023finite}. The single-stream quantizer of Section~\ref{sec: methods} applied to the full latent, with the same configuration as the Phaedra morphology channel ($d=8$, levels $[5,4,4,3,3,3,2,2]$). Because FSQ projects onto a fixed lattice rather than a learned codebook, it has no codebook parameters and cannot suffer collapse in the usual sense.
\item \textbf{IBQ} \citep{shi2025scalableimagetokenizationindex}. Index Backpropagation Quantization replaces the standard nearest-neighbor lookup with a differentiable selection over the codebook, so that gradients reach \emph{all} codebook entries at each step rather than only the selected one. This removes the partial-update pathology that limits vector quantization and allows very large codebooks to be trained at high utilization without collapse; we use a codebook of 16,384 entries. We note that IBQ is designed for codebooks substantially larger and higher-dimensional than the $d=8$ setting we impose for parity across models, and our results should be read as characterizing IBQ under that constraint rather than at its intended operating point.
\item \textbf{VQ-VAE-2} \citep{razavi2019generating}. A hierarchical tokenizer that builds two embedding sets at two spatial resolutions, a coarse map capturing global layout and a finer map capturing local detail, each quantized independently by standard vector quantization \citep{van2017neural} and recombined by concatenation and upsampling. It is the closest baseline to Phaedra in that it also factorizes before quantizing, but the factorization is across \emph{scale} rather than across amplitude and morphology, and the quantizer is learned rather than a fixed lattice.
\item \textbf{Cosmos DI} \citep{nvidia2025cosmosworldfoundationmodel}. The pretrained NVIDIA discrete image tokenizer, evaluated without retraining at 79M parameters and its native $8\times8$ and $16\times16$ compression rates. As it expects inputs in $[0,1]$, we evaluate it under a best-case per-sample normalization (Cosmos$^{0:1}$); it is included as a representative of general-purpose image tokenizers applied off the shelf to scientific fields.
\end{itemize}
 
\paragraph{Visual Autoregression (VAR) Implementation.} Our VAR implementation reconstructs embeddings through a multi-scale hierarchy rather than a single flat raster. The image is quantized into a sequence of maps at progressively higher resolutions. This process operates as a spatial residual vector quantization: the model first predicts a coarse low-resolution map, and subsequent higher-resolution maps predict the \textit{residual} error between the current reconstruction and the true embedding. We quantize each residual level using FSQ with the standard configuration ($d=8$, levels $[5,4,4,3,3,3,2,2]$). It should be noted, the original VAR implementation used VQ-VAE; however, our initial experiments found FSQ consistently provided more accurate reconstructions. Therefore, we made the decision to use this implementation for all benchmarks. We investigate two hierarchy configurations:
\begin{enumerate}
\item \textbf{Small:} A geometric progression of resolutions: [1,2,3,4,5,6,8,10,13,16,32].
\item \textbf{Large:} The standard hierarchy with an additional refinement step at the highest resolution: [...,16,32,32].
\end{enumerate}
 
\paragraph{Architecture Summary.} Table~\ref{tab:model_variants} summarizes the hyperparameter configurations. Note that the Token Count represents the total sequence length (number of discrete codes) the generative model must predict for a single $128\times128$ input. For Phaedra, this count is doubled relative to a standard FSQ model at the same resolution, as every spatial location requires both a Morphology token and an Amplitude token; it remains below VAR$_{large}$ (2,728) and above VQ-VAE-2 (1,280). That the gains are not merely a consequence of the larger budget is addressed by the codebook ablation of Table~\ref{tab: results 4x4}, where replacing the 1-D amplitude lattice with a 5-D codebook of \emph{equal cardinality and identical token count} degrades reconstruction, and by the amplitude-codebook scaling of Figure~\ref{fig:Scaling wrt amplitude codebook size}, where Phaedra still outperforms FSQ at $L_\alpha=32$.
 
\subsection{Training and Optimizer Details}
\label{sec:training and optimizer details}
All models are trained using the AdEMAMix optimizer \citep{pagliardini2024ademamixoptimizerbetterfaster} with separate EMA parameter tracking. The learning rate is held constant throughout training, and the model is trained for 1 epoch. We consider two training scenarios:
\begin{enumerate}
    \item \textbf{High-resolution Training}: This scenario is considered for the Phaedra training runs with downsampling rates of $8\times8$ and $16\times16$ to provide fair comparisons with Cosmos DI$8\times8$ and DI$16\times16$. Inputs in this setting have a spatial resolution of $512\times512$.
    \item \textbf{Low-resolution Training:} This scenario is considered for all other runs, i.e. those which use downsampling rates of $4\times4$. Inputs in this setting have a spatial resolution of $128\times 128$.
\end{enumerate}
 
\paragraph{Parameter Matching and Compute Cost.} Because the backbone is shared and only the quantization stage differs, both parameter counts and compute costs are near-identical across tokenizers. Table~\ref{tab:model_variants} reports 97M parameters for the continuous autoencoder, FSQ, IBQ, VQ-VAE-2, VAR$_{small}$, and Phaedra $4\times4$, so the headline comparison of Table~\ref{tab: results 4x4} is parameter-matched; the exceptions are VAR$_{large}$ (112M, from its additional hierarchy level), the pretrained Cosmos models (79M), and the higher-compression Phaedra $8\times8$ and $16\times16$ variants (127M and 149M), which require additional downsampling stages and are only ever compared against Cosmos at matched compression. Phaedra's own overhead relative to FSQ is a single latent channel and one channel-mixing convolution, negligible against the encoder and decoder. Each low-resolution model trains in approximately 20 hours on 4 RTX 4090 GPUs in parallel, i.e.\ 80 GPU-hours, although all models converge around 8 hours into training; high-resolution runs require approximately 160 GPU-hours. Encoding and decoding add negligible latency relative to any downstream model, and the corresponding costs for the downstream transformers and operator baselines are reported separately in \ref{sec: downstream task results}.
The improvements reported in Tables~\ref{tab: results 4x4}--\ref{tab: cosmos comparison} therefore cannot be attributed to differences in model capacity or training budget.

\begin{table}[h]
    \centering
    \caption{\textbf{Model \& quantizer configurations for the different models.} The \textbf{Codebook Size} is the number of learnable (for IBQ, VQVAE2) or product of the levels (for FSQ) codes in the codebook. All models accept $128\times128$ inputs. The \textbf{Token Count$_{128}$} is the number of tokens, assuming a $128\times128$ input. }
    
    \begin{tabular}{l c c c c c c}
    \toprule
       \textbf{Model} & \textbf{Parameters} & \textbf{Channel Mult.} & \textbf{Codebook Size} & \textbf{Token Count$_{128}$} \\
       \midrule
       \midrule
       Continuous AE  & 97M & $[2, 2, 4]$ & - & - \\
       FSQ & 97M & $[2, 2, 4]$  & 8640 & 1024 \\
       IBQ & 97M & $[2, 2, 4]$  & 16384 & 1024 \\
       VQ-VAE2 & 97M & $[2, 2, 4]$  & 20480 & 1280 \\
       VAR$_{small}$ & 97M  & $[2, 2, 4]$  & 8640 & 1704 \\
       VAR$_{large}$ & 112M & $[2, 2, 4]$ & 8640 & 2728 \\
       Cosmos DI$8\times8$ & 79M & - & 64000 & 256 \\
       Cosmos DI$16\times16$ & 79M & -  & 64000 & 64 \\
       Phaedra $4\times4$ & 97M & $[2, 2, 4]$  & 9664 & 2048 \\
       Phaedra $8\times8$ & 127M & $[1, 2, 4, 4]$  & 9664 & 512 \\
       Phaedra $16\times16$ & 149M & $[1, 2, 2, 4, 4]$  & 9664 & 128 \\
       Phaedra $4\times4$ Codebook & 97M & $[2, 2, 4]$  & 9664 & 2048 \\
       Phaedra $4\times4$ Residual & 97M & $[2, 2, 4]$  & 9664 & 2048 \\
       \bottomrule
    \end{tabular}
    \label{tab:model_variants}
\end{table}

\begin{table}[h] 
\centering 
\caption{\textbf{Training hyperparameters across all models.} We fix the autoencoder structure across all models, unless otherwise specified. All models are trained with the following hyperparameters.}
\begin{tabular}{ll} 
\toprule 
\textbf{Parameter} & \textbf{Value} \\
\midrule 
Optimizer & AdEMAMix \\
Base Learning Rate & $10^{-4}$ \\
Optimizer Betas & ($\beta_1$=0.5,$\beta_2$=0.9,$\beta_3$=0.99) \\
AdEMAMix $\alpha$ & 2.0 \\
Weight Decay & 0.01 \\
EMA Decay & 0.999 \\
Batch Size (Low-res.) & $8\times2=16$ \\
Batch Size (High-res.) & $1\times8=8$ \\
Warmup Steps & 250 \\
Epochs & 1 (Fixed LR) \\
Training Samples & 4.8M \\
Validation Samples & 50k \\
Test Samples & 100k \\
\bottomrule 
\end{tabular} 
\end{table}

\clearpage
\subsection{Datasets}
\label{sec: datasets}
\subsubsection{Training Data Details}
\label{sec: training data}
\paragraph{Physical Regimes \& Governing Equations.}
We train and evaluate on the \emph{Poseidon} pretraining suite, using the data provided by \citep{herde2024poseidonefficientfoundationmodels}. Full details regarding the dynamics, initial conditions, and construction of this dataset are available in the original work; however, we provide a brief description below.
The dataset consists of trajectories governed by two fundamental systems of conservation laws. All simulations are defined on the space-time domain $[0,1]^2\times[0,1]$.

\paragraph{1. Compressible Euler (CEU) Equations (PDEGym \cite{herde2024poseidonefficientfoundationmodels}).} The CEU datasets describe the dynamics of a compressible gas. The system in conservation form is given by:
\begin{equation}
    \partial_t \mathbf{u} + \nabla \cdot \mathbf{F} = 0, \quad \mathbf{u} = [\rho, \rho v_x, \rho v_y, E]^T,
\end{equation}
where the flux function $\mathbf{F}$ is defined as 
\begin{equation}
    \mathbf{F} = [\rho \mathbf{v}, \rho \mathbf{v} \otimes \mathbf{v} + p \mathbf{I}, (E+p)\mathbf{v}]^T.
\end{equation}
The system is closed using the ideal gas equation of state with $\gamma=1.4$, 
\begin{equation}
    E = \frac{p}{\gamma -1} + \frac{1}{2}\rho |\mathbf{v}|^2.
\end{equation}
The CE trajectories were simulated using the \textbf{ALSVINN} code \citep{lye2020alsvinn}, which employs a high-resolution finite volume scheme featuring piecewise quadrative WENO reconstructions and HLLC Riemann solvers. 

\cite{herde2024poseidonefficientfoundationmodels} provide two versions of this dataset:
\begin{itemize}
    \item The original dataset at $512\times 512$ resolution with variables stored as $[\rho, \rho v_x, \rho v_y, E]$,
    \item and a post-processed version, downsampled to $128 \times 128$, converting momentum to velocity and energy to pressure, yielding $[\rho, v_x, v_y, p]$ as variables. 
\end{itemize}

\paragraph{2. Incompressible Navier-Stokes (INS) Equations (PDEGym \cite{herde2024poseidonefficientfoundationmodels}).} The NS datasets represent fluid flow where the density is constant (set to unity). The velocity field $\textbf{v}$ and pressure $p$ are governed by
\begin{equation}
    \partial_t \textbf{v} + (\textbf{v} \cdot \nabla ) \textbf{v} + \nabla p = \nu \Delta \textbf{v}, \quad \nabla \cdot \textbf{v} = 0.
\end{equation}
To approximate the inviscid limit while maintaining numerical stability, a small viscosity $\nu \approx 4 \times 10^{-4}$ is applied to high-frequency Fourier modes. These simulations were performed by the authors of \citep{herde2024poseidonefficientfoundationmodels} using the \textbf{AZEBAN} spectral hyperviscosity solver \citep{Rohner_2024}. The solver utilizes a Fourier projection operator $P_N$ and a resolution-dependent viscosity $\epsilon_N $ to dampen problematic higher frequencies without affecting low-frequency dynamics. 

\paragraph{Initial Condition Classes.} The training suite contains 6 distinct datasets defined by unique classes of Initial Conditions (ICs). These are:
\begin{itemize}
    \item \textbf{CEU Gauss:} Initialized with a superposition of 100 Gaussians in the vorticity field, with a constant density $(\rho=0.1)$ and pressure $p=2.5$.
    \item \textbf{CEU KH:} The Kelvin-Helmholtz instability, initialized as a shear flow with a perturbed interface to induce turbulent mixing.
    \item \textbf{CEU RC:} A multi-partitioned version of the Riemann problem using sinusoidal coordinate transformations to create curved interfaces between subdomains.  
    \item \textbf{CEU Riemann:} The four-quadrant Riemann problem where the domain is divided into four quadrants, each assigned random constant states for $\rho, v_x, v_y,$ and $p$. 
    \item \textbf{INS Gauss:} Incompressible flows initialized with a superposition of 100 Gaussians in the vorticity field $\omega=\nabla \times \mathbf{v}$.
    \item \textbf{INS Sines:} The initial velocity field is a linear combination of 10 sine and cosine modes with random coefficients.
\end{itemize}

\paragraph{Data Processing and Channel Independence.} To enable a unified tokenizer architecture capable of handling diverse physical variables, we process each physical channel independently. Consequently, the model input and output dimension is always $C=1$. For example, a dataset with 4 variables $(\rho,u,v,p)$, a single trajectory of length $T$ is treated as 4 distinct sequences of length $T$.

\paragraph{Data Splits and Volume.} The training corpus consists of 10,000 trajectories for each CEU dataset and 20,000 trajectories for each INS dataset. Each trajectory spans 21 distinct time steps. This results in a total training corpus of approximately $4.9\times10^6$ single-channel samples. The details are organized in Table \ref{tab:training data splits}.

\paragraph{Experimental Configurations.} We conduct experiments across two spatial resolutions to evaluate the scalability of our tokenizer and its performance under varying levels of token compression.

\begin{itemize} 
\item \textbf{Standard Resolution (128×128):} This serves as our primary benchmark. Data is processed using a $4^2$ downsampling factor (patch size $4\times4$), reducing the spatial dimensions to a $32\times 32$ latent grid. For the Compressible Euler (CE) datasets, we utilize primitive variables $(\rho, v_x, v_y, p)$. Training is performed with a global batch size of 16 (2 mini-batches, batch size 8 per device).
\item \textbf{High Resolution (512×512):} To verify the model's robustness to extreme token compression and to facilitate comparisons with existing pretrained tokenizers, we utilize high-fidelity $512^2$ snapshots. In this regime, we test more aggressive downsampling factors of $8^2$ and $16^2$, resulting in latent grids of $64 \times 64$ and $32 \times 32$, respectively. For these experiments, the CE equations are represented via \textbf{conservative variables}: density $(\rho)$, momentum $(\rho v_x, \rho v_y)$, and total energy ($E$). Training utilizes a global batch size of 8 (8 mini-batches, batch size 1 per device).
\end{itemize}

\begin{table}[h]
    \centering
    \caption{Summary of the scientific datasets used for training. ``Vars" indicates the number of physical fields present in the simulation. Note that the standard ($128^2$) and high-res ($512^2$) runs use different variable bases for the Compressible Euler datasets (Primitive vs. Conservative).}
    \begin{tabularx}{\textwidth}{l X c c c}
    \toprule
     \textbf{Dataset Name} & \textbf{Initial Conditions} & \textbf{Vars} & \textbf{Train/Val/Test} & \textbf{Steps/Traj} \\
    \midrule
         CEU Gauss    & Gaussian density perturbations & 4 & 9{,}640/120/240 & 21 \\
         CEU KH       & Kelvin--Helmholtz instability  & 4 & 9{,}640/120/240 & 21 \\
         CEU RC   & Curved interface Riemann       & 4 & 9{,}640/120/240 & 21 \\
         CEU Riemann & 4-Quadrant Riemann interaction & 4 & 9{,}640/120/240 & 21 \\
    \midrule
        INS Gauss & Gaussian vortex field    & 2 & 19{,}640/120/240  & 21 \\
        INS Sine  & Sinusoidal perturbations & 2 & 19{,}640/120/240 & 21 \\
    \bottomrule
    \end{tabularx}
    \label{tab:training data splits}
\end{table}

\clearpage
\subsubsection{Out-of-Distribution PDE Datasets}
\label{sec: out of distribution data}
To evaluate the zero-shot generalization and robustness of our model, we consider two distinct categories of out-of-distribution (OD) benchmarks. In all cases, the model is applied directly to the test data (240 trajectories or 256 samples) without any parameter updates or fine-tuning on the target distributions.

\paragraph{OD$_1$: Distributional Shifts in Initial Conditions.}

This category utilizes the same underlying Partial Differential Equations (PDEs) encountered during training, i.e. Compressible Euler (CE) and Incompressible Navier-Stokes (INS), but introduces significantly different Initial Conditions (ICs) or boundary geometries.
\begin{itemize}
    \item \textbf{CE Riemann Kelvin-Helmholtz (RKH)} \citep{herde2024poseidonefficientfoundationmodels}. This benchmark combines the shock-dominated dynamics of a Riemann problem with uncertain interfaces that trigger Kelvin-Helmholtz instabilities. We evaluate the model's ability to reconstruct the \textbf{density} field $(\rho)$ under these complex, combined physical phenomena.
    \item \textbf{INS Sinusoidal Vortex Sheet (SVS)} \citep{herde2024poseidonefficientfoundationmodels}. Also from the Poseidon suite, this is a stochastic version of a classic Navier-Stokes benchmark. The initial condition is defined by a vorticity field $\omega_0$ concentrated along a perturbed interface. We reconstruct the \textbf{horizontal velocity} $(v_x)$. The problem involves thin shear layers that are significantly more localized than the Gaussian vortices seen during training. 
    \item \textbf{CE Airfoil} \citep{raonic2023convolutionalneuraloperatorsrobust}. This task involves the steady-state solution of the compressible Euler equations around an RAE2822 airfoil. The geometry is perturbed using 30 Hicks-Henne bump functions. We reconstruct the steady-state \textbf{density} field $(\rho)$ on a body-fitted grid, testing the model's ability to generalize to non-periodic boundary conditions and specific aerodynamic shapes.
\end{itemize}

\paragraph{OD$_2$: Generalization to Unseen Physics.}
This category tests the model's capacity for ``physics-agnostic" feature extraction by applying it to PDEs that were entirely absent from the training corpus. For these tasks, we reconstruct the primary solution variable $u$.
\begin{itemize}
    \item \textbf{Poisson Equation (POI)} \citep{raonic2023convolutionalneuraloperatorsrobust}. This dataset is described by a linear elliptic PDE given by $- \Delta u =f$ in the domain $\mathcal{D}$, with $u|_\mathcal{\partial D} = 0$. The source term $f$ is a multiscale function,
\begin{equation}
    f(x,y)=\sum_{i,j=1}^K a_{ij} \cdot (i^2 + j^2)^{-r } \sin(\pi i x) \sin (\pi j y),
\end{equation}
with $K=20$.

\item \textbf{Darcy Flow (DAR)} \citep{raonic2023convolutionalneuraloperatorsrobust}. A second-order linear elliptic PDE $- \nabla \cdot (a \nabla u) = f$ representing flow through a porous medium. We evaluate the reconstruction of the pressure field $u$. 

\item \textbf{Allen-Cahn Equation (ALC)} \citep{raonic2023convolutionalneuraloperatorsrobust}. The Allen-Cahn equation is a prototype for nonlinear parabolic PDEs, typically used to model phase separation in multi-component alloy systems. The equation is given by $\partial_t u = \Delta u - \epsilon^2 u (u^2 -1)$. The reaction rate is set to $\epsilon=220$. Our objective is to reconstruct the solution field $u$ at the final time $T=0.0002$.

\item \textbf{Acoustic Wave Layer (AWA)} \citep{herde2024poseidonefficientfoundationmodels}. We further test the tokenizer on a second-order hyperbolic equation modeling acoustic wave propagation. This dataset is described by the wave equation $\partial_{tt} u = c^2 \Delta u$. In this benchmark, the propagation speed $c(x,y)$ is spatially dependent and defined by a vertically layered medium. Each instance contains $n\in \{3,4,5,6\}$ layers with constant speeds $c_i \sim \mathcal{U}[2000,5000]$. The interfaces between layers are randomized via x-dependent frontiers $a_i(x)$. 
This task represents a significant shift in the underlying physics. The tokenizer must reconstruct complex wave patterns involving reflection and refraction at discontinuous interfaces—dynamics that are absent from the smooth or shock-based fluid datasets used during training. We specifically evaluate the reconstruction of the solution snapshot at $t=0.7$.
\end{itemize}

\clearpage
\subsubsection{Earth Observation}
\label{sec: earth observation data}

We leverage a range of diverse data products from Earth observation for the out-of-distribution evaluation of Phaedra. Specifically, we include data with vastly different characteristics: (i)~multispectral optical data from the Sentinel-2 mission, (ii)~radar data from the Sentinel-1 mission, (iii)~digital elevation maps, and (iv)~vegetation data. Data is gathered from the TerraMesh dataset \cite{blumenstiel2025terramesh}. 

\begin{itemize}
    \item \textbf{Multi-spectral optical data}: \textbf{Sentinel‑2} is part of the Copernicus Programme of the European Union. The mission consists of two identical polar‑orbiting satellites (Sentinel‑2A and ‑2B) providing systematic, global observations of the Earth’s land surface, coastal zones, and inland waters. The mission carries a MultiSpectral Instrument (MSI) acquiring imagery in 13 spectral bands ranging from the visible and near‑infrared to short‑wave infrared, with spatial resolutions of 10 meter, 20 meter, and 60 meter depending on the band. The primary data product is Level‑1C (\textbf{L1C}) which contains orthorectified top‑of‑atmosphere (TOA) reflectances in a global cartographic geometry, including radiometric calibration and geometric corrections but no atmospheric correction. The Level‑2A (\textbf{L2A}) product provides bottom‑of‑atmosphere (BOA) surface reflectances, generated from L1C through atmospheric correction, and includes additional scene classification layers such as clouds, cloud shadows, water, snow, and vegetation. Sentinel‑2 \textbf{RGB} imagery is derived by combining the red (Band 4), green (Band 3), and blue (Band 2) bands at 10 m resolution to produce natural‑color composites that resemble human vision.

    \item \textbf{Radar data}: Similar to Sentinel-2, the \textbf{Sentinel‑1} mission is equally part of the EU Copernicus Programme. The radar mission comprises two C‑band synthetic aperture radar (SAR) satellites (Sentinel‑1A and ‑1B) designed for all‑weather, day‑and‑night Earth observation. Sentinel‑1 operates primarily in Interferometric Wide Swath (IW) mode over land, acquiring dual‑polarization data (VV+VH or HH+HV) with a ground resolution of approximately 10 meter. The Radiometrically Terrain‑Corrected (\textbf{RTC}) product builds upon Ground Range Detected (GRD) data by applying terrain correction using a digital elevation model and normalizing the backscatter to a reference geometry, yielding representations that are comparable across incidence angles and terrain conditions.

    \item \textbf{Digital elevation}: \textbf{Digital Elevation Models (DEM)} provide static topographic information describing the elevation of the Earth’s surface and often serve as a geophysical context for optical and radar observations. We use a global DEM product derived from the Copernicus DEM, which offers near‑global coverage at approximately 10 meter spatial resolution. The DEM is resampled and co‑registered to the Sentinel‑1 and Sentinel‑2 grids, enabling pixel‑wise alignment across modalities. Elevation values capture terrain characteristics such as slope, aspect, and relief, which influence surface reflectance and radar backscatter. 

    \item \textbf{Vegetation data}: Vegetation is represented using the \textbf{Normalized Difference Vegetation Index (NDVI)}, a widely used spectral index quantifying vegetation greenness and photosynthetic activity. NDVI is computed from the red and near‑infrared bands of optical satellite observations and provides a normalized measure of vegetation density that is largely insensitive to absolute illumination conditions. The NDVI data is temporally aligned with the corresponding satellite acquisitions and resampled to a 10 meter grid, ensuring spatial consistency with optical and radar modalities. 
\end{itemize}

\subsubsection{ERA-5 Reanalyses Data}
\label{sec: era5 reanalysis data}

Atmospheric state variables are represented using \textbf{ERA5 reanalyses} as provided by the WeatherBench~2 benchmark \cite{rasp2024weatherbench}. ERA5 is a global atmospheric reanalysis produced by the European Centre for Medium‑Range Weather Forecasts (ECMWF), combining numerical weather prediction models with a wide range of observations to generate temporally consistent datasets. We use data from the year 2020 at a spatial resolution of {240$
\times$121}, corresponding to a regular latitude–longitude grid. Our analysis primarily focuses on {surface temperature} as well as the {zonal (U) and meridional (V) wind components} across multiple atmospheric pressure levels, capturing both near‑surface and upper‑air dynamics.

\subsubsection{ImageNet}
\label{sec: imagenet1k data description}
ILSVRC 2012 \cite{imagenet15russakovsky}, also referred to as \emph{ImageNet}, is a dataset comprising labeled images. It organizes images according to meaningful concepts from the WordNet, with approximately 1000 images for each set of synonymous concepts or ``synset`` from WordNet. 

\clearpage
\section{Numerical Results: Evaluation in Physical Space}
\subsection{ID Evaluation in Physical Space}
\label{sec: results evaluation in physical space}
In this section, we evaluate the reconstruction fidelity of the models within the physical domain. To ensure comparability across disparate physical variables, such as density $(\rho)$, velocity $(u,v)$, and pressure $(p)$, all errors are normalized by the global standard deviation $\sigma_g$ of the respective variable calculated over the entire training dataset. This normalization also prevents unreasonably large errors which arise when using relative metrics for datasets which are initialized as zero.

We employ three primary metrics to characterize the spatial error distribution:

\begin{itemize} 
\item \textbf{Normalized Mean Absolute Error (nMAE):} Measures the average magnitude of the residuals, providing a robust estimate of global reconstruction accuracy, provided in Table \ref{tab: nMAE low res}.
$$\text{nMAE} = \frac{1}{\sigma_g N} \sum_{i=1}^{N} |y_i - \hat{y}_i|$$

\item \textbf{Normalized Root Mean Squared Error (nRMSE):} Quantifies the root average squared deviation, placing a higher penalty on larger outliers, provided in Table \ref{tab: nRMSE low res}.
$\text{nRMSE} = \frac{1}{\sigma_g} \left( \frac{1}{N} \sum_{i=1}^{N} (y_i - \hat{y}_i)^2 \right)^{\frac{1}{2}}$

\item \textbf{Normalized $L_\infty$ Error (n$L_\infty$):} Represents the worst-case standardized reconstruction error across the spatial grid, critical for identifying local artifacts or gradient instabilities, provided in Table \ref{tab: nLinf low res}.
$$\text{n}L_\infty = \frac{1}{\sigma_g} \max |y_i - \hat{y}_i|$$

\item \textbf{Relative $L_1$ Error (r$L_1$):} 
$$rL_1 = \frac{\sum_{i=1}^N |y_i-\hat{y}_i|}{\sum_{i=1}^N |y_i|} $$

\item \textbf{Relative $L_2$ Error (r$L_2$):} 
$$rL_2 = \frac{\sqrt{\sum_{i=1}^N |y_i-\hat{y}_i|^2}}{\sqrt{\sum_{i=1}^N y_i^2}} $$

\item \textbf{Local Variance Error ($\Delta \sigma^2_{loc}$):} This metric identifies localized blurring or ``blocky" artifacts by comparing the spatial distribution of variance within a sliding window $\mathcal{W}$ of size $7 \times 7$.
$$\sigma^2(y) = \mathbb{E}[y^2]_{\mathcal{W}} - (\mathbb{E}[y]_{\mathcal{W}})^2, \quad \Delta \sigma^2_{loc} = \frac{\max |\sigma^2(y) - \sigma^2(\hat{y})|}{\max|\sigma^2(y)|} \times100\%$$
Local variance error quantifies the model's ability to maintain high-frequency fidelity. High values indicate that the model is either over-smoothing turbulent regions (loss of local variance) or introducing artificial high-frequency noise that does not exist in the ground truth.
\end{itemize}

\newpage
\begin{table}[h!]
\centering
\small
\caption{Comparison of nMAE$\downarrow$ across all $4\times4$ models, training datasets, and variables.}
\label{tab: nMAE low res}
\begin{tabular}{llcccc}
\toprule
Model & Dataset & $\rho$ & $u$ & $v$ & $p$ \\
\midrule
\multirow{6}{*}{Continuous AE} & CEU Gauss & 0.4413 & 0.3192 & 0.2958 & 0.3928 \\
 & CEU KH & 0.6495 & 0.3481 & 0.4031 & 0.5411 \\
 & CEU RC & 2.3518 & 1.5309 & 1.5208 & 1.4503 \\
 & CEU Riemann & 0.4697 & 0.6683 & 0.5064 & 0.3952 \\
 & INS Gauss & --- & 0.2655 & 0.2679 & --- \\
 & INS Sines & --- & 0.2888 & 0.3274 & --- \\
\midrule
\midrule
\multirow{6}{*}{FSQ} & CEU Gauss & 1.5842 & 1.1574 & 1.1879 & 1.7706 \\
 & CEU KH & 2.4296 & 1.0342 & 1.7243 & 3.8152 \\
 & CEU RC & 8.4078 & 5.5781 & 5.5651 & 5.5038 \\
 & CEU Riemann & 2.0230 & 1.8767 & 1.8243 & 1.6389 \\
 & INS Gauss & --- & 0.7782 & 0.7710 & --- \\
 & INS Sines & --- & 1.6885 & 1.7024 & --- \\
\midrule
\multirow{6}{*}{IBQ} & CEU Gauss & 5.6505 & 4.5090 & 4.6051 & 6.0485 \\
 & CEU KH & 6.4937 & 3.5596 & 6.0583 & 15.7795 \\
 & CEU RC & 20.6294 & 15.1505 & 15.0311 & 14.3478 \\
 & CEU Riemann & 7.9813 & 7.2406 & 7.1240 & 6.9787 \\
 & INS Gauss & --- & 3.1680 & 3.1375 & --- \\
 & INS Sines & --- & 8.2550 & 8.0932 & --- \\
\midrule
\multirow{6}{*}{VQ-VAE-2} & CEU Gauss  & 1.6373 & 1.1062 & 1.0255 & 1.6507 \\
 & CEU KH & 4.7605 & 1.2107 & 2.1341 & 6.3186 \\
 & CEU RC & 12.2041 & 8.4709 & 8.5027 & 8.2012 \\
 & CEU Riemann & 2.4242 & 1.8119 & 1.7743 & 1.8952 \\
 & INS Gauss & --- & 0.5333 & 0.5563  --- \\
 & INS Sines & --- & 2.0289 & 2.0057 & --- \\
\midrule
\multirow{6}{*}{VAR$_{large}$} & CEU Gauss & 2.1071 & 1.6158 & 1.6268 & 2.2636 \\
 & CEU KH & 2.7871 & 1.4450 & 2.2838 & 5.6828 \\
 & CEU RC & 9.9784 & 6.4321 & 6.3915 & 6.2769 \\
 & CEU Riemann & 2.4774 & 2.0108 & 2.0113 & 2.0608 \\
 & INS Gauss & --- & 1.1476 & 1.1400 & --- \\
 & INS Sines & --- & 2.4012 & 2.4460 & --- \\
\midrule
\multirow{6}{*}{VAR$_{small}$} & CEU Gauss & 2.6294 & 1.8349 & 1.8441 & 2.7967 \\
 & CEU KH & 3.3265 & 1.6992 & 2.9173 & 7.6570 \\
 & CEU RC & 11.7342 & 8.2315 & 8.2217 & 7.9471 \\
 & CEU Riemann & 3.1275 & 2.7895 & 2.7668 & 2.5952 \\
 & INS Gauss & --- & 1.2793 & 1.2715 & --- \\
 & INS Sines & --- & 3.2122 & 3.2374 & --- \\
\midrule
\multirow{6}{*}{Phaedra} & CEU Gauss & \textbf{0.8872} & \textbf{0.5782} & \textbf{0.5926} & \textbf{0.8919} \\
 & CEU KH & \textbf{1.4840} & \textbf{0.5870} & \textbf{0.9102} &\textbf{ 1.8920} \\
 & CEU RC & \textbf{5.5397} & \textbf{3.5207} & \textbf{3.5134} & \textbf{3.4863} \\
 & CEU Riemann & \textbf{1.1580} & \textbf{1.1195} & \textbf{1.1477} & \textbf{0.9764} \\
 & INS Gauss & --- & \textbf{0.3996} & \textbf{0.3967} & --- \\
 & INS Sines & --- & \textbf{0.6820} & \textbf{0.6861} & --- \\
\midrule
\multirow{6}{*}{Codebook Ablation} & CEU Gauss & 1.4473 & 1.1180 & 1.1355 & 1.5317 \\
 & CEU KH & 2.1341 & 0.9830 & 1.5976 & 3.4328 \\
 & CEU RC & 7.8021 & 4.9394 & 4.9293 & 4.8863 \\
 & CEU Riemann & 1.9194 & 1.7995 & 1.7681 & 1.6357 \\
 & INS Gauss & --- & 0.7614 & 0.7553 & --- \\
 & INS Sines & --- & 1.5538 & 1.5679 & --- \\
\midrule
\multirow{6}{*}{Residual Ablation} & CEU Gauss & 1.0055 & 0.5897 & 0.6010 & 1.0279 \\
 & CEU KH & 2.2374 & 0.6585 & 0.9909 & 2.3237 \\
 & CEU RC & 7.2957 & 4.2155 & 4.2065 & 4.4007 \\
 & CEU Riemann & 1.5202 & 1.1647 & 1.1392 & 1.2183 \\
 & INS Gauss & --- & 0.3506 & 0.3457 & --- \\
 & INS Sines & --- & 0.8565 & 0.8517 & --- \\
\bottomrule
\end{tabular}
\end{table}

\newpage
\begin{table}[h!]
\centering
\small
\caption{Comparison of nRMSE$\downarrow$ across all $4\times4$ models, training datasets, and variables.}
\label{tab: nRMSE low res} %Norm L1
\begin{tabular}{llcccc}
\toprule
Model & Dataset & $\rho$ & $u$ & $v$ & $p$ \\
\midrule
\multirow{6}{*}{Continuous AE} & CEU Gauss & 0.6654 & 0.4270 & 0.4450 & 0.6197 \\
 & CEU KH & 1.3745 & 0.5130 & 0.7470 & 0.8017 \\
 & CEU RC & 4.8338 & 2.5375 & 2.5300 & 2.6390 \\
 & CEU Riemann & 0.9652 & 0.9132 & 0.8846 & 0.7325 \\
 & INS Gauss & --- & 0.3753 & 0.3591 & --- \\
 & INS Sines & --- & 0.4071 & 0.4645 & --- \\
\midrule
\midrule
\multirow{6}{*}{FSQ} & CEU Gauss & 2.6164 & 1.6045 & 1.6520 & 2.9064 \\
 & CEU KH & 5.2994 & 1.6048 & 2.8778 & 5.2458 \\
 & CEU RC & 15.4156 & 8.5106 & 8.4890 & 9.4120 \\
 & CEU Riemann & 4.0658 & 3.1125 & 3.0411 & 3.2610 \\
 & INS Gauss & --- & 1.0540 & 1.0493 & --- \\
 & INS Sines & --- & 2.2983 & 2.3281 & --- \\
\midrule
\multirow{6}{*}{IBQ} & CEU Gauss & 8.9493 & 6.3105 & 6.4368 & 9.4540 \\
 & CEU KH & 15.0017 & 5.5192 & 9.5811 & 21.9039 \\
 & CEU RC & 34.7833 & 22.1880 & 22.0341 & 22.8153 \\
 & CEU Riemann & 13.5160 & 10.9087 & 10.9787 & 11.7638 \\
 & INS Gauss & --- & 4.4582 & 4.4453 & --- \\
 & INS Sines & --- & 11.3766 & 11.2976 & --- \\
\midrule
\multirow{6}{*}{VQ-VAE-2} & CEU Gauss & 2.8144 & 1.6042 & 1.6268 & 2.9209 \\
 & CEU KH & 7.2202 & 1.8412 & 3.1529 & 6.5577 \\
 & CEU RC & 19.6992 & 10.9752 & 10.9831 & 11.8936 \\
 & CEU Riemann & 4.5860 & 3.1010 & 3.1188 & 3.5560 \\
 & INS Gauss & --- & 0.7181 & 0.7307 & --- \\
 & INS Sines & --- & 2.1196 & 2.1515 & --- \\
\midrule
\multirow{6}{*}{VAR$_{large}$} & CEU Gauss & 3.1827 & 2.1603 & 2.1872 & 3.4304 \\
 & CEU KH & 5.7861 & 2.1024 & 3.5183 & 8.0035 \\
 & CEU RC & 17.8124 & 9.5782 & 9.5087 & 10.4011 \\
 & CEU Riemann & 4.7157 & 3.2230 & 3.2263 & 3.8870 \\
 & INS Gauss & --- & 1.5071 & 1.5011 & --- \\
 & INS Sines & --- & 3.2205 & 3.2899 & --- \\
\midrule
\multirow{6}{*}{VAR$_{small}$} & CEU Gauss & 3.9460 & 2.4789 & 2.4950 & 4.2456 \\
 & CEU KH & 7.0043 & 2.5221 & 4.3860 & 10.5044 \\
 & CEU RC & 20.1387 & 11.9832 & 11.9704 & 12.6990 \\
 & CEU Riemann & 5.7231 & 4.2925 & 4.2586 & 4.6633 \\
 & INS Gauss & --- & 1.6819 & 1.6748 & --- \\
 & INS Sines & --- & 4.3095 & 4.3431 & --- \\
\midrule
\multirow{6}{*}{Phaedra} & CEU Gauss & \textbf{1.3892} & \textbf{0.7938} & \textbf{0.8130} & \textbf{1.4313} \\
 & CEU KH & \textbf{3.1860} & \textbf{0.9107} & \textbf{1.4774} & \textbf{2.5807} \\
 & CEU RC & \textbf{9.9943} & \textbf{5.3645} & \textbf{5.3433} & \textbf{5.8833} \\
 & CEU Riemann & \textbf{2.2507} & \textbf{1.8456} & \textbf{1.8758} & \textbf{1.7698} \\
 & INS Gauss & --- & \textbf{0.5154} & \textbf{0.5088} & --- \\
 & INS Sines & --- & \textbf{0.9201} & \textbf{0.9273} & --- \\
\midrule
\multirow{6}{*}{Codebook Ablation} & CEU Gauss & 2.2483 & 1.5102 & 1.5367 & 2.3774 \\
 & CEU KH & 4.4492 & 1.4783 & 2.5460 & 4.7294 \\
 & CEU RC & 13.9019 & 7.3818 & 7.3576 & 8.0576 \\
 & CEU Riemann & 3.5600 & 2.7803 & 2.7529 & 2.9565 \\
 & INS Gauss & --- & 1.0098 & 1.0064 & --- \\
 & INS Sines & --- & 2.0986 & 2.1228 & --- \\
\midrule
\multirow{6}{*}{Residual Ablation} & CEU Gauss & 1.9332 & 0.9629 & 0.9811 & 2.0858 \\
 & CEU KH & 4.7696 & 1.1453 & 1.8224 & 3.3167 \\
 & CEU RC & 13.5895 & 6.6240 & 6.6109 & 7.9535 \\
 & CEU Riemann & 3.3940 & 2.3093 & 2.2799 & 2.8874 \\
 & INS Gauss & --- & 0.5822 & 0.5775 & --- \\
 & INS Sines & --- & 1.3147 & 1.3090 & --- \\
\bottomrule
\end{tabular}
\end{table}

\begin{table}[h!]
\centering
\small
\caption{Comparison of normalized maximum error (n$L_\infty \downarrow$ across all $4\times4$ models, training datasets, and variables.}
\label{tab: nLinf low res}
\begin{tabular}{llcccc}
\toprule
Model & Dataset & $\rho$ & $u$ & $v$ & $p$ \\
\midrule
\multirow{6}{*}{Continuous AE} & CEU Gauss & 10.81 & 4.94 & 5.09 & 11.11 \\
 & CEU KH & 29.07 & 6.56 & 10.57 & 7.40 \\
 & CEU RC & 80.56 & 34.85 & 34.90 & 44.92 \\
 & CEU Riemann & 22.18 & 16.26 & 16.02 & 16.22 \\
 & INS Gauss & --- & 2.00 & 2.00 & --- \\
 & INS Sines & --- & 4.22 & 4.52 & --- \\
\midrule
\midrule
\multirow{6}{*}{FSQ} & CEU Gauss & 51.23 & 18.73 & 19.39 & 56.91 \\
 & CEU KH & 74.40 & 18.59 & 37.28 & 42.85 \\
 & CEU RC & 230.92 & 103.18 & 102.19 & 173.69 \\
 & CEU Riemann & 82.38 & 49.33 & 49.00 & 80.36 \\
 & INS Gauss & --- & 7.93 & 7.91 & --- \\
 & INS Sines & --- & 21.38 & 22.15 & --- \\
\midrule
\multirow{6}{*}{IBQ} & CEU Gauss & 122.02 & 51.49 & 52.54 & 132.63 \\
 & CEU KH & 155.56 & 52.37 & 94.71 & 185.68 \\
 & CEU RC & 388.05 & 211.95 & 211.62 & 311.67 \\
 & CEU Riemann & 172.41 & 114.38 & 118.53 & 183.52 \\
 & INS Gauss & --- & 31.11 & 31.45 & --- \\
 & INS Sines & --- & 88.03 & 93.03 & --- \\
\midrule
\multirow{6}{*}{VQ-VAE-2} & CEU Gauss & 54.53 & 20.96 & 21.62 & 60.46 \\
 & CEU KH & 97.00 & 20.80 & 39.52 & 48.85 \\
 & CEU RC & 281.53 & 122.37 & 122.96 & 213.61 \\
 & CEU Riemann & 94.97 & 55.23 & 56.37 & 96.14 \\
 & INS Gauss & --- & 6.45 & 6.53 & --- \\
 & INS Sines & --- & 21.68 & 22.76 & --- \\
\midrule
\multirow{6}{*}{VAR$_{large}$} & CEU Gauss & 49.76 & 19.98 & 20.69 & 54.54 \\
 & CEU KH & 76.41 & 20.94 & 38.66 & 70.42 \\
 & CEU RC & 244.96 & 105.52 & 104.07 & 174.01 \\
 & CEU Riemann & 86.88 & 47.72 & 48.00 & 84.62 \\
 & INS Gauss & --- & 9.77 & 9.94 & --- \\
 & INS Sines & --- & 26.16 & 26.61 & --- \\
\midrule
\multirow{6}{*}{VAR$_{small}$} & CEU Gauss & 59.92 & 24.27 & 24.16 & 66.32 \\
 & CEU KH & 90.45 & 25.52 & 45.57 & 90.55 \\
 & CEU RC & 275.58 & 125.06 & 124.94 & 211.70 \\
 & CEU Riemann & 100.07 & 58.20 & 57.96 & 97.78 \\
 & INS Gauss & --- & 10.83 & 10.97 & --- \\
 & INS Sines & --- & 34.36 & 34.58 & --- \\
\midrule
\multirow{6}{*}{Phaedra} & CEU Gauss & \textbf{26.23} & \textbf{9.97} & \textbf{10.01} & \textbf{28.43} \\
 & CEU KH & \textbf{49.76} & \textbf{11.29} & \textbf{18.35} & \textbf{19.89} \\
 & CEU RC & \textbf{148.}39 & \textbf{64.86} & \textbf{63.87} & \textbf{104.}96 \\
 & CEU Riemann & \textbf{47.34} & \textbf{30.89} & \textbf{30.88} & \textbf{42.16} \\
 & INS Gauss & --- & \textbf{3.53} & \textbf{3.52} & --- \\
 & INS Sines & --- & \textbf{9.23} & \textbf{9.75} & --- \\
\midrule
\multirow{6}{*}{Codebook Ablation} & CEU Gauss & 38.74 & 15.45 & 15.96 & 42.27 \\
 & CEU KH & 61.36 & 16.10 & 31.68 & 39.95 \\
 & CEU RC & 205.03 & 86.50 & 85.31 & 143.86 \\
 & CEU Riemann & 68.39 & 39.47 & 39.61 & 66.55 \\
 & INS Gauss & --- & 7.29 & 7.36 & --- \\
 & INS Sines & --- & 18.13 & 19.03 & --- \\
\midrule
\multirow{6}{*}{Residual Ablation} & CEU Gauss & 46.27 & 17.07 & 17.26 & 52.83 \\
 & CEU KH & 66.90 & 17.79 & 27.84 & 33.59 \\
 & CEU RC & 212.96 & 85.07 & 84.47 & 155.21 \\
 & CEU Riemann & 75.64 & 43.33 & 42.55 & 77.98 \\
 & INS Gauss & --- & 10.07 & 10.19 & --- \\
 & INS Sines & --- & 20.55 & 20.86 & --- \\
\bottomrule
\end{tabular}
\end{table}

\begin{table}[h!]
\centering
\small
\caption{Comparison of local variance error $\Delta\sigma^2_{loc} \downarrow$ across all $4\times4$ models, datasets, and variables.}
\begin{tabular}{llcccc} %local var
\toprule
Model & Dataset & $\rho$ & $u$ & $v$ & $p$ \\
\midrule
\multirow{6}{*}{AE Continuous} & CEU Gauss & 7.1022 & 1.4656 & 1.4148 & 6.4110 \\ 
 & CEU KH & 2.5600 & 1.5330 & 6.2576 & 6.4077 \\
 & CEU RC & 2.4214 & 2.2839 & 2.1390 & 2.2797 \\
 & CEU Riemann & 2.0933 & 5.9525 & 3.1555 & 1.7560 \\
 & INS Gauss & --- & 1.2037 & 1.2210 & --- \\
 & INS Sines & --- & 0.9001 & 1.0462 & --- \\
\midrule
\midrule
\multirow{6}{*}{AE FSQ} & CEU Gauss & 13.1460 & 7.2716 & 7.3850 & 14.4460 \\ 
 & CEU KH & 12.6578 & 6.4351 & 12.9247 & 14.1667 \\
 & CEU RC & 20.8387 & 13.4452 & 13.9370 & 21.5758 \\
 & CEU Riemann & 12.1791 & 13.6554 & 10.5422 & 10.4616 \\
 & INS Gauss & --- & 4.8224 & 4.6141 & --- \\
 & INS Sines & --- & 5.4708 & 5.7247 & --- \\
\midrule
\multirow{6}{*}{AE IPQ} & CEU Gauss & 55.4958 & 45.3449 & 45.6992 & 57.6184 \\
 & CEU KH & 43.7444 & 28.4937 & 33.3352 & 42.5119 \\
 & CEU RC & 64.3039 & 55.9095 & 53.8193 & 71.2099 \\
 & CEU Riemann & 43.1572 & 50.8144 & 45.4075 & 42.6998 \\
 & INS Gauss & --- & 29.9442 & 26.4051 & --- \\
 & INS Sines & --- & 28.5017 & 28.6824 & --- \\
\midrule
\multirow{6}{*}{AE VQVAE2} & CEU Gauss & 16.1925 & 6.9803 & 7.4218 & 16.8609  \\ 
 & CEU KH & 26.9932 & 7.4908 & 12.3255 & 15.1002 \\
 & CEU RC & 34.9857 & 19.3438 & 19.8505 & 36.9111 \\
 & CEU Riemann & 17.5794 & 15.1483 & 13.2076 & 17.6304 \\
 & INS Gauss & --- & 3.5796 & 3.4126 & --- \\
 & INS Sines & --- & 4.6725 & 4.6263 & --- \\
\midrule
\multirow{6}{*}{VAR$_{large}$} & CEU Gauss & 14.7919 & 8.8846 & 9.0220 & 15.3468 \\ 
 & CEU KH & 13.8038 & 7.7105 & 13.7219 & 21.1367 \\
 & CEU RC & 29.2065 & 16.7669 & 16.9259 & 21.1900 \\
 & CEU Riemann & 13.5286 & 12.5818 & 10.2089 & 13.1864 \\
 & INS Gauss & --- & 6.4256 & 6.4061 & --- \\
 & INS Sines & --- & 7.3671 & 8.3635 & --- \\
\midrule
\multirow{6}{*}{VAR$_{small}$} & CEU Gauss & 17.0333 & 11.2408 & 11.6997 & 17.5699 \\ 
 & CEU KH & 17.0183 & 10.3782 & 15.7022 & 23.7020 \\
 & CEU RC & 31.8274 & 19.3497 & 19.7310 & 26.8133 \\
 & CEU Riemann & 17.2686 & 17.5919 & 13.9959 & 14.2574 \\
 & INS Gauss & --- & 8.2847 & 7.5344 & --- \\
 & INS Sines & --- & 9.7012 & 10.0351 & --- \\
\midrule
\multirow{6}{*}{Phaedra} & CEU Gauss & \textbf{8.0820} & \textbf{2.7267} & \textbf{2.9978} & \textbf{8.1773} \\ 
 & CEU KH & \textbf{7.2102} & \textbf{3.0264} & \textbf{7.5021} & \textbf{8.0339} \\
 & CEU RC & \textbf{9.5770} & \textbf{6.1343} & \textbf{6.3184} & \textbf{8.5875} \\
 & CEU Riemann & \textbf{5.5694} & \textbf{9.4949} & \textbf{5.6252} & \textbf{4.6193} \\
 & INS Gauss & --- & \textbf{4.4532} & \textbf{4.7492} & --- \\
 & INS Sines & --- & \textbf{3.3483} & \textbf{2.9431} & --- \\
\midrule
\multirow{6}{*}{Codebook Ablation} & CEU Gauss & 12.3104 & 6.5689 & 6.3608 & 12.1637 \\ 
 & CEU KH & 11.0185 & 5.2732 & 12.3157 & 14.3675 \\
 & CEU RC & 19.2576 & 11.9405 & 12.3364 & 15.6704 \\
 & CEU Riemann & 11.2302 & 10.0266 & 8.2394 & 10.2028 \\
 & INS Gauss & --- & 4.6574 & 4.5683 & --- \\
 & INS Sines & --- & 5.6768 & 5.5701 & --- \\
\midrule
\multirow{6}{*}{Residual Ablation} & CEU Gauss & 11.5755 & 3.9624 & 3.7376 & 11.8012 \\ 
 & CEU KH & 13.6535 & 3.4593 & 9.1220 & 10.9740 \\
 & CEU RC & 20.0013 & 11.1732 & 11.4838 & 18.0761 \\
 & CEU Riemann & 11.6014 & 10.4490 & 8.0780 & 10.9935 \\
 & INS Gauss & --- & 2.2278 & 2.0472 & --- \\
 & INS Sines & --- & 2.8229 & 2.9113 & --- \\
\bottomrule
\end{tabular}
\end{table}

\clearpage
\subsection{ID Evaluation in Spectral Space}
\label{sec:results evaluation in spectral space}
The ability to resolve multi-scale features is a cornerstone of effective neural operators in fluid dynamics. Evaluation in physical space often fails to penalize ``blurry" reconstructions that satisfy mean-error constraints but lack high-frequency turbulent structures. We analyze the spectral response using the following metrics:

\begin{itemize} 

\item \textbf{Minimum Spectral Coherence ($\gamma_{min}$):} Measures the minimum correlation between truth and reconstruction across frequency bands, identifying the ``cutoff" frequency where the model ceases to be physically accurate. A value of $0\%$ represents no coherence, while $100\%$ represents perfect coherence.
$$\gamma^2(k) = \frac{|G_{y\hat{y}}(k)|^2}{G_{yy}(k)G_{\hat{y}\hat{y}}(k)}, \quad \gamma_{min} = \min_{k} \gamma(k) \times100\%$$

\item \textbf{Log Spectral Energy Fidelity ($F_{\log}$):} Measures the similarity in the decibel-scaled (logarithmic) domain. This metric minimizes the impact of energy scale differences, allowing the fidelity of the high-frequency dissipation range to be evaluated on equal footing with the large-scale structures.
$$F_{\log} = \left(1 - \frac{\int |\log_{10} E(k) - \log_{10} \hat{E}(k)| dk}{\int |\log_{10} E(k)| dk}\right) \times100\%$$

\item \textbf{Max Spectral Difference ($\Delta P_{max}$):} We identify periodic artifacts by finding the maximum discrepancy in the 2D power spectrum $P(\mathbf{k})$ in log-space, using a floor $\epsilon$ to avoid machine-precision noise.
$$\Delta P_{max} = \max \left| \log_{10}(P(\mathbf{k}) + \epsilon) - \log_{10}(\hat{P}(\mathbf{k}) + \epsilon) \right|$$
While radially averaged metrics can hide directional artifacts, the maximum spectral difference is sensitive to ``spurs" or ``checkerboard" patterns. It ensures that no single frequency mode—often associated with upsampling artifacts—dominates the reconstruction error.

\end{itemize}

\begin{table}[h!]
\centering
\small
\caption{Comparison of $\gamma_{min} \uparrow$ across all $4\times4$ models, training datasets, and variables.}
\label{tab: min spectral coherence} %spec coh
\begin{tabular}{llcccc} 
\toprule
Model & Dataset & $\rho$ & $u$ & $v$ & $p$ \\
\midrule
\multirow{6}{*}{Continuous AE} &  CEU Gauss & 99.99 & 99.83 & 99.82 & 99.99 \\
 & CEU KH & 99.98 & 99.86 & 93.92 & 100.00 \\
 & CEU RC & 96.65 & 89.54 & 90.00 & 99.91 \\
 & CEU Riemann & 99.88 & 99.00 & 99.04 & 99.89 \\
 & INS Gauss & --- & 99.83 & 99.85 & --- \\
 & INS Sines & --- & 99.70 & 99.67 & --- \\
\midrule
\midrule
\multirow{6}{*}{FSQ} & CEU Gauss & 99.73 & 94.29 & 94.18 & 99.50 \\
 & CEU KH & 98.04 & 94.44 & 71.55 & 100.00 \\
 & CEU RC & 65.41 & 37.62 & 38.34 & 94.31 \\
 & CEU Riemann & 96.59 & 85.49 & 85.16 & 96.77 \\
 & INS Gauss & --- & 96.13 & 96.16 & --- \\
 & INS Sines & --- & 80.92 & 80.45 & --- \\
\midrule
\multirow{6}{*}{IBQ} & CEU Gauss & 95.51 & 29.83 & 29.27 & 92.43 \\
 & CEU KH & 69.28 & 36.70 & 8.50 & 100.00 \\
 & CEU RC & 8.57 & 0.68 & 0.78 & 52.30 \\
 & CEU Riemann & 60.25 & 20.22 & 18.11 & 59.81 \\
 & INS Gauss & --- & 35.21 & 35.79 & --- \\
 & INS Sines & --- & 2.05 & 1.97 & --- \\
\midrule
\multirow{6}{*}{VQ-VAE-2} & CEU Gauss & 99.86 & 90.61 & 90.55 & 99.71 \\
 & CEU KH & 96.54 & 90.64 & 55.87 & 100.00 \\
 & CEU RC & 42.38 & 14.37 & 14.41 & 90.43 \\
 & CEU Riemann & 95.72 & 79.62 & 78.11 & 96.28 \\
 & INS Gauss & --- & 96.82 & 97.01 & --- \\
 & INS Sines & --- & 76.68 & 75.65 & --- \\
\midrule
\multirow{6}{*}{VAR$_{large}$} & CEU Gauss & 99.53 & 88.27 & 88.19 & 99.22 \\
 & CEU KH & 97.65 & 90.25 & 67.18 & 100.00 \\
 & CEU RC & 59.36 & 33.57 & 34.20 & 92.20 \\
 & CEU Riemann & 95.37 & 85.12 & 84.40 & 95.29 \\
 & INS Gauss & --- & 90.65 & 91.01 & --- \\
 & INS Sines & --- & 71.60 & 71.29 & --- \\
\midrule
\multirow{6}{*}{VAR$_{small}$} & CEU Gauss & 99.25 & 83.43 & 83.52 & 98.72 \\
 & CEU KH & 95.79 & 85.44 & 55.76 & 100.00 \\
 & CEU RC & 49.47 & 21.03 & 21.33 & 86.64 \\
 & CEU Riemann & 92.73 & 76.66 & 75.28 & 93.01 \\
 & INS Gauss & --- & 88.16 & 88.15 & --- \\
 & INS Sines & --- & 59.17 & 59.28 & --- \\
\midrule
\multirow{6}{*}{Phaedra} & CEU Gauss & \textbf{99.97} & \textbf{98.23} & \textbf{98.26} & \textbf{99.95} \\
 & CEU KH & \textbf{99.35} & \textbf{98.18} & \textbf{86.88} & \textbf{100.0}0 \\
 & CEU RC & \textbf{84.27} & \textbf{66.56} & \textbf{67.23} & \textbf{97.92} \\
 & CEU Riemann & \textbf{99.03} & \textbf{94.37} & \textbf{93.97} & \textbf{99.16} \\
 & INS Gauss & --- & \textbf{99.05} & \textbf{99.02} & --- \\
 & INS Sines & --- & \textbf{95.77} & \textbf{95.79} & --- \\
\midrule
\multirow{6}{*}{Codebook Ablation} & CEU Gauss & 99.80 & 94.65 & 94.63 & 99.68 \\
 & CEU KH & 98.52 & 95.31 & 77.81 & 100.00 \\
 & CEU RC & 72.37 & 51.65 & 52.56 & 95.64 \\
 & CEU Riemann & 97.36 & 89.21 & 88.67 & 97.30 \\
 & INS Gauss & --- & 96.05 & 96.08 & --- \\
 & INS Sines & --- & 84.35 & 84.77 & --- \\
\midrule
\multirow{6}{*}{Residual Ablation} & CEU Gauss & 99.93 & 95.58 & 95.47 & 99.83 \\
 & CEU KH & 98.73 & 97.10 & 80.88 & 100.00 \\
 & CEU RC & 74.11 & 54.91 & 54.99 & 96.60 \\
 & CEU Riemann & 97.79 & 89.56 & 88.93 & 97.31 \\
 & INS Gauss & --- & 97.19 & 97.60 & --- \\
 & INS Sines & --- & 90.37 & 90.59 & --- \\
\bottomrule
\end{tabular}
\end{table}

\begin{table}[h!]
\centering
\small
\caption{Comparison of Log spectral energy fidelity $F_{\log} \uparrow$ across all $4\times4$ models, training datasets, and variables.}
\label{tab: Flog}
\begin{tabular}{llcccc}
\toprule
Model & Dataset & $\rho$ & $u$ & $v$ & $p$ \\
\midrule
\multirow{6}{*}{Continuous AE} & CEU Gauss & 99.5375 & 98.6984 & 98.6770 & 99.5449 \\
 & CEU KH & 99.1300 & 97.3549 & 89.8986 & 99.7914 \\
 & CEU RC & 97.7406 & 96.5632 & 96.6366 & 98.1140 \\
 & CEU Riemann & 99.1136 & 98.7110 & 98.7952 & 99.3907 \\
 & INS Gauss & --- & 98.8140 & 98.7005 & --- \\
 & INS Sines & --- & 98.1300 & 98.0971 & --- \\
\midrule
\midrule
\multirow{6}{*}{FSQ} & CEU Gauss & 97.3897 & 92.8934 & 93.4634 & 97.0950 \\
 & CEU KH & 94.2305 & 88.0622 & 77.1086 & 99.2448 \\
 & CEU RC & 88.3988 & 86.3705 & 86.4274 & 88.4324 \\
 & CEU Riemann & 95.8999 & 94.4410 & 95.1606 & 96.8478 \\
 & INS Gauss & --- & 94.8299 & 94.8754 & --- \\
 & INS Sines & --- & 90.6183 & 90.5738 & --- \\
\midrule
\multirow{6}{*}{IBQ} & CEU Gauss & 80.2842 & 71.0614 & 71.2729 & 77.8525 \\
 & CEU KH & 78.1060 & 71.3236 & 67.9919 & 93.9853 \\
 & CEU RC & 58.0125 & 61.1280 & 60.6907 & 46.3902 \\
 & CEU Riemann & 79.5127 & 82.7325 & 82.4159 & 81.5519 \\
 & INS Gauss & --- & 74.0509 & 74.1365 & --- \\
 & INS Sines & --- & 72.0738 & 71.5063 & --- \\
\midrule
\multirow{6}{*}{VQ-VAE-2} & CEU Gauss & 96.6785 & 89.1131 & 88.7911 & 96.3025 \\
 & CEU KH & 89.3462 & 81.6161 & 72.1835 & 99.0675 \\
 & CEU RC & 84.3196 & 80.0007 & 80.0342 & 80.8448 \\
 & CEU Riemann & 93.1672 & 92.1933 & 92.4201 & 94.7460 \\
 & INS Gauss & --- & 94.9956 & 94.8881 & --- \\
 & INS Sines & --- & 88.9598 & 88.8400 & --- \\
\midrule
\multirow{6}{*}{VAR$_{large}$} & CEU Gauss & 97.2534 & 92.4595 & 92.6781 & 96.9146 \\
 & CEU KH & 93.7048 & 89.0913 & 77.2612 & 98.5192 \\
 & CEU RC & 86.1282 & 85.4420 & 85.4413 & 87.8285 \\
 & CEU Riemann & 95.7478 & 94.7405 & 95.2965 & 96.6915 \\
 & INS Gauss & --- & 92.5613 & 92.6857 & --- \\
 & INS Sines & --- & 88.7444 & 88.5143 & --- \\
\midrule
\multirow{6}{*}{VAR$_{small}$} & CEU Gauss & 96.2144 & 90.6662 & 90.8125 & 95.3245 \\
 & CEU KH & 93.3232 & 86.2077 & 75.9352 & 98.4822 \\
 & CEU RC & 85.5665 & 82.2292 & 82.6465 & 84.3503 \\
 & CEU Riemann & 94.7157 & 93.4295 & 93.7998 & 95.6554 \\
 & INS Gauss & --- & 90.8588 & 91.3032 & --- \\
 & INS Sines & --- & 86.4380 & 86.4006 & --- \\
\midrule
\multirow{6}{*}{Phaedra} & CEU Gauss & \textbf{98.7787} & \textbf{97.1623} & \textbf{97.2565} & \textbf{98.6961} \\
 & CEU KH & \textbf{97.3386} & \textbf{93.8611} & \textbf{83.9861} & \textbf{99.6212} \\
 & CEU RC & \textbf{94.2088} & \textbf{92.5639} & \textbf{92.3111} & \textbf{94.2493} \\
 & CEU Riemann & \textbf{97.8088} & \textbf{97.4333} & \textbf{97.4803} & \textbf{98.5297} \\
 & INS Gauss & --- & \textbf{98.0049} & \textbf{97.9743} & --- \\
 & INS Sines & --- & \textbf{95.5512} & \textbf{95.5021} & --- \\
\midrule
\multirow{6}{*}{Codebook Ablation} & CEU Gauss & 97.9914 & 93.9751 & 93.6093 & 97.8939 \\
 & CEU KH & 95.3311 & 91.4694 & 80.1426 & 99.3537 \\
 & CEU RC & 91.3980 & 89.7692 & 89.4794 & 91.0116 \\
 & CEU Riemann & 96.6902 & 95.7339 & 95.9641 & 97.2131 \\
 & INS Gauss & --- & 94.9830 & 95.0838 & --- \\
 & INS Sines & --- & 91.5237 & 91.6349 & --- \\
\midrule
\multirow{6}{*}{Residual Ablation} & CEU Gauss & 97.5400 & 95.2212 & 95.2031 & 96.5910 \\
 & CEU KH & 94.2214 & 93.7963 & 82.4042 & 99.1737 \\
 & CEU RC & 89.4408 & 89.4758 & 89.3621 & 89.3775 \\
 & CEU Riemann & 95.3566 & 95.8081 & 95.9204 & 96.0842 \\
 & INS Gauss & --- & 90.4897 & 90.8223 & --- \\
 & INS Sines & --- & 90.5721 & 90.6601 & --- \\
\bottomrule
\end{tabular}
\end{table}

\begin{table}[h!]
\centering
\small
\caption{Comparison of maximum spectral difference $\Delta P_{max}$ across all $4\times4$ models, training datasets, and variables.}
\begin{tabular}{llcccc}
\toprule
Model & Dataset & $\rho$ & $u$ & $v$ & $p$ \\
\midrule
\multirow{6}{*}{Continuous AE} & CEU Gauss & 1.7245 & 3.1961 & 3.1904 & 1.8379 \\
 & CEU KH & 2.7869 & 3.3842 & 4.3291 & 0.5914 \\
 & CEU RC & 3.4748 & 3.7626 & 3.7706 & 3.1601 \\
 & CEU Riemann & 2.8192 & 3.3568 & 3.3694 & 2.5334 \\
 & INS Gauss & --- & 6.1458 & 6.2652 & --- \\
 & INS Sines & --- & 6.3145 & 6.3756 & --- \\
\midrule
\midrule
\multirow{6}{*}{FSQ} & CEU Gauss & 2.7542 & 4.0910 & 4.0706 & 2.9249 \\
 & CEU KH & 3.6352 & 4.1515 & 5.1767 & 1.8626 \\
 & CEU RC & 4.3127 & 4.6508 & 4.6238 & 3.9972 \\
 & CEU Riemann & 3.6928 & 4.1484 & 4.1291 & 3.5492 \\
 & INS Gauss & --- & 7.0809 & 7.0763 & --- \\
 & INS Sines & --- & 7.8109 & 7.8228 & --- \\
\midrule
\multirow{6}{*}{IBQ} & CEU Gauss & 3.7298 & 5.1042 & 5.1033 & 3.9029 \\
 & CEU KH & 4.3250 & 4.8353 & 5.6677 & 3.0003 \\
 & CEU RC & 5.1864 & 5.6314 & 5.6474 & 4.8969 \\
 & CEU Riemann & 4.4723 & 4.8715 & 4.8924 & 4.4445 \\
 & INS Gauss & --- & 7.9481 & 7.9392 & --- \\
 & INS Sines & --- & 8.9313 & 8.8548 & --- \\
\midrule
\multirow{6}{*}{VQ-VAE-2} & CEU Gauss & 2.9147 & 4.2811 & 4.2803 & 3.0996 \\
 & CEU KH & 3.9518 & 4.4162 & 5.3355 & 2.2973 \\
 & CEU RC & 4.5179 & 4.9215 & 4.9169 & 4.2708 \\
 & CEU Riemann & 3.8905 & 4.3145 & 4.3166 & 3.8014 \\
 & INS Gauss & --- & 6.8195 & 7.0097 & --- \\
 & INS Sines & --- & 7.6848 & 7.7590 & --- \\
\midrule
\multirow{6}{*}{VAR$_{large}$} & CEU Gauss & 2.8300 & 4.2049 & 4.1977 & 3.0261 \\
 & CEU KH & 3.7014 & 4.1969 & 5.2397 & 2.0644 \\
 & CEU RC & 4.4381 & 4.6878 & 4.6987 & 4.0347 \\
 & CEU Riemann & 3.7416 & 4.1548 & 4.1615 & 3.6284 \\
 & INS Gauss & --- & 7.1097 & 7.1102 & --- \\
 & INS Sines & --- & 7.9863 & 7.9974 & --- \\
\midrule
\multirow{6}{*}{VAR$_{small}$}& CEU Gauss & 3.0148 & 4.3044 & 4.2987 & 3.2048 \\
 & CEU KH & 3.7892 & 4.2744 & 5.2674 & 2.3395 \\
 & CEU RC & 4.4682 & 4.8438 & 4.8188 & 4.1571 \\
 & CEU Riemann & 3.8582 & 4.2690 & 4.2723 & 3.7514 \\
 & INS Gauss & --- & 7.4189 & 7.2857 & --- \\
 & INS Sines & --- & 8.2338 & 8.2276 & --- \\
\midrule
\multirow{6}{*}{Phaedra} & CEU Gauss & \textbf{2.3412} & \textbf{3.6393} & \textbf{3.6331} & \textbf{2.4757} \\
 & CEU KH & \textbf{3.2888} & \textbf{3.7852} & \textbf{4.8457} & \textbf{1.3705} \\
 & CEU RC & \textbf{3.9565} & \textbf{4.2154} & \textbf{4.2451} & \textbf{3.6434} \\
 & CEU Riemann & \textbf{3.3399} & \textbf{3.8041} & \textbf{3.8077} & \textbf{3.1563} \\
 & INS Gauss & --- & \textbf{6.4060} & \textbf{6.4636} & --- \\
 & INS Sines & --- & \textbf{7.0518} & \textbf{7.0917} & --- \\
\midrule
\multirow{6}{*}{Codebook Ablation} & CEU Gauss & 2.6425 & 3.9857 & 4.0012 & 2.7870 \\
 & CEU KH & 3.4913 & 3.9756 & 5.0169 & 1.7441 \\
 & CEU RC & 4.1540 & 4.4053 & 4.4156 & 3.8440 \\
 & CEU Riemann & 3.5880 & 4.0353 & 4.0399 & 3.4534 \\
 & INS Gauss & --- & 6.9872 & 6.9443 & --- \\
 & INS Sines & --- & 7.7840 & 7.7232 & --- \\
\midrule
\multirow{6}{*}{Residual Ablation} & CEU Gauss & 2.7282 & 4.1136 & 4.1163 & 2.9371 \\
 & CEU KH & 3.6152 & 4.0780 & 4.9722 & 1.5290 \\
 & CEU RC & 4.2448 & 4.4427 & 4.4441 & 3.9020 \\
 & CEU Riemann & 3.5835 & 4.0173 & 4.0424 & 3.4933 \\
 & INS Gauss & --- & 6.3805 & 6.4059 & --- \\
 & INS Sines & --- & 7.1282 & 7.2441 & --- \\
\bottomrule
\end{tabular}
\end{table}

\clearpage
\subsection{ID Evaluation of Token Usage}
\label{sec: results evaluation of token usage}
For quantized models (Phaedra, FSQ, VAR), we evaluate the efficiency and utilization of the latent discrete bottleneck. These metrics characterize whether the model is effectively utilizing its available vocabulary or suffering from codebook collapse. For Phaedra and its ablations, we only calculate these metrics on the morphological stream. This is because we are primarily interested in ensuring the high-dimensional codebook does not suffer from codebook collapse. The amplitude stream is a fixed, dense 1-D scalar lattice (1024 levels) whose values are trivially separable and whose coverage follows the data's amplitude distribution, not learning dynamics, so utilization there says little that would inform tuning.

\begin{itemize} \item \textbf{Codebook Utilization ($U$):} The percentage of the total vocabulary V that is actively utilized across the test set.
$$U=\frac{1}{|V|} \sum_{i=1}^{|V|}1(\text{count}_i >0)\times100\%$$

\item \textbf{Token Entropy ($H$):} Measures the information density of the token distribution. A high entropy suggests that the model uses its codes uniformly to represent the physics.
$$H = - \sum_{i=1}^{|V|} p(z_i) \log_2 p(z_i)$$

\item \textbf{Token Redundancy ($R$):} Quantifies the inefficiency of the encoding relative to a perfectly uniform distribution.
$$R = \left( 1 - \frac{H}{\log_2 |V|} \right) \times 100\%$$
\end{itemize}

\begin{table}[h!]
\centering
\small
\caption{Comparison of codebook utilization $U \uparrow$ across all $4\times4$ models, training datasets, and variables.}
\label{tab: codebook utilization low res}
\begin{tabular}{llcccc}
\toprule
Model & Dataset & $\rho$ & $u$ & $v$ & $p$ \\
\midrule
\multirow{6}{*}{FSQ} & CEU Gauss & 94.3750 & 89.1551 & 89.3403 & 94.2361 \\
 & CEU KH & 93.4259 & 87.8704 & 92.8356 & 94.7222 \\
 & CEU RC & 98.4838 & 99.0162 & 98.9815 & 98.3681 \\
 & CEU Riemann & 96.0301 & 94.2477 & 94.2245 & 92.9398 \\
 & INS Gauss & --- & 53.7500 & 54.4792 & --- \\
 & INS Sines & --- & 86.7014 & 88.2523 & --- \\
\midrule
\multirow{6}{*}{IBQ} & CEU Gauss & 92.2668 & 92.4011 & 92.1570 & 92.2607 \\
 & CEU KH & 91.0645 & 91.7786 & 92.2424 & 92.0349 \\
 & CEU RC & 92.3828 & 92.3584 & 92.3645 & 92.2852 \\
 & CEU Riemann & 92.2363 & 91.9495 & 92.2180 & 92.1204 \\
 & INS Gauss & --- & 92.1631 & 92.2546 & --- \\
 & INS Sines & --- & 92.2363 & 92.2729 & --- \\
\midrule
\multirow{6}{*}{VQ-VAE-2} & CEU Gauss & 3.1189 & 3.1067 & 3.1128 & 3.1189 \\
 & CEU KH & 3.0762 & 3.0640 & 3.1128 & 3.1128 \\
 & CEU RC & 3.1189 & 3.1128 & 3.1128 & 3.1189 \\
 & CEU Riemann & 3.1006 & 3.1128 & 3.1067 & 3.1006 \\
 & INS Gauss & --- & 3.0701 & 3.0457 & --- \\
 & INS Sines & --- & 3.1128 & 3.1128 & --- \\
\midrule
\multirow{6}{*}{VAR$_{large}$} & CEU Gauss & 82.4537 & 84.5139 & 84.1204 & 82.0370 \\
 & CEU KH & 85.6713 & 85.2778 & 80.7292 & 69.7454 \\
 & CEU RC & 73.8310 & 74.3287 & 74.0278 & 76.1227 \\
 & CEU Riemann & 84.4444 & 85.9144 & 85.6481 & 85.5671 \\
 & INS Gauss & --- & 85.9954 & 85.8449 & --- \\
 & INS Sines & --- & 75.9144 & 76.1921 & --- \\
\midrule
\multirow{6}{*}{VAR$_{small}$} & CEU Gauss & 75.6366 & 79.4676 & 79.1204 & 75.4398 \\
 & CEU KH & 81.6435 & 80.7870 & 72.0139 & 49.8380 \\
 & CEU RC & 57.6042 & 54.8264 & 55.7176 & 58.7847 \\
 & CEU Riemann & 80.2894 & 83.1713 & 83.1366 & 82.5926 \\
 & INS Gauss & --- & 81.8287 & 80.9259 & --- \\
 & INS Sines & --- & 63.4838 & 63.9815 & --- \\
\midrule
\multirow{6}{*}{Phaedra} & CEU Gauss & 99.8495 & 91.7361 & 92.5694 & 99.8032 \\
 & CEU KH & 100.0000 & 91.5741 & 98.8889 & 99.9884 \\
 & CEU RC & 100.0000 & 100.0000 & 100.0000 & 100.0000 \\
 & CEU Riemann & 99.9884 & 99.7917 & 99.5718 & 99.4560 \\
 & INS Gauss & --- & 25.1389 & 24.1319 & --- \\
 & INS Sines & --- & 79.9537 & 83.2755 & --- \\
\midrule
\multirow{6}{*}{Codebook Ablation} & CEU Gauss & 91.1921 & 92.8819 & 92.9051 & 90.9028 \\
 & CEU KH & 91.7130 & 92.2917 & 89.9537 & 72.1296 \\
 & CEU RC & 67.2685 & 66.0417 & 66.0995 & 69.0741 \\
 & CEU Riemann & 91.4468 & 93.2755 & 93.1366 & 92.4421 \\
 & INS Gauss & --- & 94.8032 & 94.9306 & --- \\
 & INS Sines & --- & 85.9722 & 85.9722 & --- \\
\midrule
\multirow{6}{*}{Residual Ablation} & CEU Gauss & 98.0440 & 88.8079 & 89.3750 & 97.9630 \\
 & CEU KH & 99.3056 & 87.2917 & 90.9722 & 86.8056 \\
 & CEU RC & 99.9421 & 99.2477 & 99.2014 & 99.8958 \\
 & CEU Riemann & 99.1435 & 96.5278 & 96.8056 & 99.0509 \\
 & INS Gauss & --- & 58.3218 & 59.1088 & --- \\
 & INS Sines & --- & 82.4653 & 80.5440 & --- \\
\bottomrule
\end{tabular}
\end{table}

\begin{table}[h!]
\centering
\small
\caption{Comparison of token entropy $H \uparrow$ across all $4\times4$ models, training datasets, and variables.}
\label{tab: token entropy low res}
\begin{tabular}{llcccc}
\toprule
Model & Dataset & $\rho$ & $u$ & $v$ & $p$ \\
\midrule
\multirow{6}{*}{FSQ} & CEU Gauss & 10.4052 & 9.9868 & 9.9988 & 10.3707 \\
 & CEU KH & 9.7379 & 9.8794 & 10.5237 & 11.5640 \\
 & CEU RC & 12.0170 & 12.0983 & 12.1081 & 11.9914 \\
 & CEU Riemann & 10.3948 & 10.1950 & 10.2560 & 10.0298 \\
 & INS Gauss & --- & 8.7338 & 8.7793 & --- \\
 & INS Sines & --- & 10.4527 & 10.5076 & --- \\
\midrule
\multirow{6}{*}{IBQ} & CEU Gauss & 13.2099 & 13.3916 & 13.3718 & 13.2100 \\
 & CEU KH & 12.5941 & 13.0446 & 13.1207 & 13.1206 \\
 & CEU RC & 13.3900 & 13.4048 & 13.4021 & 13.4136 \\
 & CEU Riemann & 13.2279 & 13.1815 & 13.1578 & 13.2261 \\
 & INS Gauss & --- & 13.3423 & 13.3621 & --- \\
 & INS Sines & --- & 13.3270 & 13.3797 & --- \\
\midrule
\multirow{6}{*}{VQ-VAE-2} & CEU Gauss & 8.1152 & 7.8774 & 7.9037 & 8.0978 \\
 & CEU KH & 7.7807 & 7.7601 & 8.1538 & 8.5364 \\
 & CEU RC & 8.3999 & 8.5027 & 8.5068 & 8.4887 \\
 & CEU Riemann & 8.0811 & 8.1386 & 8.1070 & 7.9660 \\
 & INS Gauss & --- & 7.2617 & 7.2919 & --- \\
 & INS Sines & --- & 8.5078 & 8.5068 & --- \\
\midrule
\multirow{6}{*}{VAR$_{large}$} & CEU Gauss & 9.1903 & 9.2200 & 9.2183 & 9.1598 \\
 & CEU KH & 9.4007 & 9.2950 & 9.1263 & 8.9824 \\
 & CEU RC & 9.0556 & 9.0543 & 9.0502 & 9.0693 \\
 & CEU Riemann & 9.2814 & 9.3750 & 9.3583 & 9.3325 \\
 & INS Gauss & --- & 9.3461 & 9.3395 & --- \\
 & INS Sines & --- & 8.9444 & 8.9386 & --- \\
\midrule
\multirow{6}{*}{VAR$_{small}$} & CEU Gauss & 9.2246 & 9.2915 & 9.2808 & 9.2284 \\
 & CEU KH & 9.3548 & 9.3389 & 9.1479 & 8.5886 \\
 & CEU RC & 8.7283 & 8.7190 & 8.7278 & 8.8065 \\
 & CEU Riemann & 9.3077 & 9.4131 & 9.4067 & 9.3784 \\
 & INS Gauss & --- & 9.3463 & 9.3623 & --- \\
 & INS Sines & --- & 8.8798 & 8.8808 & --- \\
\midrule
\multirow{6}{*}{Phaedra} & CEU Gauss & 9.1402 & 8.5534 & 8.6189 & 9.0934 \\
 & CEU KH & 9.1498 & 8.6559 & 9.5233 & 11.0973 \\
 & CEU RC & 12.2623 & 12.2658 & 12.2754 & 12.1710 \\
 & CEU Riemann & 9.4154 & 8.8793 & 8.9665 & 8.6407 \\
 & INS Gauss & --- & 4.7851 & 4.8563 & --- \\
 & INS Sines & --- & 9.3984 & 9.4578 & --- \\
\midrule
\multirow{6}{*}{Codebook Ablation} & CEU Gauss & 9.7855 & 9.8618 & 9.8548 & 9.7976 \\
 & CEU KH & 9.7251 & 9.8785 & 9.7606 & 9.0702 \\
 & CEU RC & 8.8350 & 8.9170 & 8.9234 & 8.9906 \\
 & CEU Riemann & 9.7236 & 9.8212 & 9.8497 & 9.8034 \\
 & INS Gauss & --- & 9.9260 & 9.9352 & --- \\
 & INS Sines & --- & 9.4265 & 9.3923 & --- \\
\midrule
\multirow{6}{*}{Residual Ablation} & CEU Gauss & 10.5090 & 10.1459 & 10.1638 & 10.4989 \\
 & CEU KH & 10.4531 & 9.8494 & 10.6353 & 11.2747 \\
 & CEU RC & 12.0073 & 12.0298 & 12.0281 & 12.0282 \\
 & CEU Riemann & 10.8142 & 10.6561 & 10.6447 & 10.4639 \\
 & INS Gauss & --- & 8.6662 & 8.7133 & --- \\
 & INS Sines & --- & 10.8044 & 10.8467 & --- \\
\bottomrule
\end{tabular}
\end{table}

\begin{table}[h!]
\centering
\small
\caption{Comparison of token redundancy $R\downarrow$ across all $4\times4$ models, training datasets, and variables.}
\label{tab: token redundancy low res}
\begin{tabular}{llcccc}
\toprule
Model & Dataset & $\rho$ & $u$ & $v$ & $p$ \\
\midrule
\multirow{6}{*}{FSQ} & CEU Gauss & 20.4305 & 23.6296 & 23.5377 & 20.6941 \\
 & CEU KH & 25.5333 & 24.4507 & 19.5239 & 11.5686 \\
 & CEU RC & 8.1044 & 7.4829 & 7.4079 & 8.3005 \\
 & CEU Riemann & 20.5100 & 22.0374 & 21.5712 & 23.3010 \\
 & INS Gauss & --- & 33.2113 & 32.8633 & --- \\
 & INS Sines & --- & 20.0666 & 19.6474 & --- \\
\midrule
\multirow{6}{*}{IBQ} & CEU Gauss & 5.6433 & 4.3459 & 4.4870 & 5.6431 \\
 & CEU KH & 10.0419 & 6.8240 & 6.2804 & 6.2812 \\
 & CEU RC & 4.3570 & 4.2514 & 4.2704 & 4.1887 \\
 & CEU Riemann & 5.5150 & 5.8464 & 6.0159 & 5.5278 \\
 & INS Gauss & --- & 4.6979 & 4.5565 & --- \\
 & INS Sines & --- & 4.8072 & 4.4309 & --- \\
\midrule
\multirow{6}{*}{VQ-VAE-2} & CEU Gauss & 42.0343 & 43.7330 & 43.5450 & 42.1584 \\
 & CEU KH & 44.4236 & 44.5704 & 41.7586 & 39.0257 \\
 & CEU RC & 40.0010 & 39.2665 & 39.2373 & 39.3667 \\
 & CEU Riemann & 42.2778 & 41.8669 & 42.0932 & 43.1000 \\
 & INS Gauss & --- & 48.1306 & 47.9150 & --- \\
 & INS Sines & --- & 39.2303 & 39.2373 & --- \\
\midrule
\multirow{6}{*}{VAR$_{large}$} & CEU Gauss & 29.7203 & 29.4939 & 29.5067 & 29.9538 \\
 & CEU KH & 28.1113 & 28.9204 & 30.2103 & 31.3103 \\
 & CEU RC & 30.7510 & 30.7604 & 30.7919 & 30.6462 \\
 & CEU Riemann & 29.0241 & 28.3080 & 28.4361 & 28.6334 \\
 & INS Gauss & --- & 28.5294 & 28.5799 & --- \\
 & INS Sines & --- & 31.6013 & 31.6457 & --- \\
\midrule
\multirow{6}{*}{VAR$_{small}$} & CEU Gauss & 29.4582 & 28.9466 & 29.0283 & 29.4296 \\
 & CEU KH & 28.4624 & 28.5846 & 30.0446 & 34.3220 \\
 & CEU RC & 33.2535 & 33.3244 & 33.2572 & 32.6557 \\
 & CEU Riemann & 28.8226 & 28.0172 & 28.0656 & 28.2824 \\
 & INS Gauss & --- & 28.5274 & 28.4056 & --- \\
 & INS Sines & --- & 32.0951 & 32.0874 & --- \\
\midrule
\multirow{6}{*}{Phaedra} & CEU Gauss & 30.1041 & 34.5912 & 34.0901 & 30.4619 \\
 & CEU KH & 30.0307 & 33.8069 & 27.1744 & 15.1373 \\
 & CEU RC & 6.2290 & 6.2019 & 6.1286 & 6.9270 \\
 & CEU Riemann & 27.9990 & 32.0990 & 31.4325 & 33.9232 \\
 & INS Gauss & --- & 63.4078 & 62.8630 & --- \\
 & INS Sines & --- & 28.1294 & 27.6753 & --- \\
\midrule
\multirow{6}{*}{Codebook Ablation} & CEU Gauss & 25.1690 & 24.5854 & 24.6390 & 25.0768 \\
 & CEU KH & 25.6308 & 24.4579 & 25.3598 & 30.6394 \\
 & CEU RC & 32.4376 & 31.8104 & 31.7613 & 31.2481 \\
 & CEU Riemann & 25.6421 & 24.8962 & 24.6782 & 25.0322 \\
 & INS Gauss & --- & 24.0950 & 24.0242 & --- \\
 & INS Sines & --- & 27.9141 & 28.1756 & --- \\
\midrule
\multirow{6}{*}{Residual Ablation} & CEU Gauss & 19.6364 & 22.4128 & 22.2759 & 19.7138 \\
 & CEU KH & 20.0635 & 24.6807 & 18.6704 & 13.7813 \\
 & CEU RC & 8.1788 & 8.0068 & 8.0197 & 8.0187 \\
 & CEU Riemann & 17.3022 & 18.5113 & 18.5989 & 19.9812 \\
 & INS Gauss & --- & 33.7284 & 33.3686 & --- \\
 & INS Sines & --- & 17.3777 & 17.0540 & --- \\
\bottomrule
\end{tabular}
\end{table}

\clearpage
\subsection{OD$_1$ \& OD$_2$ Summary}
\label{sec: results out of distribution od1 od2 PDEs}
\begin{table}[h!]
    \centering
    \caption{Results for out of distribution datasets. CEU RKH, AIR, and INS SVS comprise OD$_1$, while all other datasets comprise OD$_2$.}
    \resizebox{0.7\textwidth}{!}{
    \begin{tabular}{llccccc}
    \toprule
       Model & Dataset & nMAE$\downarrow$ & nRMSE$\downarrow$ & $\Delta \sigma^2_{loc}\downarrow$ & $\gamma_{min}\uparrow$ & Utilization$\uparrow$\\
       \midrule
          \multirow{7}{*}{Continuous AE} & CEU RKH, $\rho$ & 2.6836 & 4.5891 & 5.1315 & 0.9788 & --- \\
           & CEU AIR, $\rho$ & 0.4410 & 1.1844 & 4.0184 & 0.9998 & --- \\
           & INS SVS, $u$ & 0.3382 & 0.4635 & 1.0631 & 0.9985 & --- \\
           & POI, $u$ & 0.7140 & 1.0170 & 1.27e-08 & 0.9671 & --- \\
           & DAR, $u$ & 0.4245 & 0.6117 & 1.6570 & 0.9981 & --- \\
           & ALC, $u$ & 1.4171 & 1.9240 & 3.6923 & 0.9894 & --- \\
           & AWA, $u$ & 5.3110 & 7.7318 & 6.7124 & 0.9250 & --- \\
        \midrule 
        \midrule
        \multirow{7}{*}{FSQ} & CEU RKH, $\rho$ & 2.7441 & 6.0136 & 18.6689 & 0.8824 & 38.5185 \\
           & CEU AIR, $\rho$ & 1.9414 & 5.4816 & 36.0707 & 0.9955 & 17.9861 \\
           & INS SVS, $u$ & 0.9487 & 1.4973 & 6.9149 & 0.9385 & 23.1019 \\
           & POI, $u$ & 6.1093 & 8.1953 & 9.37e-08 & 0.0089 & 61.3426 \\
           & DAR, $u$ & 3.7399 & 4.8796 & 16.6083 & 0.9096 & 14.4444 \\
           & ALC, $u$ & 6.7952 & 9.8960 & 19.4266 & 0.7951 & 31.1921 \\
           & AWA, $u$ & 2.8160 & 3.7289 & 10.5929 & 0.8587 & 37.0718 \\
        \midrule 
          \multirow{7}{*}{IBQ} & CEU RKH, $\rho$ & 11.6534 & 19.3687 & 57.4506 & 0.2665 & 3.9185 \\
           & CEU AIR, $\rho$ & 6.1183 & 13.1568 & 64.96 & 0.9441 & 28.7781 \\
           & INS SVS, $u$ & 3.8131 & 6.8924 & 28.0073 & 0.2030 & 2.6794 \\
           & POI, $u$ & 25.0840 & 33.1060 & 3.35e-07 & 0.0010 & 63.0798 \\
           & DAR, $u$ & 15.3045 & 18.7602 & 92.1098 & 0.4491 & 1.4343 \\
           & ALC, $u$ & 21.7913 & 32.0495 & 86.0836 & 0.2358 & 2.2705 \\
           & AWA, $u$ & 12.1177 & 15.7567 & 41.8770 & 0.0540 & 3.8330 \\
        \midrule 
          \multirow{7}{*}{VQ-VAE-2} & CEU RKH, $\rho$ & 3.9253 & 9.5092 & 27.9232 & 0.6416 & 2.0081 \\
           & CEU AIR, $\rho$ & 1.6594 & 6.8367 & 31.5205 & 0.9982 & 1.5137 \\
           & INS SVS, $u$ & 0.7023 & 1.1821 & 5.2197 & 0.9503 & 1.6052 \\
           & POI, $u$ & 11.7889 & 15.5432 & 1.22e-07 & 4.36e-04 & 0.8240 \\
           & DAR, $u$ & 2.6377 & 3.3503 & 11.2195 & 0.9500 & 1.0071 \\
           & ALC, $u$ & 15.0539 & 22.7377 & 48.1380 & 0.2244 & 1.5625 \\
           & AWA, $u$ & 2.1241 & 2.7352 & 8.8961 & 0.8802 & 2.0264 \\
        \midrule
          \multirow{7}{*}{VAR$_{large}$} & CEU RKH, $\rho$ & 3.3204 & 6.7568 & 19.4989 & 0.8594 & 23.8426 \\
           & CEU AIR, $\rho$ & 1.9713 & 4.8634 & 21.7771 & 0.9957 & 19.2130 \\
           & INS SVS, $u$ & 1.2462 & 1.8364 & 6.9619 & 0.9041 & 22.3032 \\
           & POI, $u$ & 11.3193 & 14.7730 & 1.93e-07 & 0.0017 & 20.8449 \\
           & DAR, $u$ & 4.4314 & 5.7686 & 19.7863 & 0.7902 & 12.8588 \\
           & ALC, $u$ & 7.0034 & 10.9251 & 26.5046 & 0.7649 & 26.1458 \\
           & AWA, $u$ & 4.0899 & 5.1140 & 12.8110 & 0.7944 & 23.2407 \\      
           \midrule
          \multirow{7}{*}{VAR$_{small}$} & CEU RKH, $\rho$ & 4.1443 & 8.1576 & 25.2500 & 0.7950 & 28.9931 \\
           & CEU AIR, $\rho$ & 2.2530 & 5.5115 & 21.2046 & 0.9928 & 25.3704 \\
           & INS SVS, $u$ & 1.4933 & 2.4126 & 10.6312 & 0.8555 & 28.9468 \\
           & POI, $u$ & 20.5913 & 28.1804 & 4.57e-07 & 9.37e-04 & 17.1528 \\
           & DAR, $u$ & 6.4258 & 8.7358 & 35.5045 & 0.1246 & 15.4282 \\
           & ALC, $u$ & 8.5392 & 13.6259 & 36.2196 & 0.7014 & 28.0208 \\
           & AWA, $u$ & 5.0149 & 6.3306 & 14.8620 & 0.7032 & 29.4097 \\
         \midrule
          \multirow{8}{*}{Phaedra} & CEU RKH, $\rho$ & \textbf{2.1480} & \textbf{3.9824} & \textbf{8.4787} & \textbf{0.9514} & {45.5787} \\
           & CEU AIR, $\rho$ & \textbf{1.0813} & \textbf{2.7147} & \textbf{8.5552} & \textbf{0.9997} & {31.2153} \\
           & INS SVS, $u$ & \textbf{0.4424} & \textbf{0.6084} & \textbf{2.3872} & \textbf{0.9914} & {29.1551} \\
           & POI, $u$ & \textbf{4.5620} & \textbf{5.9460} & \textbf{6.19e-08} & \textbf{0.2883} & {46.4815} \\
           & DAR, $u$ & \textbf{1.2634} & \textbf{1.7240} & \textbf{4.6661} & \textbf{0.9868} & {25.3356} \\
           & ALC, $u$ & \textbf{4.4200} & \textbf{6.0337} & \textbf{12.5190} & \textbf{0.8876} & {44.3866} \\
           & AWA, $u$ & \textbf{2.3419} & \textbf{3.2453} & \textbf{6.1183} & \textbf{0.9428} & {40.7639} \\
    \bottomrule
    \end{tabular}}
    \label{tab:out of distribution results}
\end{table}

\clearpage

\subsection{Scaling}
\label{sec: scaling results with respect to bottleneck resolution}
\begin{figure}[h]
     \centering
     \begin{subfigure}[b]{0.8\textwidth}
         \centering
         \includegraphics[width=\textwidth]{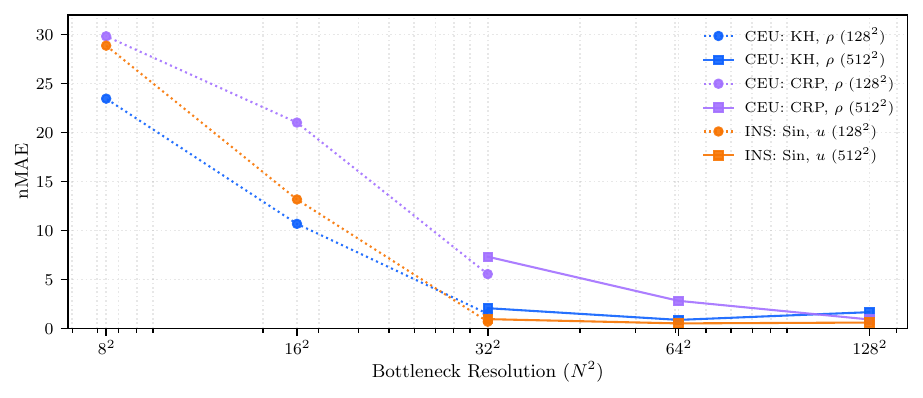}
     \end{subfigure}

     \begin{subfigure}[b]{0.8\textwidth}
         \centering
         \includegraphics[width=\textwidth]{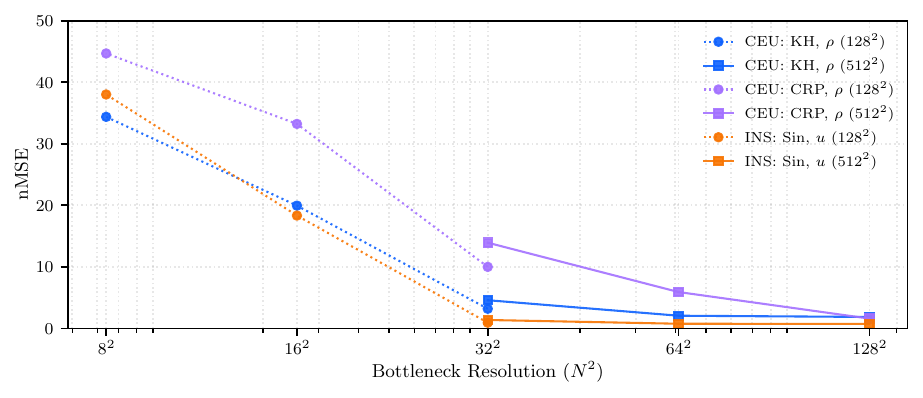}
     \end{subfigure}

     \begin{subfigure}[b]{0.8\textwidth}
         \centering
         \includegraphics[width=\textwidth]{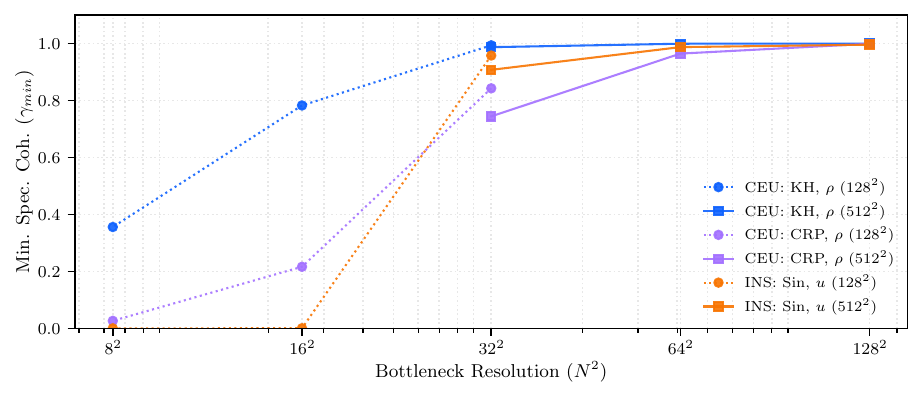}
     \end{subfigure}

     \caption{\textbf{Scaling with respect to bottleneck resolution.} We observe that the model scales with respect to the absolute number of tokens as opposed to the ratio of downsampling. That is, Phaedra performs equally well using $16^2$ downsampling on high resolution ($512^2$) data as when using $4^2$ downsampling on low resolution ($128^2$) data, as each compresses the input to $32\times32$ tokens. We also observe a major drop-off in performance when using fewer than $32\times32$ tokens, with diminishing returns as the number of tokens increases. This is illustrated above for the normalized MAE, MSE, and minimal spectral coherence.}
     \label{fig:scaling results}
\end{figure}

\begin{figure}
    \centering
    \includegraphics[width=0.8\linewidth]{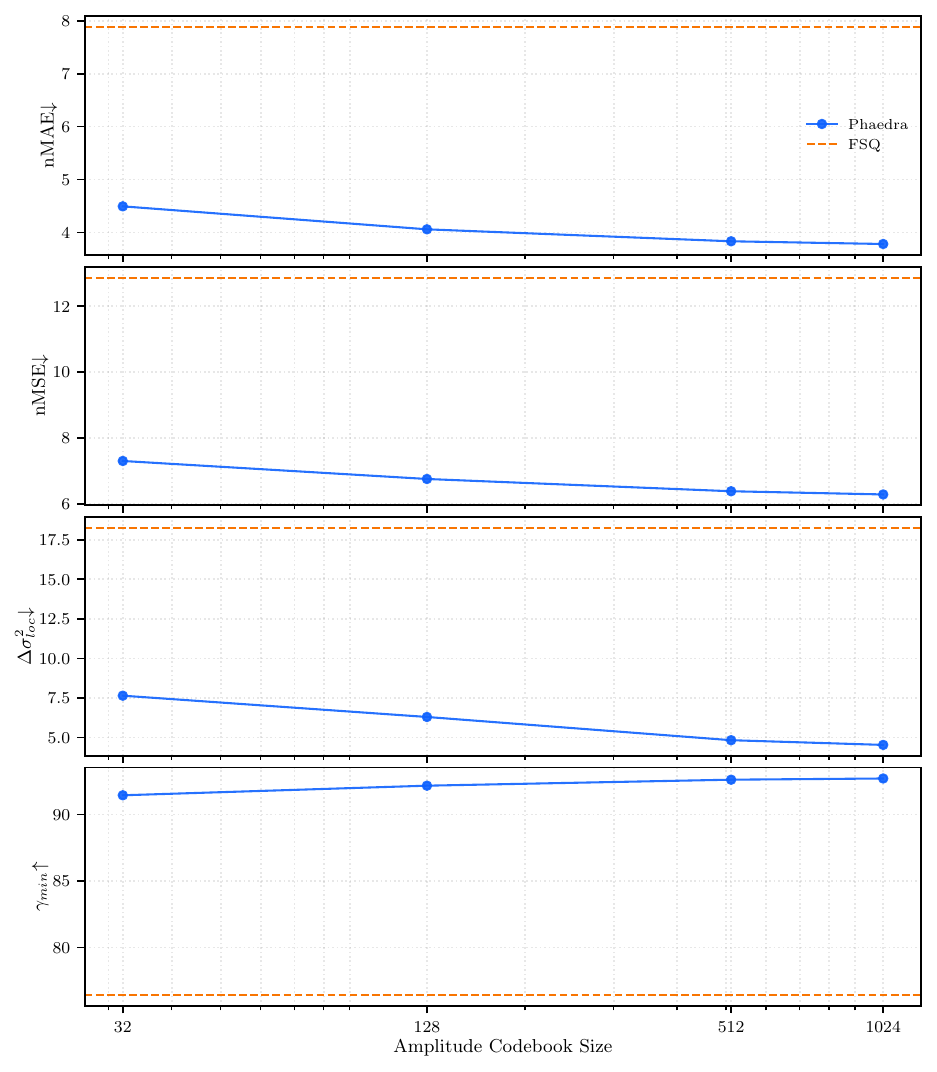}
    \caption{\textbf{Scaling with respect to amplitude codebook size.} We trained Phaedra and FSQ on the CEU RC $\rho$ dataset to observe scaling performance as the number of available amplitude tokens increases. While increasing the size of the amplitude codebook consistently yields better results, these gains become extremely marginal after $\sim 512$ tokens. Even with a codebook size of 32, Phaedra already exhibits considerable gains compared to the FSQ baseline.}
    \label{fig:Scaling wrt amplitude codebook size}
\end{figure}

\clearpage
\subsection{Earth Observation: Global Evaluations}
\label{sec: earth observation numerical results}

In Table~\ref{tab:combined_eo_results}, we evaluate the performance of Phaedra and baseline models on globally distributed Earth observation. None of the data products have been seen during pretraining and, therefore, form strong out-of-distribution tasks. Across tasks, we observe desirable performances of Phaedra compared to existing quantization approaches, thereby closing the gap to continuous tokenizers.

We observe that Phaedra$_4$ consistently narrows the gap to the continuous tokenizer upper bound across all modalities, while outperforming FSQ on the majority of metrics and datasets. This pattern is most pronounced for high‑dimensional optical data (Sentinel‑2 L2A/L1C), where Phaedra$_4$ reduces nMAE and nRMSE substantially relative to FSQ and closely tracks the continuous reference (e.g., S2‑L2A nRMSE: 31.76$\pm$67.89 for Phaedra$_4$ vs. 52.85$\pm$113.70 for FSQ, and 30.17$\pm$69.42 for continuous). These results indicate that Phaedra retains much of the fidelity of continuous representations while using a compact tokenization scheme.

In line with our expectation, when increasing the patch size, performance declines. Across optical modalities, Phaedra$_8$ frequently underperforms Phaedra$_4$ (e.g., S2‑L1C nRMSE: 36.81$\pm$56.12 vs. 31.84$\pm$57.45), indicating that larger patches per codebook token increase compression losses. A similar degradation is observed for RGB data, where Phaedra$_8$ exhibits noticeably higher error than both Phaedra$_4$ and FSQ, while Phaedra$_4$ remains competitive with the Continuous reference. Performance trends on C‑band SAR data further support the robustness of Phaedra$_4$. On Sentinel‑1 RTC, Phaedra$_4$ delivers a clear improvement over FSQ (nRMSE 0.084$\pm$0.046 vs. 0.101$\pm$0.040) and recovers a substantial fraction of the performance of the continuous tokenizer (0.047$\pm$0.024). This behavior demonstrates effective generalization to speckle‑dominated backscatter and to signal distributions that differ markedly from optical imagery.

Static and smoothly varying geophysical fields particularly benefit from Phaedra’s representation. For digital elevation data, Phaedra$_4$ closely approaches the continuous upper bound (nRMSE 11.41$\pm$35.28 vs. 9.27$\pm$39.88) while substantially outperforming FSQ (23.63$\pm$86.39), indicating strong modeling of low‑frequency structure and terrain coherence. A similar pattern emerges for NDVI, where absolute errors are low across all methods, yet Phaedra$_4$ remains consistently closer to the Continuous reference than FSQ (nRMSE  0.002$\pm$0.001 vs. 0.003$\pm$0.005), suggesting minimal distortion of vegetation dynamics under discretization.

Overall, these results highlight Phaedra as a strong quantization model on unseen, physical data across heterogeneous Earth observation modalities, including multispectral optical data, SAR, topography, and vegetation indices. Phaedra closely approaches the performance of continuous tokenizers while consistently surpassing our FSQ‑based baseline.

\begin{table}[ht!] 
\centering
\caption{Comprehensive evaluation across Earth Observation and Satellite datasets. Metrics are reported as $\text{mean} \pm \text{std}$. Relative metrics (r$L_1$/r$L_2$) for DEM and NDVI are omitted due to numerical instability caused by zero-valued pixels in the ground truth. Results in this table for Sentinel-2 L1C differ from those in the main text because these are for a large-scale global evaluation, while the main text presented results for specific regions.}
\label{tab:combined_eo_results}
\resizebox{\linewidth}{!}{
\begin{tabular}{llcccc}
\toprule

\textbf{Dataset} & \textbf{Model} & \textbf{nMAE} $\downarrow$ & \textbf{nRMSE} $\downarrow$ & \textbf{r$L_1$} $\downarrow$ & \textbf{r$L_2$} $\downarrow$ \\
\midrule
\multirow{4}{*}{Sentinel-2 L2A} & Continuous & 21.583 $\pm$ 60.032 & 30.169 $\pm$ 69.422 & 3.687 $\pm$ 6.950 & 5.790 $\pm$ 8.281 \\
                                & FSQ        & 31.650 $\pm$ 79.865 & 52.845 $\pm$ 113.701 & 5.880 $\pm$ 9.838 & 10.113 $\pm$ 15.647 \\
                                & Phaedra$_4$ & 23.719 $\pm$ 58.275 & 31.756 $\pm$ 67.894 & 4.572 $\pm$ 7.132 & 6.446 $\pm$ 8.490 \\
                                & Phaedra$_8$ & 30.178 $\pm$ 56.766 & 41.157 $\pm$ 65.523 & 6.394 $\pm$ 6.389 & 9.276 $\pm$ 7.392 \\
\midrule
\multirow{4}{*}{Sentinel-2 L1C} & Continuous & 21.198 $\pm$ 50.801 & 33.021 $\pm$ 60.200 & 4.214 $\pm$ 6.814 & 7.303 $\pm$ 8.156 \\
                                & FSQ        & 22.098 $\pm$ 21.393 & 51.321 $\pm$ 47.126 & 5.904 $\pm$ 4.410 & 13.675 $\pm$ 9.907 \\
                                & Phaedra$_4$ & 21.658 $\pm$ 47.982 & 31.837 $\pm$ 57.445 & 4.557 $\pm$ 6.493 & 7.153 $\pm$ 7.819 \\
                                & Phaedra$_8$ & 26.564 $\pm$ 47.131 & 36.808 $\pm$ 56.123 & 6.008 $\pm$ 6.125 & 8.721 $\pm$ 7.226 \\
\midrule
\multirow{4}{*}{Sentinel-2 RGB} & Continuous & 0.410 $\pm$ 0.719 & 0.549 $\pm$ 0.828 & 4.022 $\pm$ 8.876 & 5.614 $\pm$ 12.121 \\
                                & FSQ        & 0.534 $\pm$ 0.261 & 0.787 $\pm$ 0.407 & 5.815 $\pm$ 3.559 & 8.572 $\pm$ 5.858 \\
                                & Phaedra$_4$ & 0.547 $\pm$ 0.755 & 0.734 $\pm$ 0.855 & 5.390 $\pm$ 4.610 & 7.438 $\pm$ 5.716 \\
                                & Phaedra$_8$ & 0.989 $\pm$ 0.773 & 1.375 $\pm$ 0.874 & 10.784 $\pm$ 7.556 & 15.539 $\pm$ 9.924 \\
\midrule
\multirow{4}{*}{Sentinel-1 RTC} & Continuous & 0.035 $\pm$ 0.021 & 0.047 $\pm$ 0.024 & 2.540 $\pm$ 1.395 & 3.406 $\pm$ 1.816 \\
                                & FSQ        & 0.078 $\pm$ 0.035 & 0.101 $\pm$ 0.040 & 5.628 $\pm$ 1.867 & 7.301 $\pm$ 2.518 \\
                                & Phaedra$_4$ & 0.065 $\pm$ 0.041 & 0.084 $\pm$ 0.046 & 4.645 $\pm$ 1.989 & 6.045 $\pm$ 2.330 \\
                                & Phaedra$_8$ & 0.138 $\pm$ 0.022 & 0.174 $\pm$ 0.026 & 10.292 $\pm$ 2.907 & 13.067 $\pm$ 3.738 \\
\midrule
\multirow{4}{*}{Dig. Elevation} & Continuous & 8.767 $\pm$ 39.608 & 9.265 $\pm$ 39.882 & --- & --- \\
                                & FSQ        & 22.855 $\pm$ 85.443 & 23.626 $\pm$ 86.390 & --- & --- \\
                                & Phaedra$_4$ & 10.933 $\pm$ 35.323 & 11.408 $\pm$ 35.280 & --- & --- \\
                                & Phaedra$_8$ & 12.046 $\pm$ 33.090 & 12.767 $\pm$ 33.217 & --- & --- \\
\midrule
\multirow{4}{*}{NDVI}           & Continuous & 0.001 $\pm$ 0.003 & 0.002 $\pm$ 0.003 & --- & --- \\
                                & FSQ        & 0.002 $\pm$ 0.001 & 0.002 $\pm$ 0.001 & --- & --- \\
                                & Phaedra$_4$ & 0.002 $\pm$ 0.004 & 0.003 $\pm$ 0.005 & --- & --- \\
                                & Phaedra$_8$ & 0.004 $\pm$ 0.004 & 0.005 $\pm$ 0.004 & --- & --- \\
\bottomrule
\end{tabular}}
\end{table}

\clearpage
\subsection{ERA-5 Evaluations}
\label{sec: era5 numerical results}

In Table~\ref{tab:era5_models_x_metrics_mean_std}, we evaluate the performance of Phaedra and baseline models on ERA5 {zonal (U) and meridional (V) wind components}. These atmospheric reanalyses differ fundamentally from optical and land‑surface data, exhibiting smooth spatiotemporal structure alongside sharp gradients in dynamically active regions. As such, they constitute a good out-of-distribution test for discrete tokenization under physically constrained, continuous flow fields.

Across all metrics, Phaedra$_4$ substantially improves upon the FSQ baseline while consistently narrowing the gap to the continuous tokenizer upper bound. In terms of normalized errors, Phaedra$_4$ reduces nMAE from 0.062$\pm$0.001 (FSQ) to 0.042$\pm$0.000 and nMSE from 0.091$\pm$0.001 to 0.059$\pm$0.001, recovering a large fraction of the performance lost under discretization. While a gap to the continuous reference remains (nMAE 0.023$\pm$0.000, nMSE 0.034$\pm$0.001), Phaedra$_4$ consistently halves the error increase incurred by FSQ, indicating higher fidelity in representing variables from global reanalyses data. A similar pattern is observed in the rescaled error metrics. Phaedra$_4$ achieves rMAE 6.391$\pm$0.250 and rMSE 9.117$\pm$0.362, compared to substantially higher errors for FSQ (rMAE 9.567$\pm$0.330, rMSE 14.003$\pm$0.482). This improvement suggests that Phaedra more effectively preserves relative variations in wind magnitude, which is critical for downstream tasks sensitive to gradients and flow structure.

\begin{table}[h]
\centering
\caption{ERA5 reanalyses: Zonal and meridional winds. Metrics measured in mean $\pm$ std.}\label{tab:era5_models_x_metrics_mean_std}
\begin{tabular}{lcccc}
\toprule
 & nMAE$\downarrow$ & nMSE$\downarrow$ & r$L_1\downarrow$ & r$L_2\downarrow$ \\
\midrule
Continuous & 0.023 ± 0.000 & 0.034 ± 0.001 & 3.575 ± 0.144 & 5.254 ± 0.216\\\midrule
FSQ & 0.062 ± 0.001 & 0.091 ± 0.001 & 9.567 ± 0.330 & 14.003 ± 0.482 \\
Phaedra$_4$ & 0.042 ± 0.000 & 0.059 ± 0.001 & 6.391 ± 0.250 & 9.117 ± 0.362 \\
Phaedra$_8$ & 0.156 ± 0.003 & 0.231 ± 0.006 & 23.912 ± 1.098 & 35.491 ± 1.679  \\
\bottomrule
\end{tabular}
\end{table}

For global surface temperature, we observe the same patterns and come to the same conclusions as for zonal and meridional winds.

\begin{table}[h]
\centering
\caption{ERA5 reanalyses: Temperature. Metrics measured in mean $\pm$ std.}\label{tab:era5_models_t2m_metrics_mean_std}
\begin{tabular}{lcccc}
\toprule
 & nMAE$\downarrow$ & nMSE$\downarrow$ & r$L_1\downarrow$ & r$L_2\downarrow$ \\
\midrule
Continuous & 0.022 ± 0.001 & 0.037 ± 0.001 & 0.080 ± 0.002 & 0.133 ± 0.004\\\midrule
FSQ & 0.062 ± 0.002 & 0.095 ± 0.002 & 0.224 ± 0.006 & 0.342 ± 0.007 \\
Phaedra$_4$ & 0.038 ± 0.001 & 0.063 ± 0.002 & 0.138 ± 0.004 & 0.226 ± 0.007 \\
Phaedra$_8$ & 0.121 ± 0.004 & 0.192 ± 0.006 & 0.437 ± 0.015 & 0.690 ± 0.022 \\
\bottomrule
\end{tabular}
\end{table}

Overall, these results reinforce Phaedra’s suitability for discretizing \textit{unseen} smoothly varying yet dynamically rich physical systems, extending its effectiveness beyond Earth‑surface observations to atmospheric dynamics.

\subsection{3D Tokenizer Pilot}
As a brief investigation into whether the Phaedra tokenizer may be extended into 3D, we run a minimal pilot. In this experiment, we tokenize a 3D version of the CEU Riemann Curved dataset using a 3D extension of our 2D CNN-based autoencoder. An FSQ-only tokenizer is trained as a baseline. The input resolution is $128^3$ and the tokenized resolution is $32^3$. Results are presented in Table \ref{tab:3d pilot experiment}.

\begin{table}[]
    \centering
    \caption{Relative $L_1$ error for the 3D CEU RC reconstructions.}
    \begin{tabular}{lccc}
        Model & Density & Velocity (avg.) & Pressure \\
        \midrule
        FSQ & 0.98 & 5.94 & 0.79 \\
        Phaedra & \textbf{0.52} & $2.81$ & 0.37 
    \end{tabular}
    \label{tab:3d pilot experiment}
\end{table}

\clearpage
\section{Reconstruction Figures}
\label{sec: reconstruction figures}
\subsection{ID Samples}
\label{sec: id low resolution figures samples}
\begin{figure}[htbp]
     \centering
     \begin{subfigure}[b]{0.3\textwidth}
         \centering
         \includegraphics[width=\textwidth]{images/CRP_rho/Input_fields.png}
         \caption{Ground truth}
     \end{subfigure}
     \hfill
     \begin{subfigure}[b]{0.3\textwidth}
         \centering
         \includegraphics[width=\textwidth]{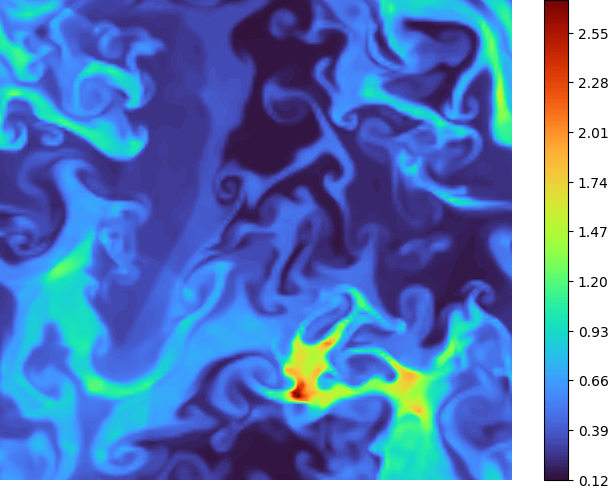}
         \caption{Cont. AE, nMAE: 2.5666}
     \end{subfigure}
     \hfill
     \begin{subfigure}[b]{0.3\textwidth}
         \centering
         \includegraphics[width=\textwidth]{images/CRP_rho/Phaedra_AE_FSQ_4x4_CEU_2D_RiemannCurvedLowRes_rho_recon_fields.png}
         \caption{Phaedra, nMAE: 5.5578}
     \end{subfigure}

     \vspace{10pt} % Add vertical space between rows

     \begin{subfigure}[b]{0.3\textwidth}
         \centering
         \includegraphics[width=\textwidth]{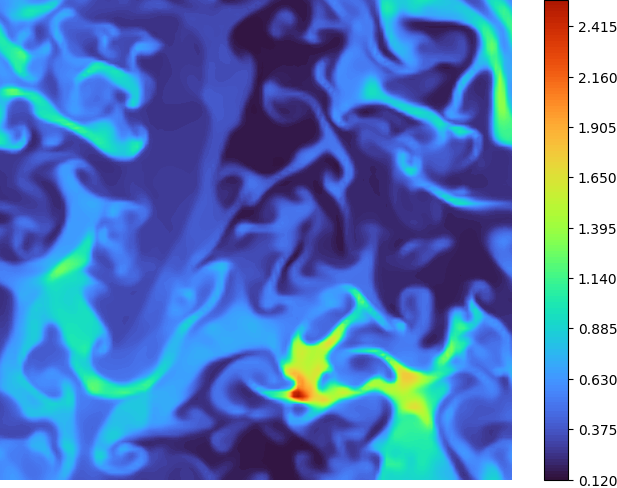}
         \caption{FSQ, nMAE: 8.5492}
     \end{subfigure}
     \hfill
     \begin{subfigure}[b]{0.3\textwidth}
         \centering
         \includegraphics[width=\textwidth]{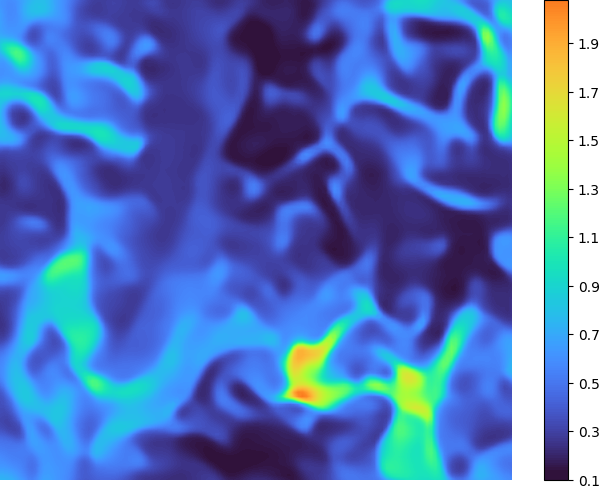}
         \caption{IBQ, nMAE: 19.5397}
     \end{subfigure}
     \hfill
     \begin{subfigure}[b]{0.3\textwidth}
         \centering
         \includegraphics[width=\textwidth]{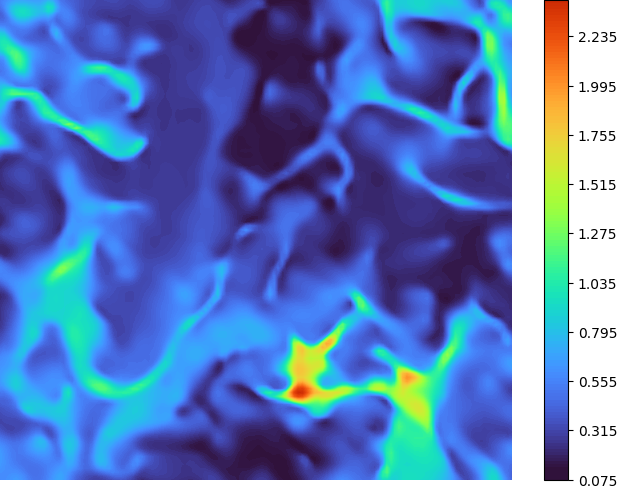}
         \caption{VQ-VAE-2, nMAE: 14.4420}
     \end{subfigure}

     \vspace{10pt}

     \begin{subfigure}[b]{0.3\textwidth}
         \centering
         \includegraphics[width=\textwidth]{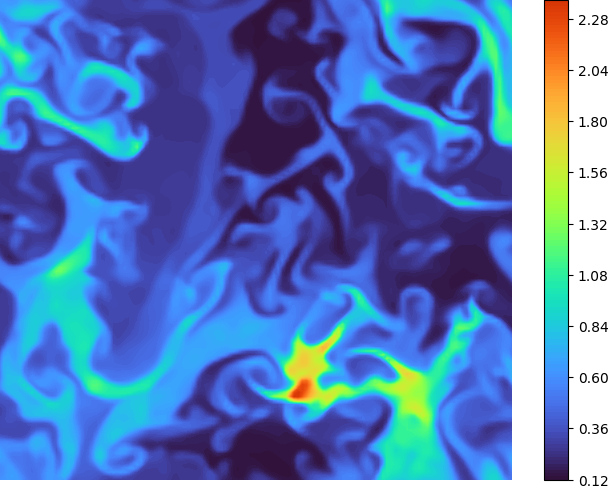}
         \caption{VAR$_{large}$, nMAE: 9.6958}
     \end{subfigure}
     \hfill
     \begin{subfigure}[b]{0.3\textwidth}
         \centering
         \includegraphics[width=\textwidth]{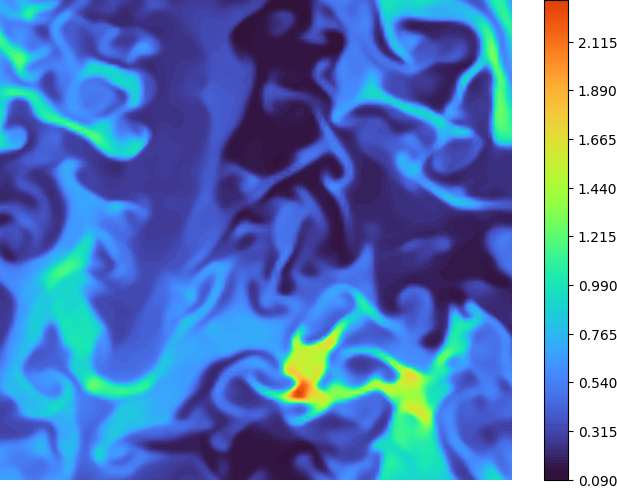}
         \caption{VAR$_{small}$, nMAE: 11.6758}
     \end{subfigure}
     \hfill
     \begin{subfigure}[b]{0.3\textwidth}
         \centering
         \includegraphics[width=\textwidth]{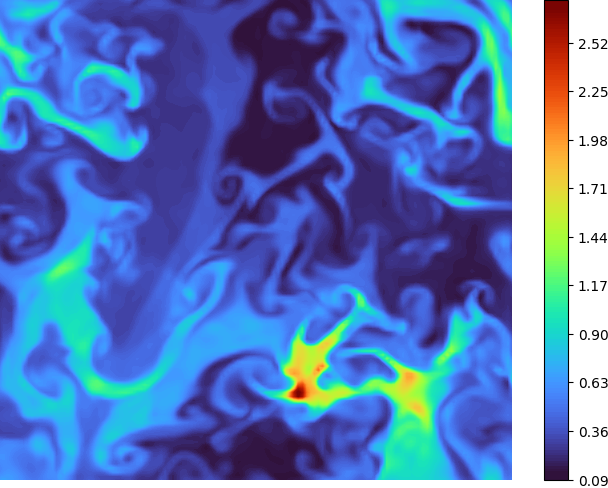}
         \caption{Residual Ablation, nMAE: 7.0496}
     \end{subfigure}

     \caption{\textbf{CEU RC, $\rho$}: Ground truth and reconstructions for density at the final timestep in the first trajectory of the CEU RC dataset.}
     \label{fig:CEU RC rho reconstructions}
\end{figure}

\begin{figure}[htbp]
     \centering
     \begin{subfigure}[b]{0.3\textwidth}
         \centering
         \includegraphics[width=\textwidth]{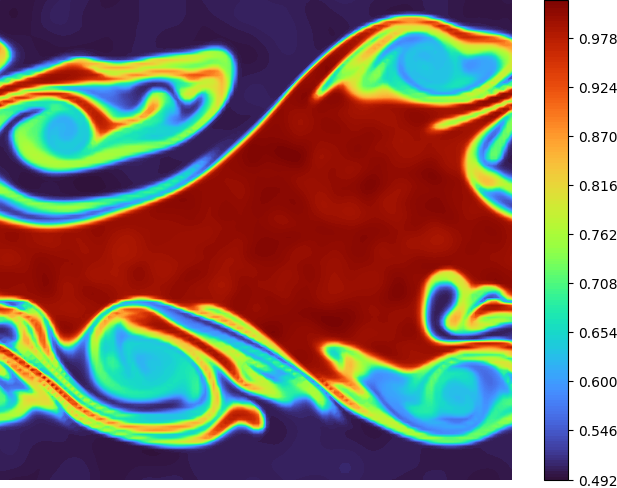}
         \caption{Ground truth}
     \end{subfigure}
     \hfill
     \begin{subfigure}[b]{0.3\textwidth}
         \centering
         \includegraphics[width=\textwidth]{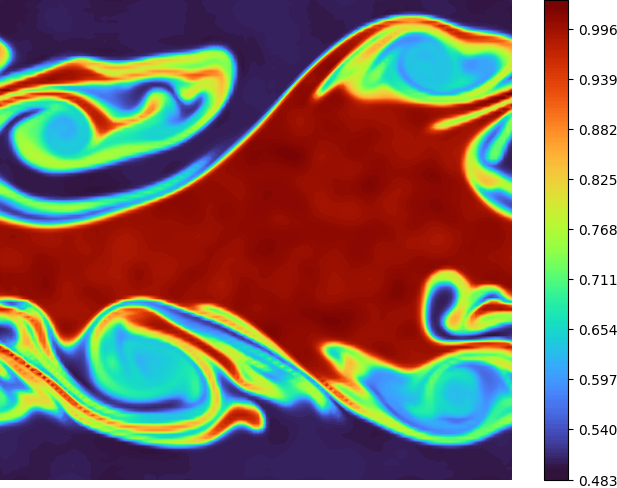}
         \caption{Cont. AE, nMAE: 0.8552}
     \end{subfigure}
     \hfill
     \begin{subfigure}[b]{0.3\textwidth}
         \centering
         \includegraphics[width=\textwidth]{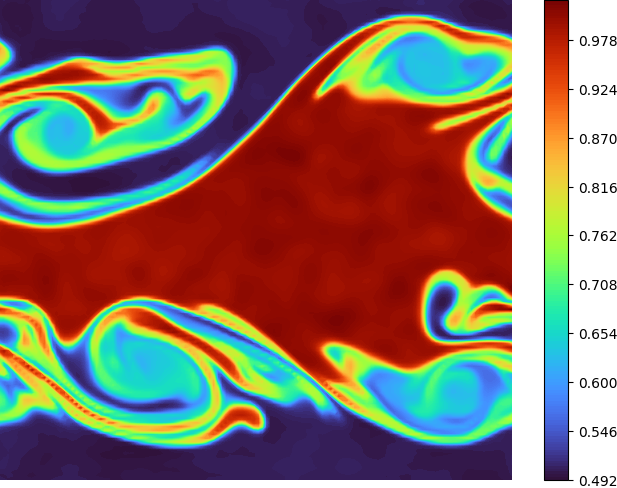}
         \caption{Phaedra, nMAE: 2.0551}
     \end{subfigure}

     \vspace{10pt} % Add vertical space between rows

     \begin{subfigure}[b]{0.3\textwidth}
         \centering
         \includegraphics[width=\textwidth]{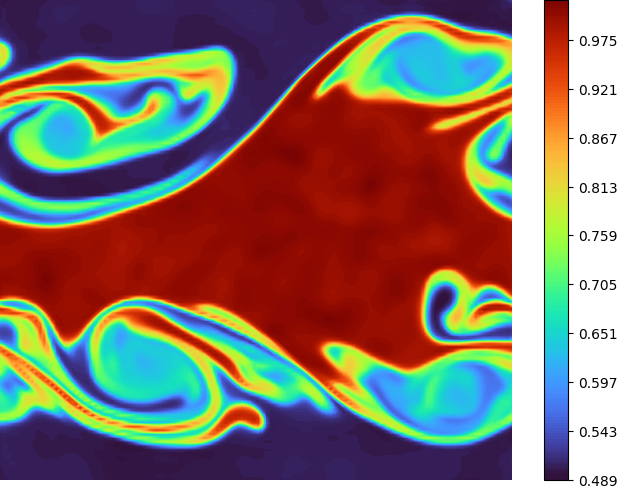}
         \caption{FSQ, nMAE: 3.3900}
     \end{subfigure}
     \hfill
     \begin{subfigure}[b]{0.3\textwidth}
         \centering
         \includegraphics[width=\textwidth]{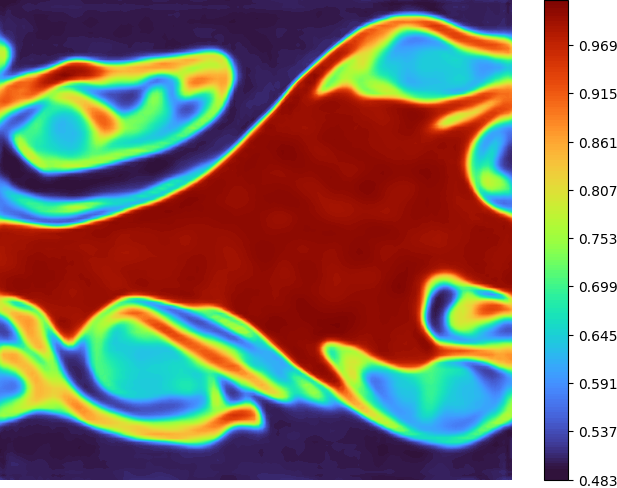}
         \caption{IBQ, nMAE: 9.6695}
     \end{subfigure}
     \hfill
     \begin{subfigure}[b]{0.3\textwidth}
         \centering
         \includegraphics[width=\textwidth]{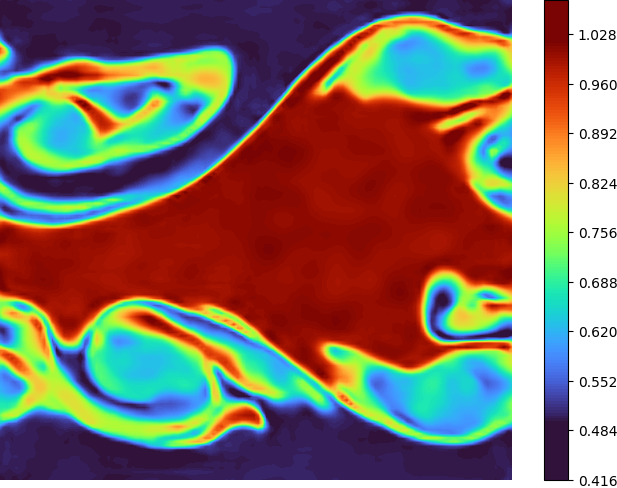}
         \caption{VQ-VAE-2, nMAE: 6.6353}
     \end{subfigure}

     \vspace{10pt}

     \begin{subfigure}[b]{0.3\textwidth}
         \centering
         \includegraphics[width=\textwidth]{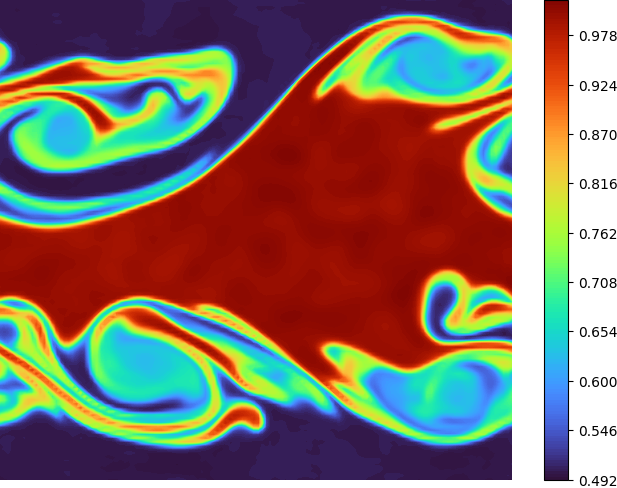}
         \caption{VAR$_{large}$, nMAE: 3.7847}
     \end{subfigure}
     \hfill
     \begin{subfigure}[b]{0.3\textwidth}
         \centering
         \includegraphics[width=\textwidth]{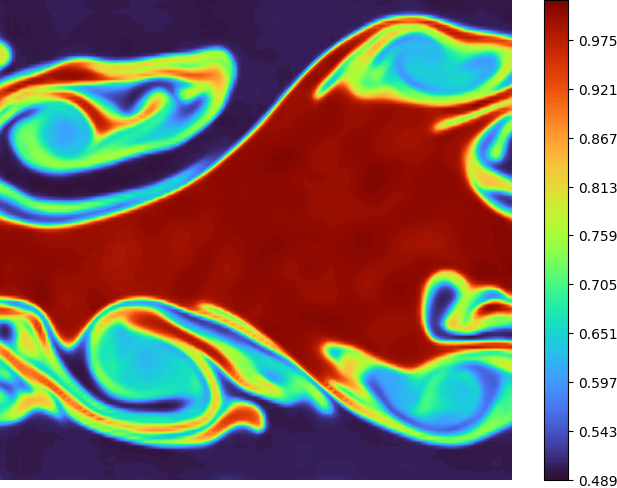}
         \caption{VAR$_{small}$, nMAE: 4.6899}
     \end{subfigure}
     \hfill
     \begin{subfigure}[b]{0.3\textwidth}
         \centering
         \includegraphics[width=\textwidth]{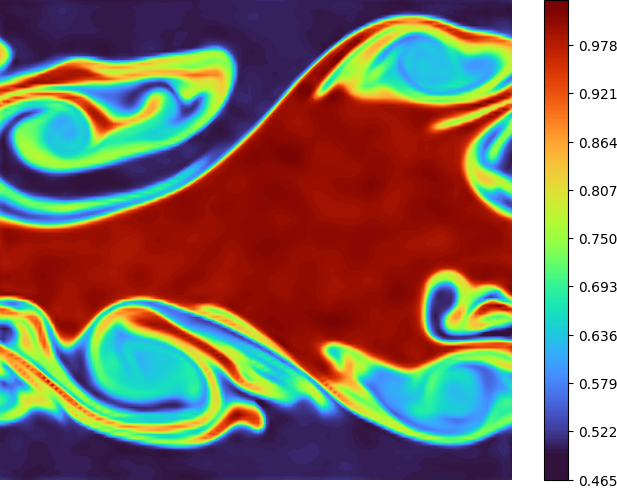}
         \caption{Residual Ablation, nMAE: 2.7994}
     \end{subfigure}

     \caption{\textbf{CEU KH, $\rho$}: Ground truth and reconstructions for density at the final timestep in the first trajectory of the CEU KH dataset.}
     \label{fig:CEU KH rho reconstructions}
\end{figure}

\begin{figure}[htbp]
     \centering
     \begin{subfigure}[b]{0.3\textwidth}
         \centering
         \includegraphics[width=\textwidth]{images/KH_p/Input_fields.png}
         \caption{Ground truth}
     \end{subfigure}
     \hfill
     \begin{subfigure}[b]{0.3\textwidth}
         \centering
         \includegraphics[width=\textwidth]{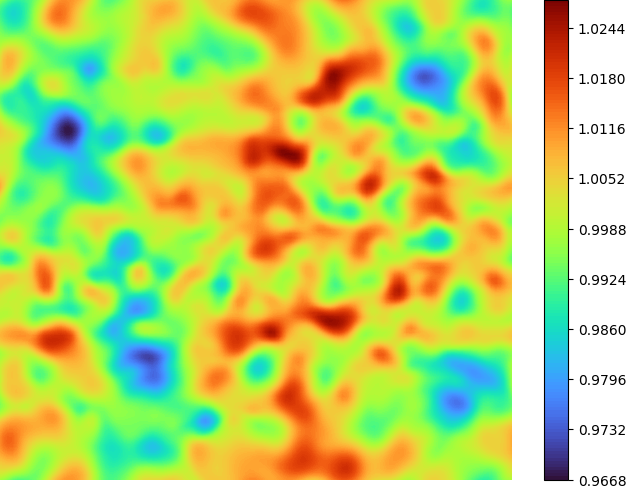}
         \caption{Cont. AE, nMAE: 0.4086}
     \end{subfigure}
     \hfill
     \begin{subfigure}[b]{0.3\textwidth}
         \centering
         \includegraphics[width=\textwidth]{images/KH_p/Phaedra_AE_FSQ_4x4_CEU_2D_KelvinHelmholtzLowRes_p_recon_fields.png}
         \caption{Phaedra, nMAE: 1.3366}
     \end{subfigure}

     \vspace{10pt} % Add vertical space between rows

     \begin{subfigure}[b]{0.3\textwidth}
         \centering
         \includegraphics[width=\textwidth]{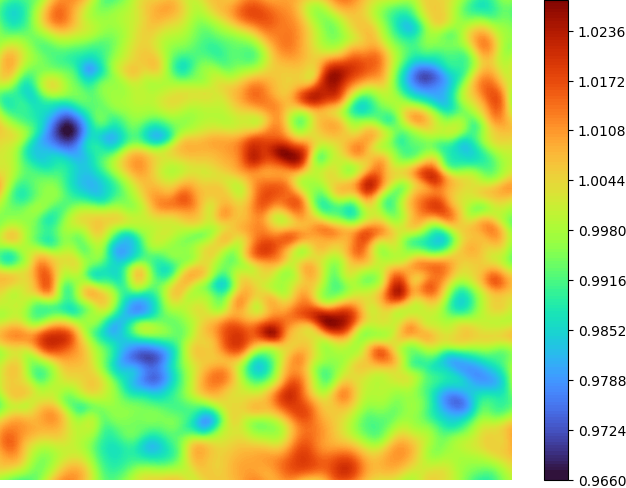}
         \caption{FSQ, nMAE: 2.9358}
     \end{subfigure}
     \hfill
     \begin{subfigure}[b]{0.3\textwidth}
         \centering
         \includegraphics[width=\textwidth]{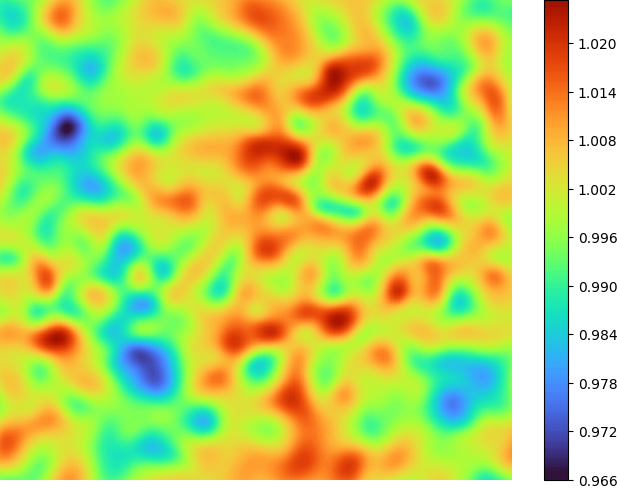}
         \caption{IBQ, nMAE: 12.0917}
     \end{subfigure}
     \hfill
     \begin{subfigure}[b]{0.3\textwidth}
         \centering
         \includegraphics[width=\textwidth]{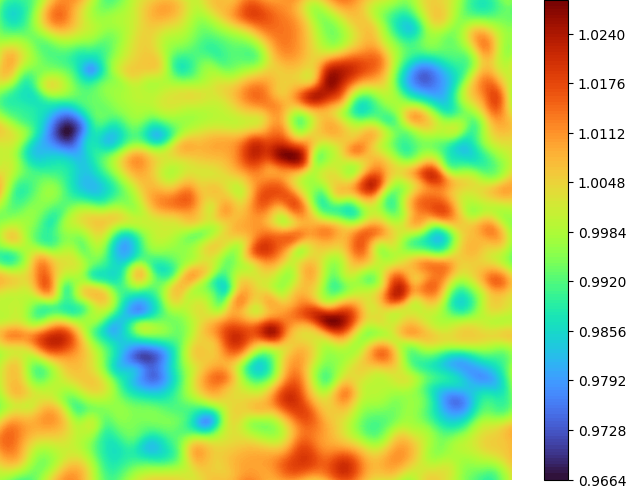}
         \caption{VQ-VAE-2, nMAE: 4.3613}
     \end{subfigure}

     \vspace{10pt}

     \begin{subfigure}[b]{0.3\textwidth}
         \centering
         \includegraphics[width=\textwidth]{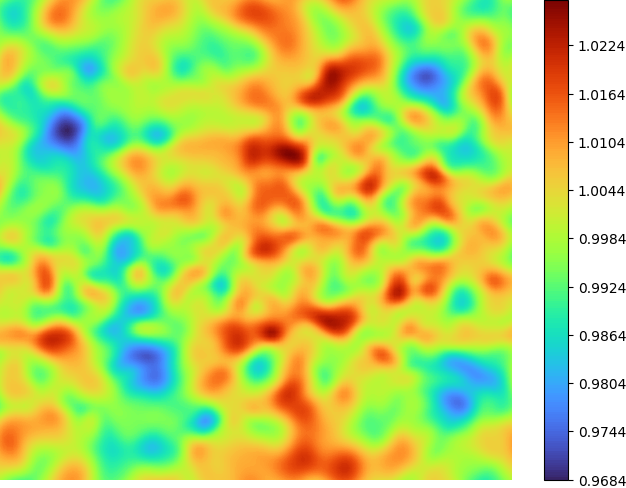}
         \caption{VAR$_{large}$, nMAE: 4.0569}
     \end{subfigure}
     \hfill
     \begin{subfigure}[b]{0.3\textwidth}
         \centering
         \includegraphics[width=\textwidth]{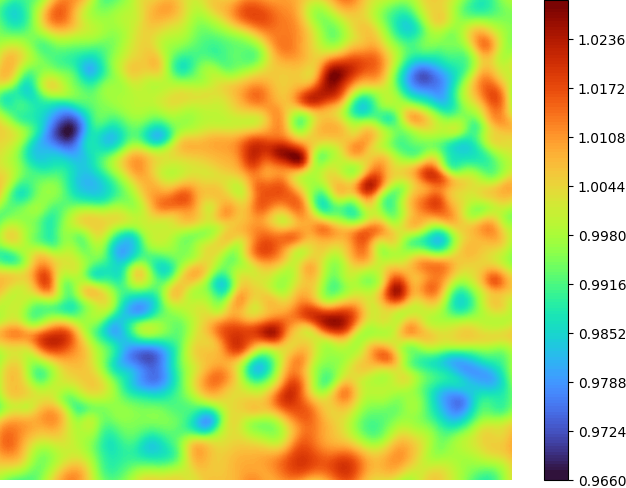}
         \caption{VAR$_{small}$, nMAE: 5.8759}
     \end{subfigure}
     \hfill
     \begin{subfigure}[b]{0.3\textwidth}
         \centering
         \includegraphics[width=\textwidth]{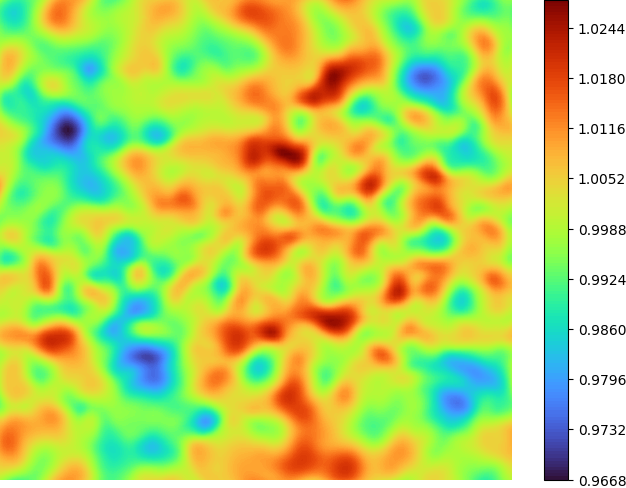}
         \caption{Residual Ablation, nMAE: 1.6057}
     \end{subfigure}

     \caption{\textbf{CEU KH, $p$}: Ground truth and reconstructions for pressure at the final timestep in the first trajectory of the CEU KH dataset.}
     \label{fig:CEU KH pressure reconstructions}
\end{figure}

\clearpage
\subsection{OD$_1$ Samples}
\label{sec: od1 low resolution figures samples}
\begin{figure}[htbp]
     \centering
     \begin{subfigure}[b]{0.3\textwidth}
         \centering
         \includegraphics[width=\textwidth]{images/Airfoil/Input_fields.png}
         \caption{Ground truth}
     \end{subfigure}
     \hfill
     \begin{subfigure}[b]{0.3\textwidth}
         \centering
         \includegraphics[width=\textwidth]{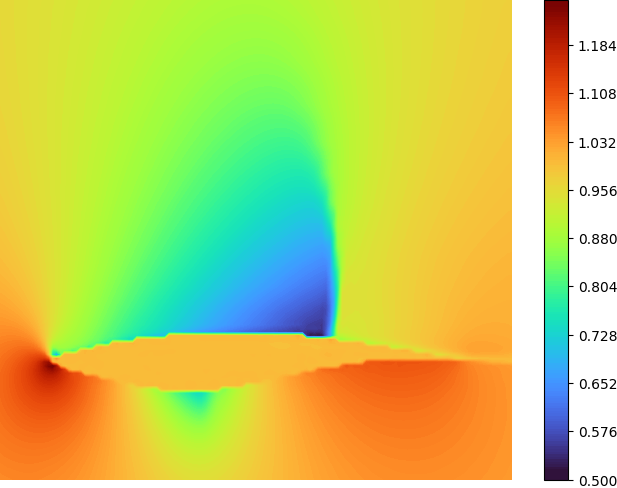}
         \caption{Cont. AE, nMAE: 0.4264}
     \end{subfigure}
     \hfill
     \begin{subfigure}[b]{0.3\textwidth}
         \centering
         \includegraphics[width=\textwidth]{images/Airfoil/Phaedra_AE_FSQ_4x4_CEU_2D_RPBAirfoil_rho_recon_fields.png}
         \caption{Phaedra, nMAE: 1.0802}
     \end{subfigure}

     \vspace{10pt} % Add vertical space between rows

     \begin{subfigure}[b]{0.3\textwidth}
         \centering
         \includegraphics[width=\textwidth]{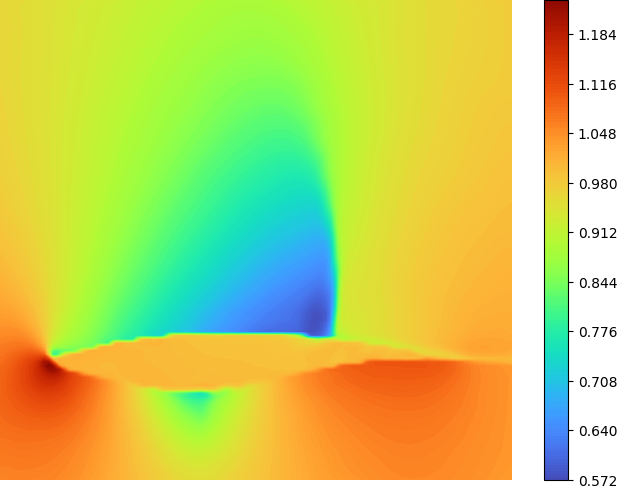}
         \caption{FSQ, nMAE: 2.2186}
     \end{subfigure}
     \hfill
     \begin{subfigure}[b]{0.3\textwidth}
         \centering
         \includegraphics[width=\textwidth]{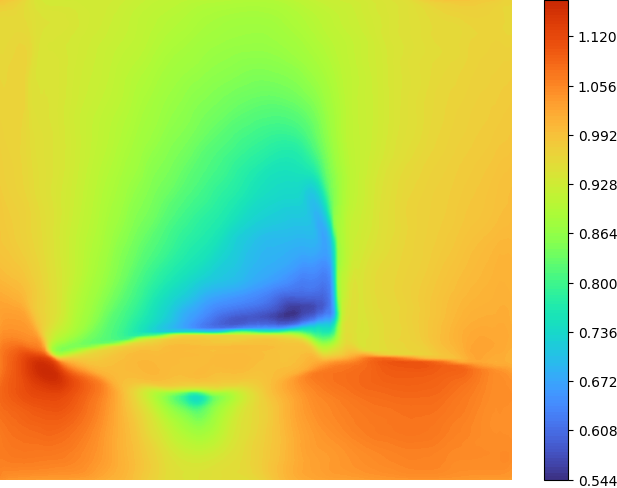}
         \caption{IBQ, nMAE: 6.7628}
     \end{subfigure}
     \hfill
     \begin{subfigure}[b]{0.3\textwidth}
         \centering
         \includegraphics[width=\textwidth]{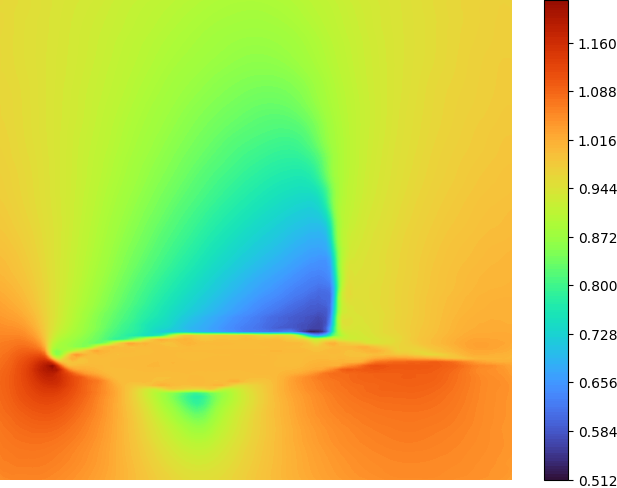}
         \caption{VQ-VAE-2, nMAE: 1.7547}
     \end{subfigure}

     \vspace{10pt}

     \begin{subfigure}[b]{0.3\textwidth}
         \centering
         \includegraphics[width=\textwidth]{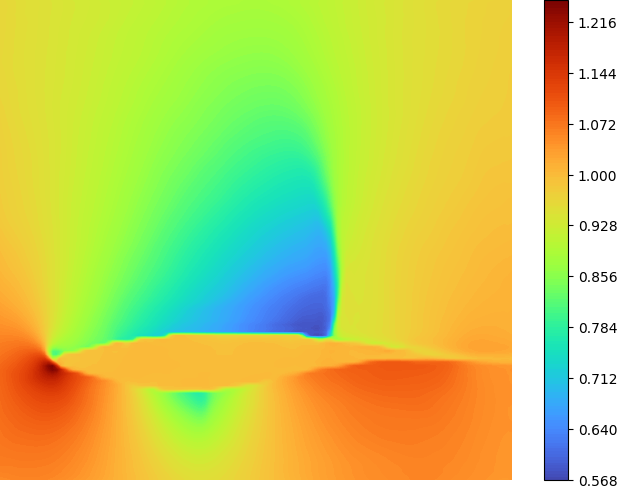}
         \caption{VAR$_{large}$, nMAE: 2.0221}
     \end{subfigure}
     \hfill
     \begin{subfigure}[b]{0.3\textwidth}
         \centering
         \includegraphics[width=\textwidth]{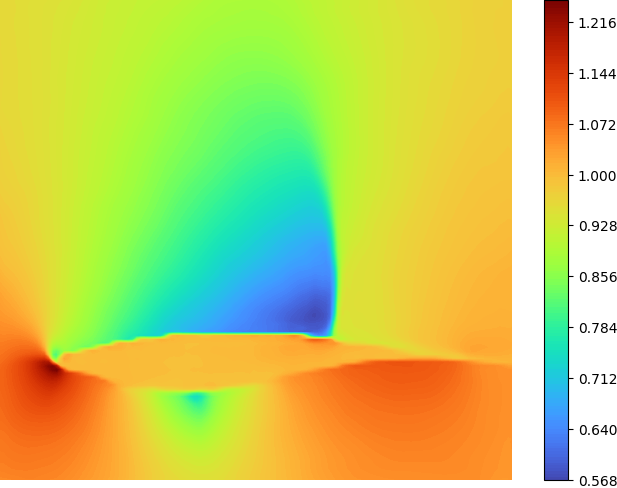}
         \caption{VAR$_{small}$, nMAE: 2.4270}
     \end{subfigure}
     \hfill
     \begin{subfigure}[b]{0.3\textwidth}
         \centering
         \includegraphics[width=\textwidth]{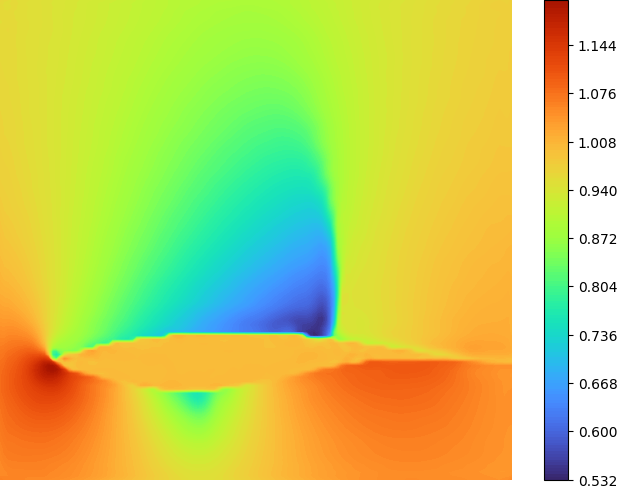}
         \caption{Residual Ablation, nMAE: 1.0952}
     \end{subfigure}

     \caption{\textbf{CEU AIR, $\rho$}: Ground truth and reconstructions for density in the first sample of the CEU Airfoil dataset.}
     \label{fig:CEU air reconstructions}
\end{figure}

\clearpage
\subsection{OD$_2$ Samples}
\label{sec: od2 low resolution figures samples}
\begin{figure}[htbp]
     \centering
     \begin{subfigure}[b]{0.3\textwidth}
         \centering
         \includegraphics[width=\textwidth]{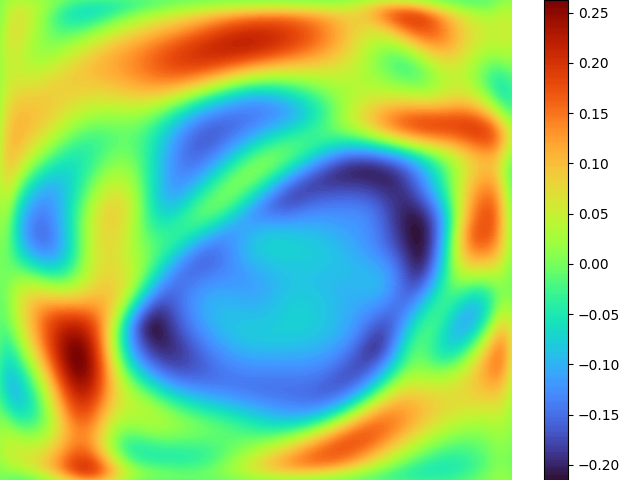}
         \caption{Ground truth}
     \end{subfigure}
     \hfill
     \begin{subfigure}[b]{0.3\textwidth}
         \centering
         \includegraphics[width=\textwidth]{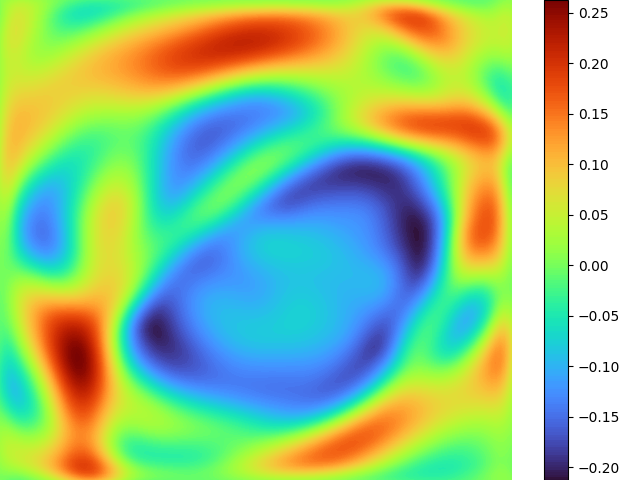}
         \caption{Cont. AE, nMAE: 0.3242}
     \end{subfigure}
     \hfill
     \begin{subfigure}[b]{0.3\textwidth}
         \centering
         \includegraphics[width=\textwidth]{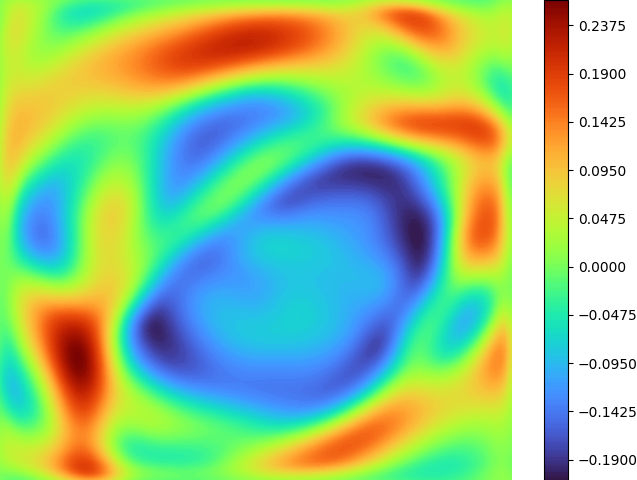}
         \caption{Phaedra, nMAE: 0.9231}
     \end{subfigure}

     \vspace{10pt} % Add vertical space between rows

     \begin{subfigure}[b]{0.3\textwidth}
         \centering
         \includegraphics[width=\textwidth]{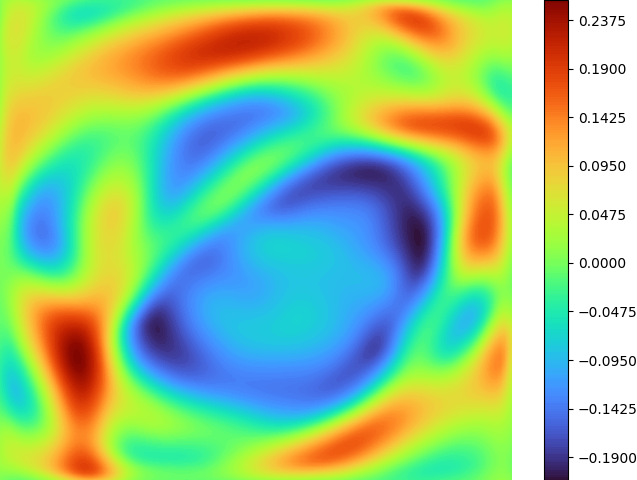}
         \caption{FSQ, nMAE: 1.2219}
     \end{subfigure}
     \hfill
     \begin{subfigure}[b]{0.3\textwidth}
         \centering
         \includegraphics[width=\textwidth]{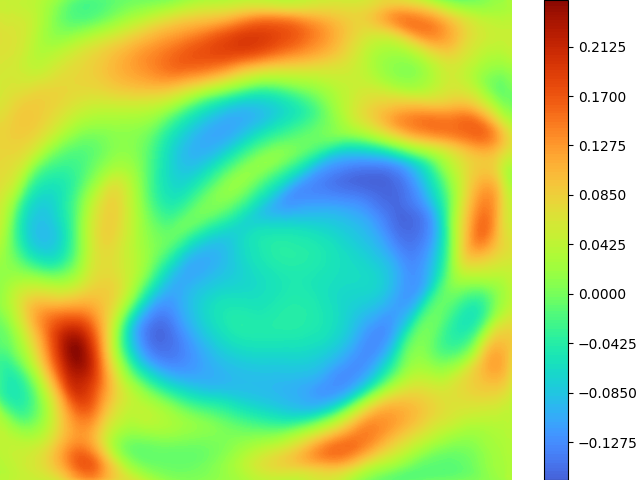}
         \caption{IBQ, nMAE: 21.3735}
     \end{subfigure}
     \hfill
     \begin{subfigure}[b]{0.3\textwidth}
         \centering
         \includegraphics[width=\textwidth]{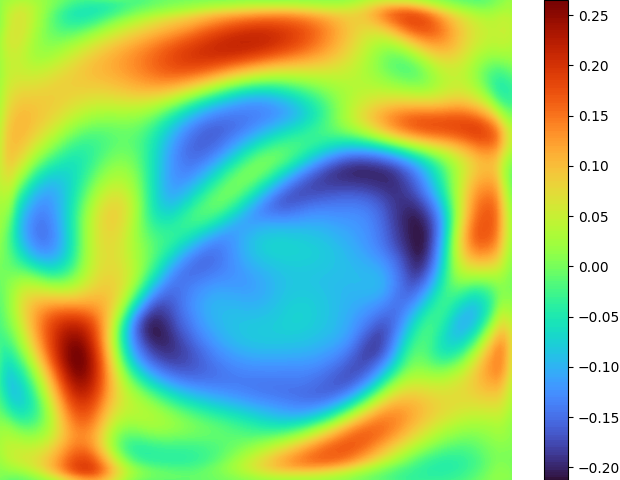}
         \caption{VQ-VAE-2, nMAE: 1.0513}
     \end{subfigure}

     \vspace{10pt}

     \begin{subfigure}[b]{0.3\textwidth}
         \centering
         \includegraphics[width=\textwidth]{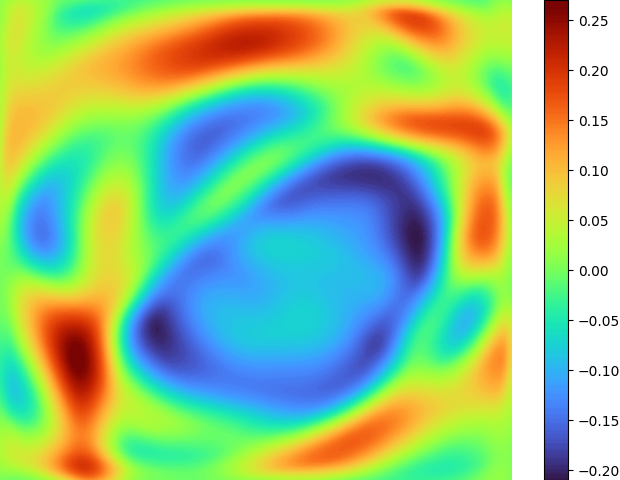}
         \caption{VAR$_{large}$, nMAE: 2.3942}
     \end{subfigure}
     \hfill
     \begin{subfigure}[b]{0.3\textwidth}
         \centering
         \includegraphics[width=\textwidth]{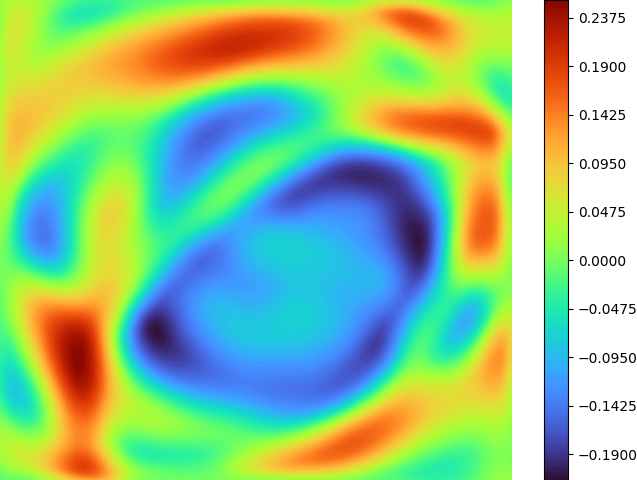}
         \caption{VAR$_{small}$, nMAE: 1.7859}
     \end{subfigure}
     \hfill
     \begin{subfigure}[b]{0.3\textwidth}
         \centering
         \includegraphics[width=\textwidth]{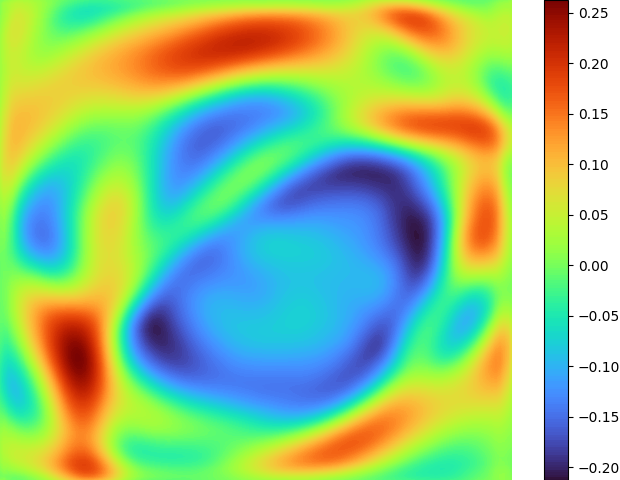}
         \caption{Residual Ablation, nMAE: 0.4606}
     \end{subfigure}

     \caption{\textbf{AWA}: Ground truth and reconstructions for the solution at the final timestep in the first trajectory of the Acoustic Wave dataset.}
     \label{fig:AWA reconstructions}
\end{figure}

\clearpage
\subsection{High-Resolution Samples}
\label{sec: high resolution figures samples}
\begin{figure}[htbp]
     \centering
     \begin{subfigure}[b]{0.3\textwidth}
         \centering
         \includegraphics[width=\textwidth]{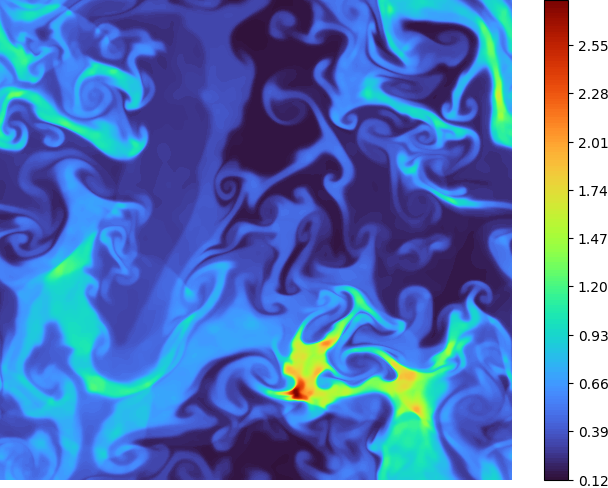}
         \caption{Ground truth}
     \end{subfigure}
     \hfill

     \vspace{10pt} % Add vertical space between rows

     \begin{subfigure}[b]{0.3\textwidth}
         \centering
         \includegraphics[width=\textwidth]{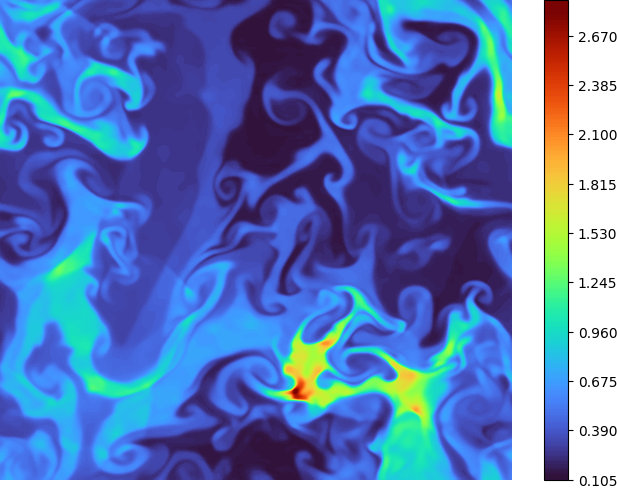}
         \caption{Phaedra$_8$, nMAE: 2.7239}
     \end{subfigure}
     \hfill
     \begin{subfigure}[b]{0.3\textwidth}
         \centering
         \includegraphics[width=\textwidth]{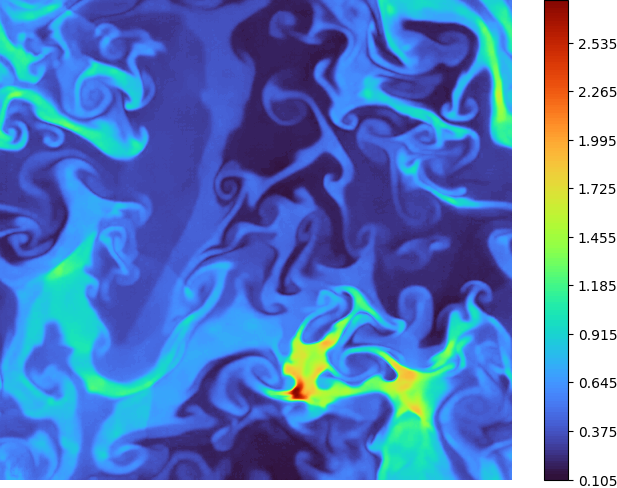}
         \caption{Cosmos$_8^{0:1}$, nMAE: 8.5508}
     \end{subfigure}
     \hfill
     \begin{subfigure}[b]{0.3\textwidth}
         \centering
         \includegraphics[width=\textwidth]{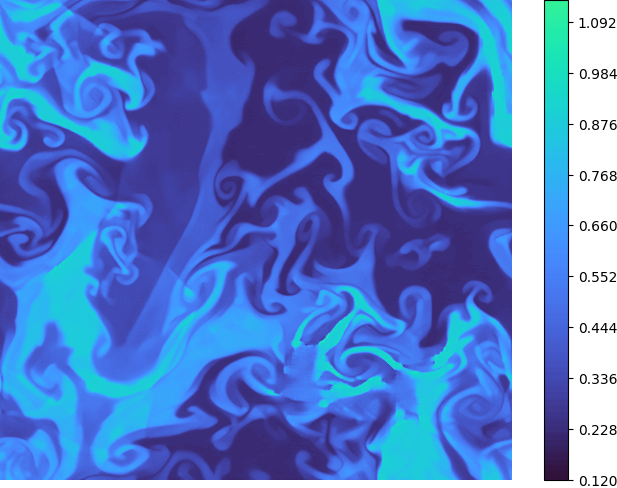}
         \caption{Cosmos$_8$, nMAE: 17.4769}
     \end{subfigure}

     \vspace{10pt}

     \begin{subfigure}[b]{0.3\textwidth}
         \centering
         \includegraphics[width=\textwidth]{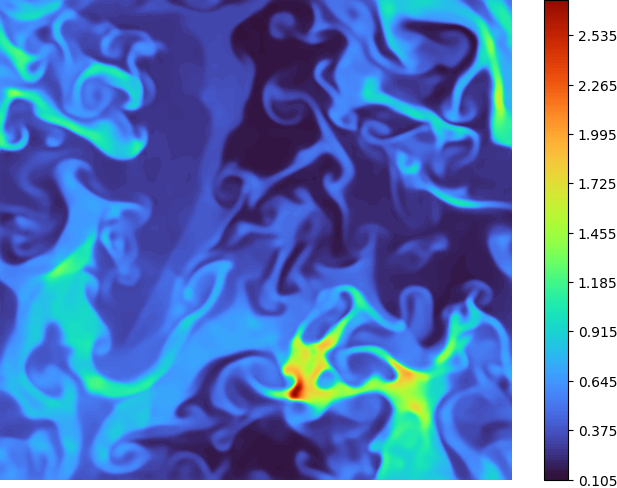}
         \caption{Phaedra$_{16}$, nMAE: 7.1981}
     \end{subfigure}
     \hfill
     \begin{subfigure}[b]{0.3\textwidth}
         \centering
         \includegraphics[width=\textwidth]{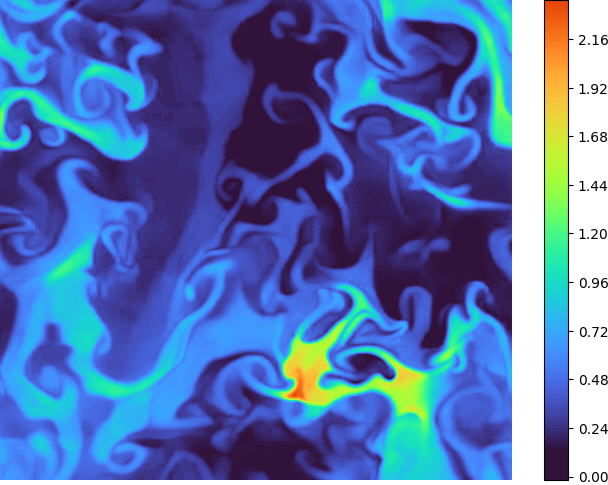}
         \caption{Cosmos$_{16}^{0:1}$, nMAE: 19.2073}
     \end{subfigure}
     \hfill
     \begin{subfigure}[b]{0.3\textwidth}
         \centering
         \includegraphics[width=\textwidth]{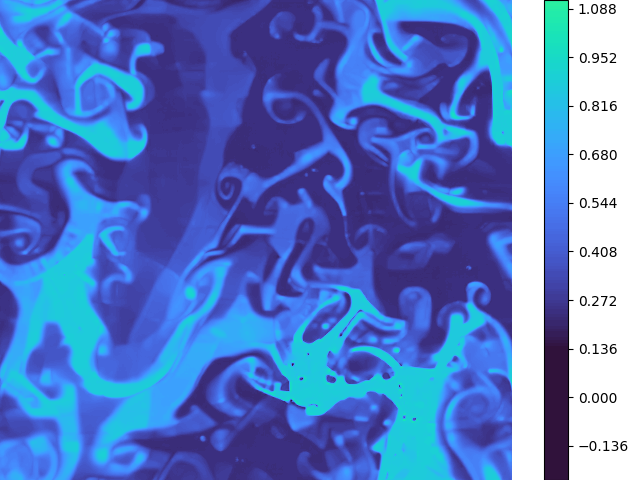}
         \caption{Cosmos$_{16}$, nMAE: 21.7191}
     \end{subfigure}

     \caption{\textbf{CEU RC$_{512}$, $\rho$}: Ground truth and reconstructions for \emph{density} at the final timestep in the first trajectory of the high-resolution CEU RC dataset.}
     \label{fig:CEU RC 512 rho reconstructions}
\end{figure}

\begin{figure}[htbp]
     \centering
     \begin{subfigure}[b]{0.3\textwidth}
         \centering
         \includegraphics[width=\textwidth]{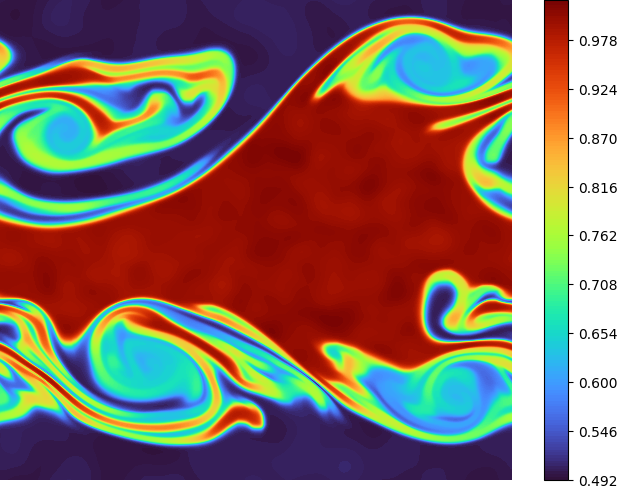}
         \caption{Ground truth}
     \end{subfigure}
     \hfill

     \vspace{10pt} % Add vertical space between rows

     \begin{subfigure}[b]{0.3\textwidth}
         \centering
         \includegraphics[width=\textwidth]{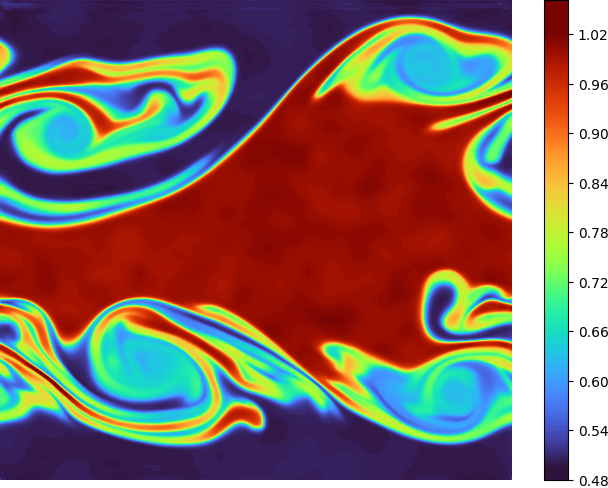}
         \caption{Phaedra$_8$, nMAE: 1.0492}
     \end{subfigure}
     \hfill
     \begin{subfigure}[b]{0.3\textwidth}
         \centering
         \includegraphics[width=\textwidth]{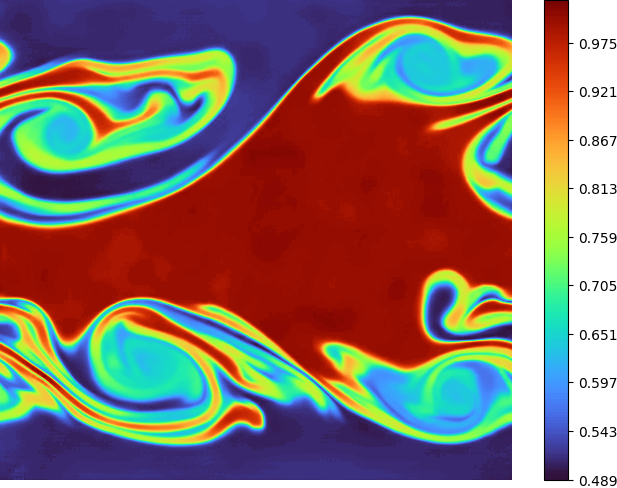}
         \caption{Cosmos$_8^{0:1}$, nMAE: 2.9245}
     \end{subfigure}
     \hfill
     \begin{subfigure}[b]{0.3\textwidth}
         \centering
         \includegraphics[width=\textwidth]{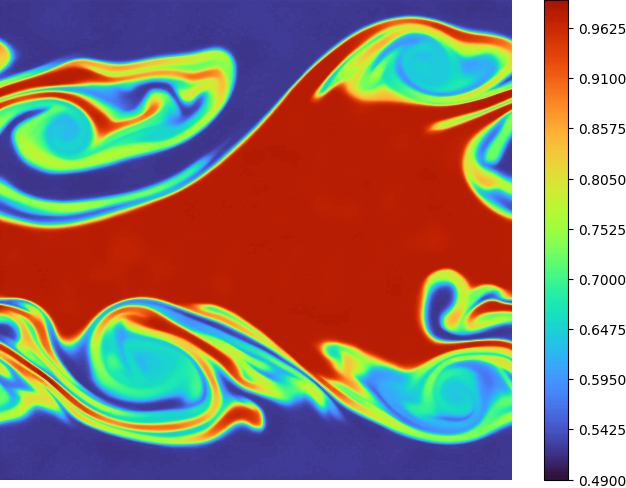}
         \caption{Cosmos$_8$, nMAE: 7.5107}
     \end{subfigure}

     \vspace{10pt}

     \begin{subfigure}[b]{0.3\textwidth}
         \centering
         \includegraphics[width=\textwidth]{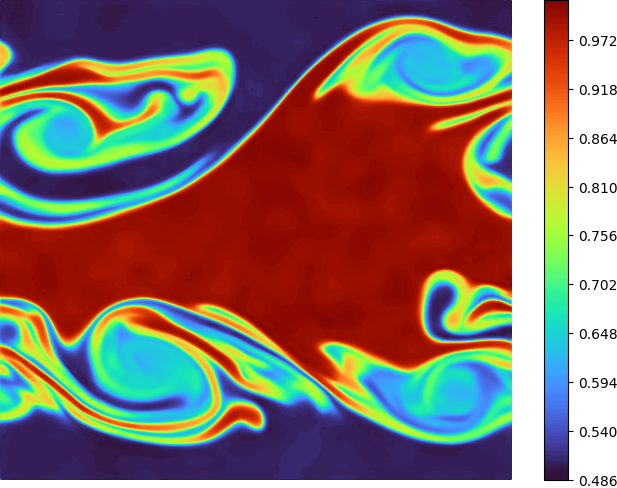}
         \caption{Phaedra$_{16}$, nMAE: 2.7128}
     \end{subfigure}
     \hfill
     \begin{subfigure}[b]{0.3\textwidth}
         \centering
         \includegraphics[width=\textwidth]{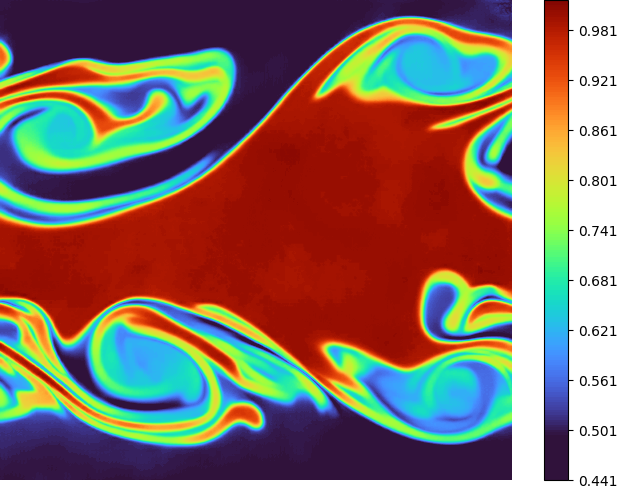}
         \caption{Cosmos$_{16}^{0:1}$, nMAE: 7.0264}
     \end{subfigure}
     \hfill
     \begin{subfigure}[b]{0.3\textwidth}
         \centering
         \includegraphics[width=\textwidth]{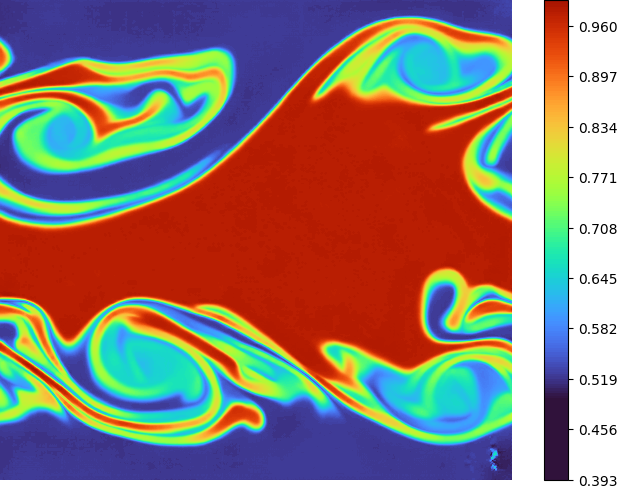}
         \caption{Cosmos$_{16}$, nMAE: 9.5627}
     \end{subfigure}

     \caption{\textbf{CEU KH$_{512}$, $\rho$}: Ground truth and reconstructions for \emph{density} at the final timestep in the first trajectory of the high-resolution CEU KH dataset.}
     \label{fig:CEU KH 512 rho reconstructions}
\end{figure}

\begin{figure}[htbp]
     \centering
     \begin{subfigure}[b]{0.3\textwidth}
         \centering
         \includegraphics[width=\textwidth]{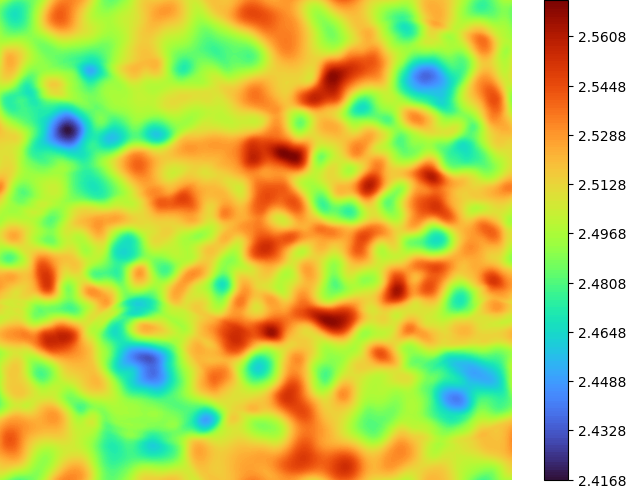}
         \caption{Ground truth}
     \end{subfigure}
     \hfill

     \vspace{10pt} % Add vertical space between rows

     \begin{subfigure}[b]{0.3\textwidth}
         \centering
         \includegraphics[width=\textwidth]{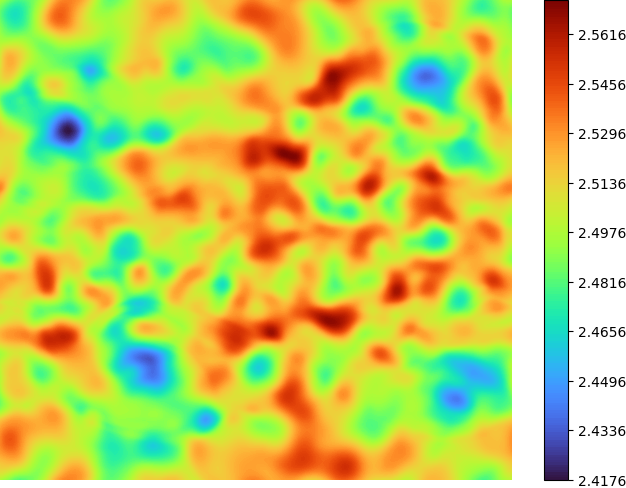}
         \caption{Phaedra$_8$, nMAE: 0.6878}
     \end{subfigure}
     \hfill
     \begin{subfigure}[b]{0.3\textwidth}
         \centering
         \includegraphics[width=\textwidth]{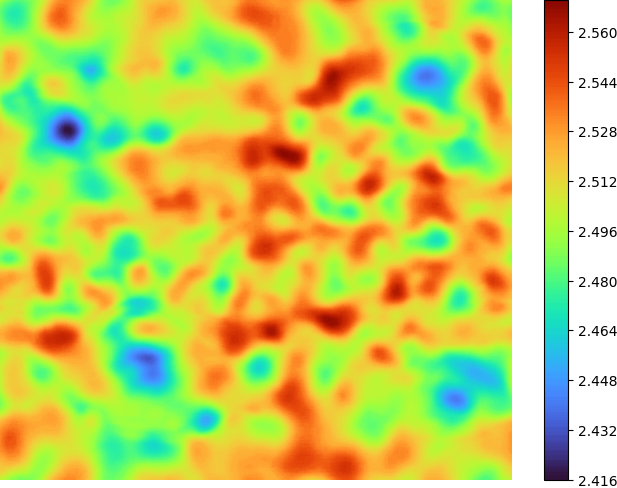}
         \caption{Cosmos$_8^{0:1}$, nMAE: 6.5960}
     \end{subfigure}
     \hfill
     \begin{subfigure}[b]{0.3\textwidth}
         \centering
         \includegraphics[width=\textwidth]{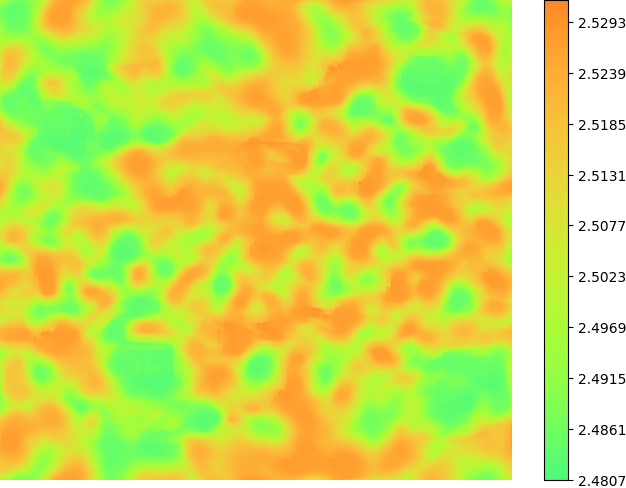}
         \caption{Cosmos$_8$, nMAE: 27.5057}
     \end{subfigure}

     \vspace{10pt}

     \begin{subfigure}[b]{0.3\textwidth}
         \centering
         \includegraphics[width=\textwidth]{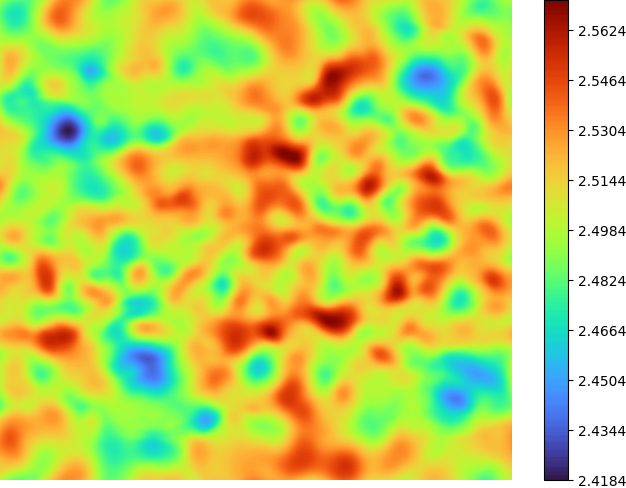}
         \caption{Phaedra$_{16}$, nMAE: 1.8797}
     \end{subfigure}
     \hfill
     \begin{subfigure}[b]{0.3\textwidth}
         \centering
         \includegraphics[width=\textwidth]{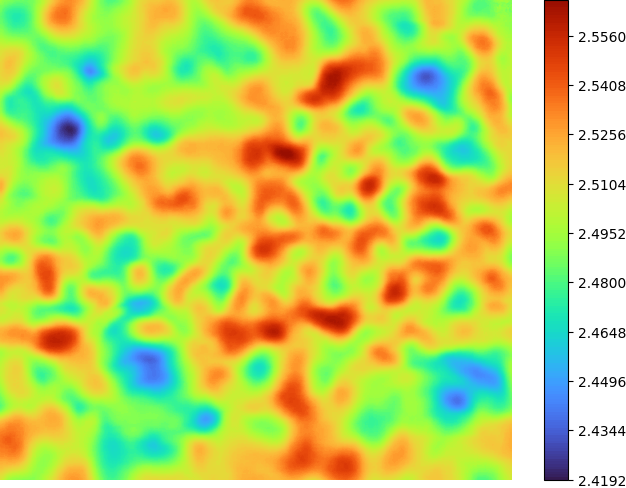}
         \caption{Cosmos$_{16}^{0:1}$, nMAE: 16.7553}
     \end{subfigure}
     \hfill
     \begin{subfigure}[b]{0.3\textwidth}
         \centering
         \includegraphics[width=\textwidth]{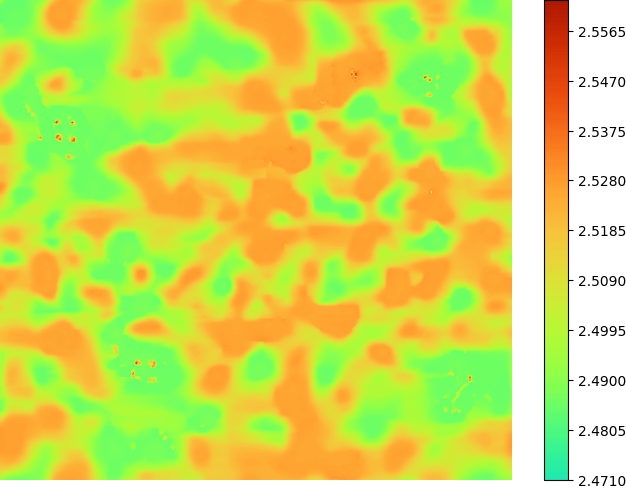}
         \caption{Cosmos$_{16}$, nMAE: 29.4885}
     \end{subfigure}

     \caption{\textbf{CEU KH$_{512}$, $E$}: Ground truth and reconstructions for \emph{energy} at the final timestep in the first trajectory of the high-resolution CEU KH dataset.}
     \label{fig:CEU KH 512 E reconstructions}
\end{figure}

\clearpage
\subsection{Earth Observation Samples}
\label{sec: earth observation eo data figures samples}

\begin{figure}[h]
     \centering
     \begin{subfigure}[b]{0.3\textwidth}
         \centering
         \includegraphics[width=\textwidth]{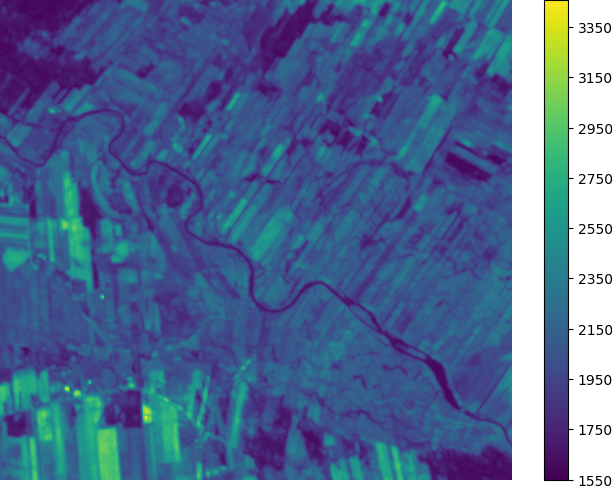}
         \caption{Original}
     \end{subfigure}
     \hspace{0.5cm}
     \begin{subfigure}[b]{0.3\textwidth}
         \centering
         \includegraphics[width=\textwidth]{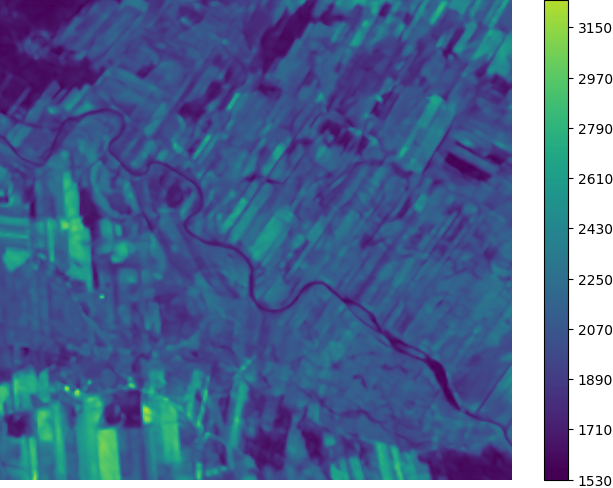}
         \caption{Phaedra$_4$}
     \end{subfigure}
     \hspace{0.5cm}
     \begin{subfigure}[b]{0.3\textwidth}
         \centering
         \includegraphics[width=\textwidth]{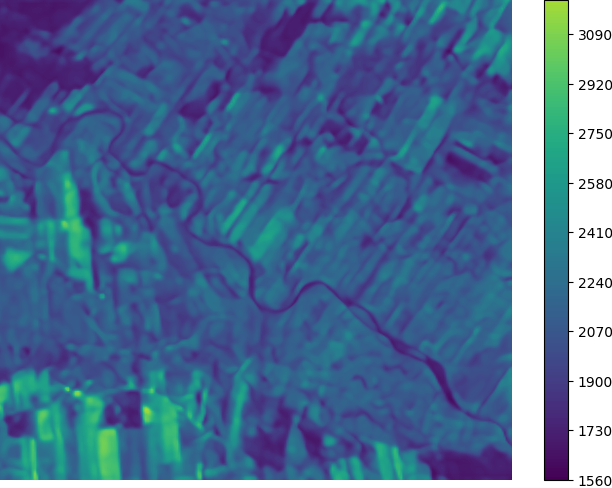}
         \caption{FSQ$_4$}
     \end{subfigure}

     \vspace{10pt}

     \begin{subfigure}[b]{0.3\textwidth}
         \centering
         \includegraphics[width=\textwidth]{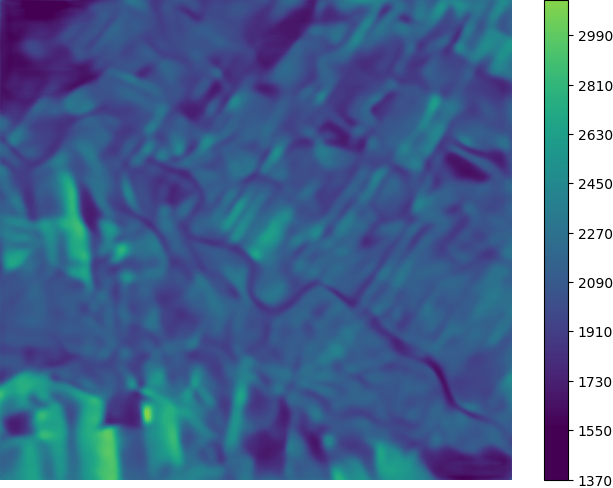}
         \caption{Phaedra$_{8}$}
     \end{subfigure}
     \hspace{0.5cm}
     \begin{subfigure}[b]{0.3\textwidth}
         \centering
         \includegraphics[width=\textwidth]{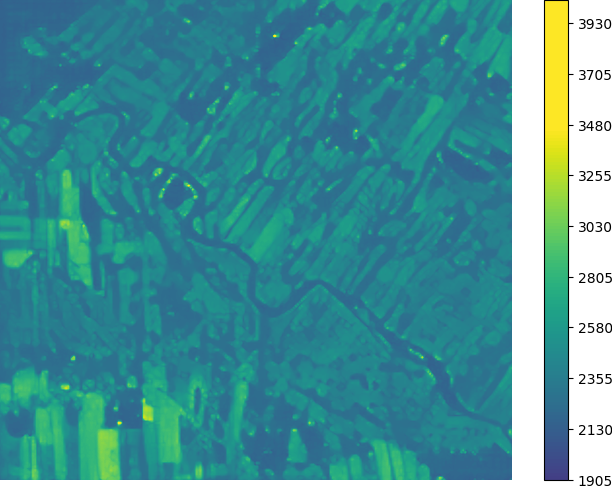}
         \caption{Cosmos$_{8}$}
     \end{subfigure}

     \caption{\textbf{Sentinel-2 L1C-Band 3}: Ground truth and reconstructions for the first sample of Band 3. All models are applied without any fine-tuning on Sentinel-2 or other earth observation data. We use a dataset-wide 0:1 normalization across all models. }
     \label{fig:S2L1C reconstructions}
\end{figure}

In the following, we provide qualitative results for reconstructions on Earth observation data. For reasons of interpretability, we primarily focus on multispectral optical data in this subsection. In Figure~\ref{fig:S2L1C reconstructions}, we provide a visual comparison of the performance of different reconstruction models. We observe that Phaedra is well able to capture spatial details and manages to reconstruct the overall magnitude of the data which represents the reflectance values in the Earth observation domain.

\begin{figure}[h]
    \centering
    \includegraphics[width=0.6\linewidth]{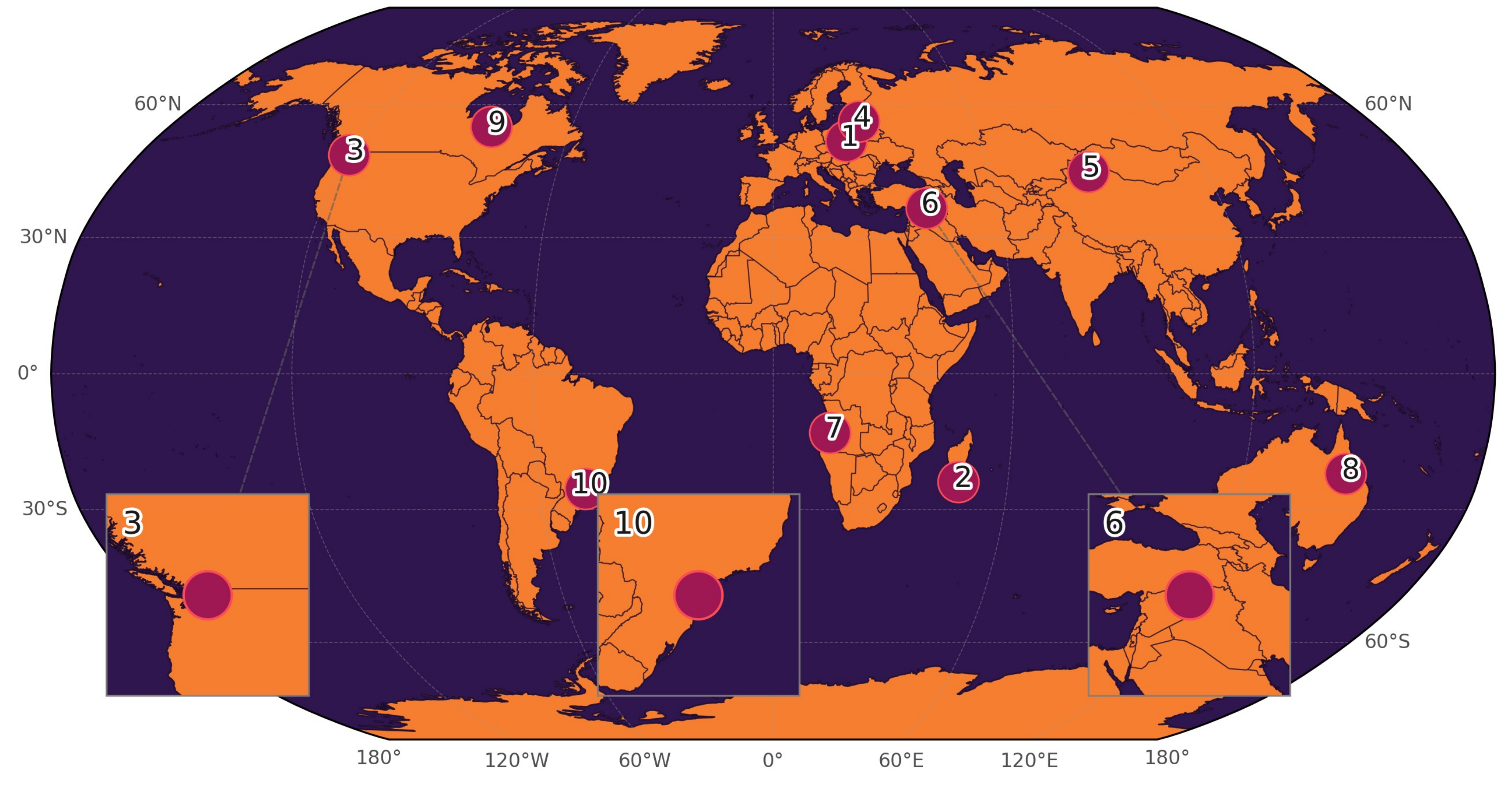}
    \caption{Global positions of the 10 locations reconstructed in Fig.~\ref{fig:s2l1c global reconstructions}.}
    \label{fig:S2L1C diverse sample locations}
\end{figure}

\begin{figure}[h]
    \centering
        \begin{subfigure}[b]{0.24\textwidth}
            \centering
            \includegraphics[width=\textwidth]{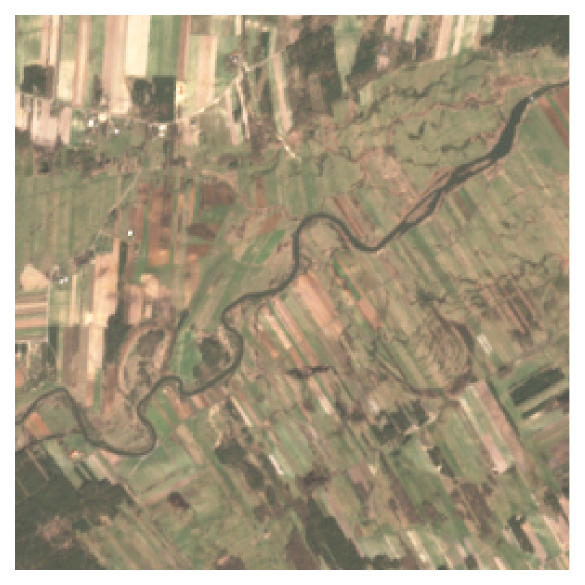}
        \end{subfigure}
        \begin{subfigure}[b]{0.24\textwidth}
            \centering
            \includegraphics[width=\textwidth]{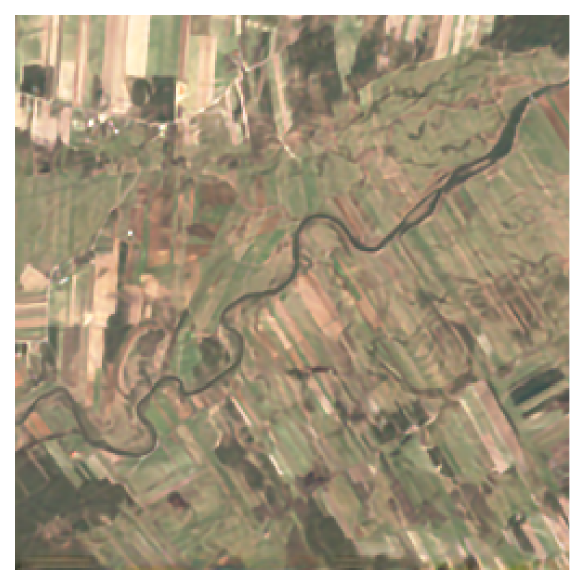}
        \end{subfigure}
        \begin{subfigure}[b]{0.24\textwidth}
            \centering
            \includegraphics[width=\textwidth]{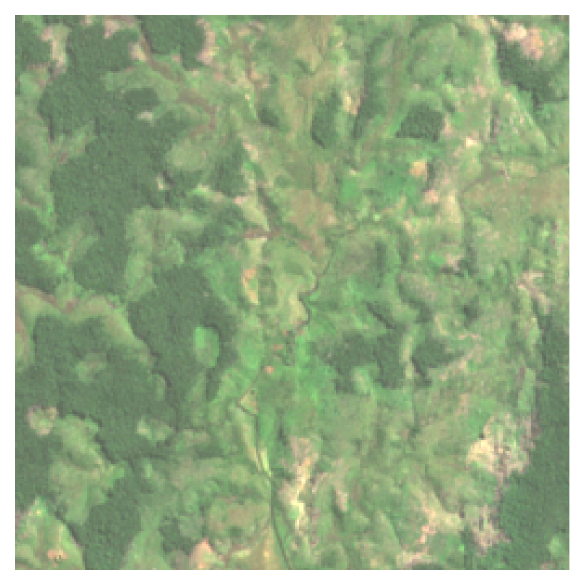}
        \end{subfigure}
        \begin{subfigure}[b]{0.24\textwidth}
            \centering
            \includegraphics[width=\textwidth]{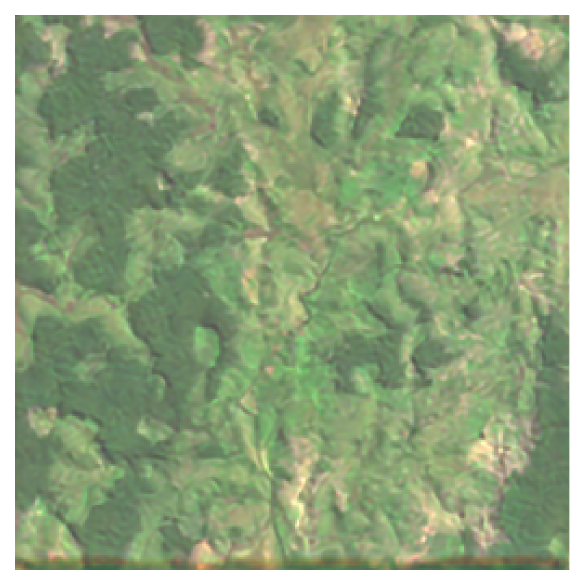}
        \end{subfigure}
        \par\vspace{0.5em}
        \begin{subfigure}[b]{0.24\textwidth}
            \centering
            \includegraphics[width=\textwidth]{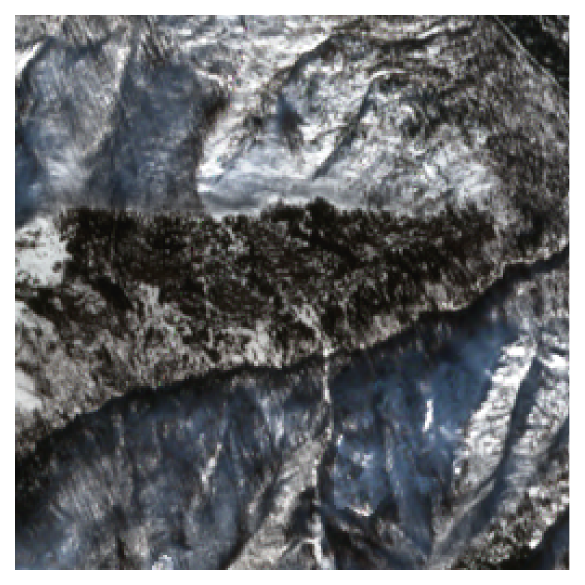}
        \end{subfigure}
        \begin{subfigure}[b]{0.24\textwidth}
            \centering
            \includegraphics[width=\textwidth]{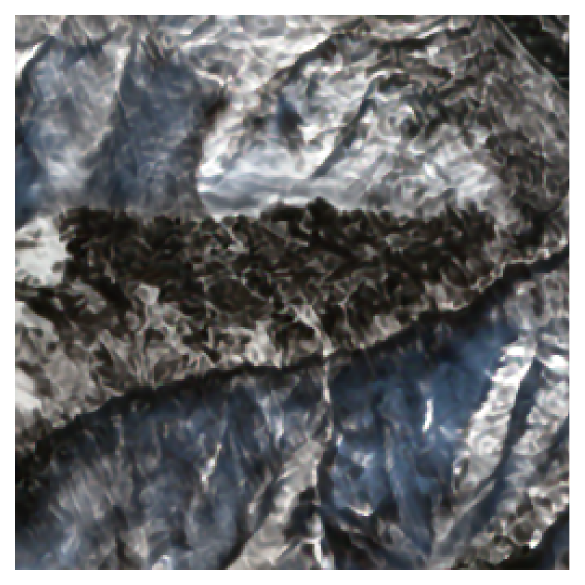}
        \end{subfigure}
        \begin{subfigure}[b]{0.24\textwidth}
            \centering
            \includegraphics[width=\textwidth]{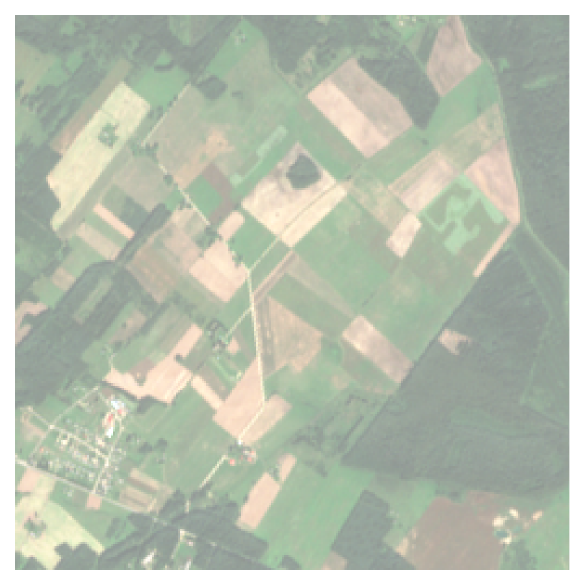}
        \end{subfigure}
        \begin{subfigure}[b]{0.24\textwidth}
            \centering
            \includegraphics[width=\textwidth]{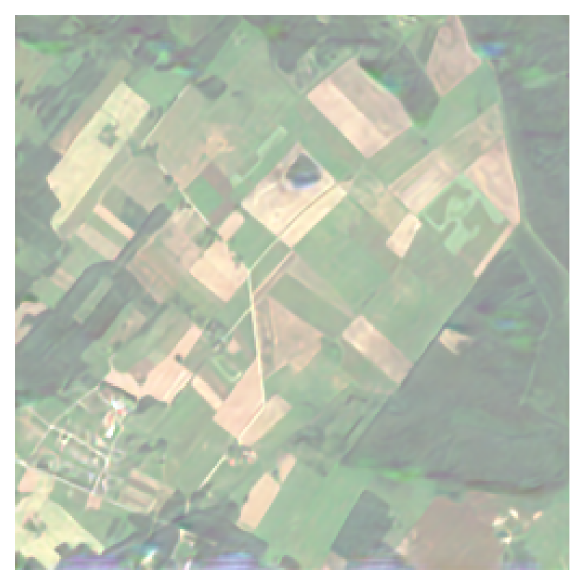}
        \end{subfigure}
        \par\vspace{0.5em}        
                \begin{subfigure}[b]{0.24\textwidth}
            \centering
            \includegraphics[width=\textwidth]{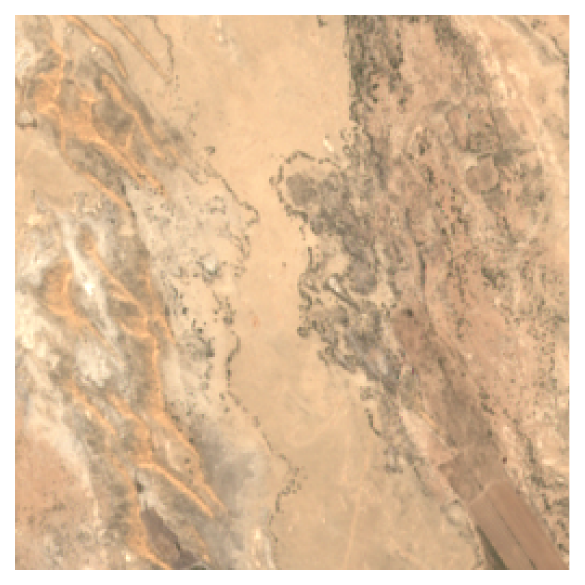}
        \end{subfigure}
        \begin{subfigure}[b]{0.24\textwidth}
            \centering
            \includegraphics[width=\textwidth]{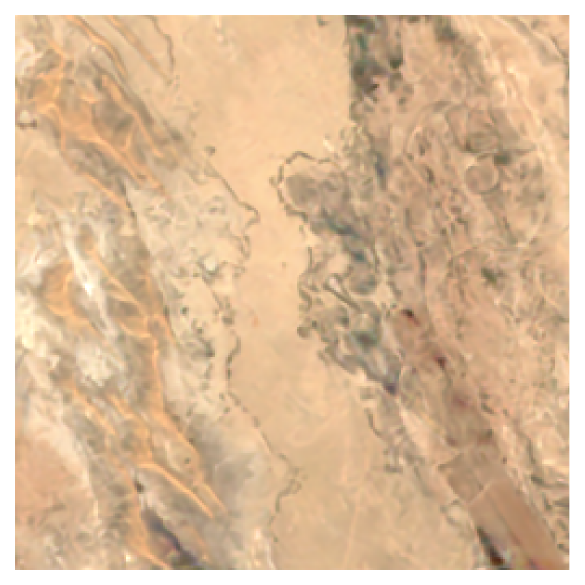}
        \end{subfigure}
        \begin{subfigure}[b]{0.24\textwidth}
            \centering
            \includegraphics[width=\textwidth]{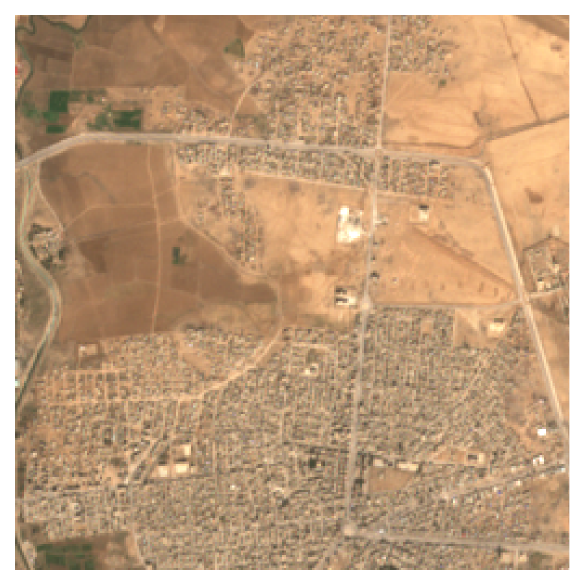}
        \end{subfigure}
        \begin{subfigure}[b]{0.24\textwidth}
            \centering
            \includegraphics[width=\textwidth]{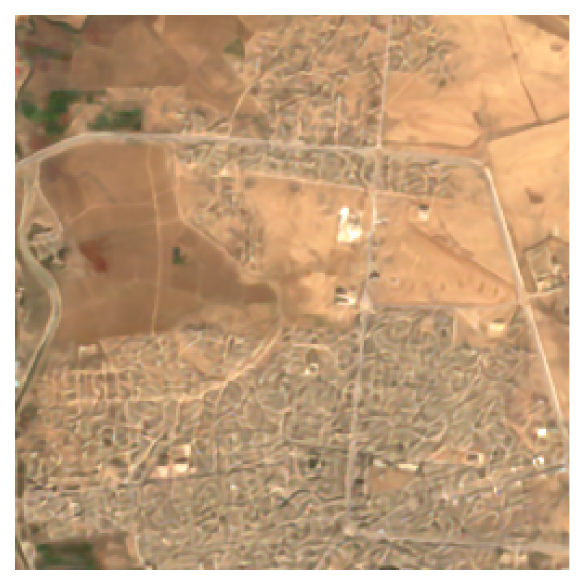}
        \end{subfigure}
        \par\vspace{0.5em} 
                \begin{subfigure}[b]{0.24\textwidth}
            \centering
            \includegraphics[width=\textwidth]{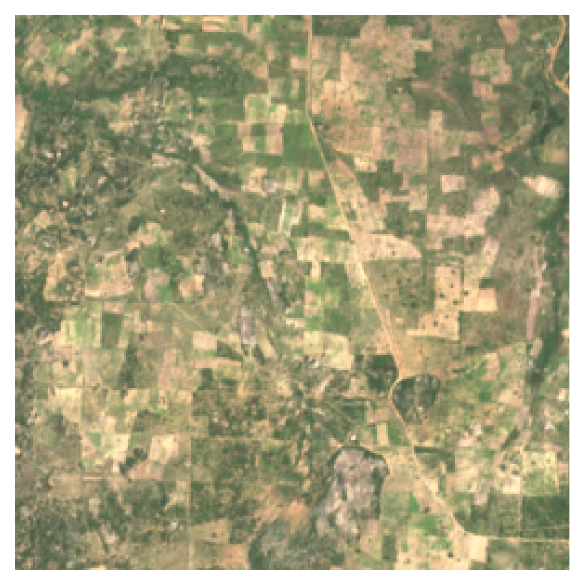}
        \end{subfigure}
        \begin{subfigure}[b]{0.24\textwidth}
            \centering
            \includegraphics[width=\textwidth]{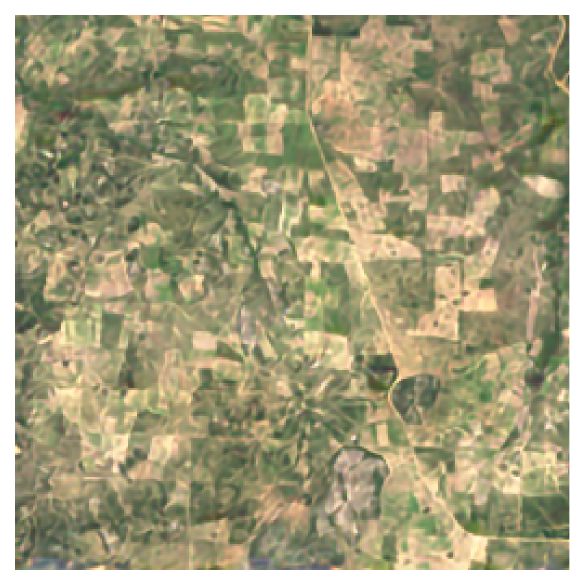}
        \end{subfigure}
        \begin{subfigure}[b]{0.24\textwidth}
            \centering
            \includegraphics[width=\textwidth]{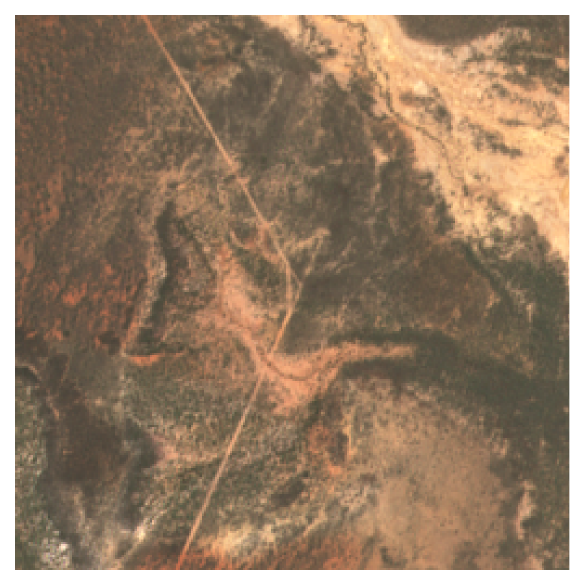}
        \end{subfigure}
        \begin{subfigure}[b]{0.24\textwidth}
            \centering
            \includegraphics[width=\textwidth]{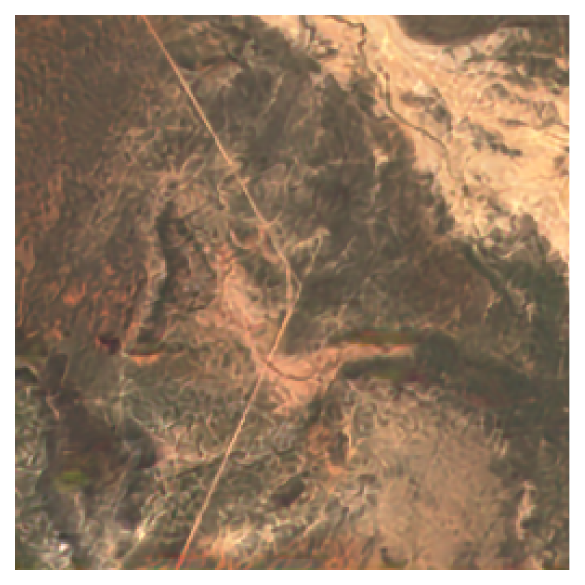}
        \end{subfigure}
        \par\vspace{0.5em} 
 \begin{subfigure}[b]{0.24\textwidth}
            \centering
            \includegraphics[width=\textwidth]{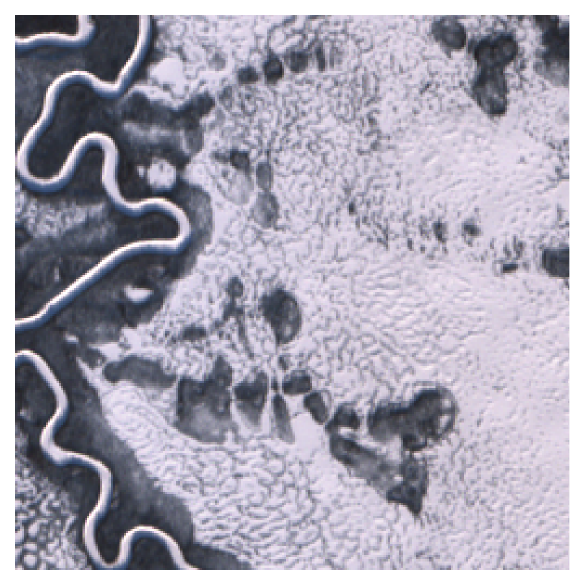}
        \end{subfigure}
        \begin{subfigure}[b]{0.24\textwidth}
            \centering
            \includegraphics[width=\textwidth]{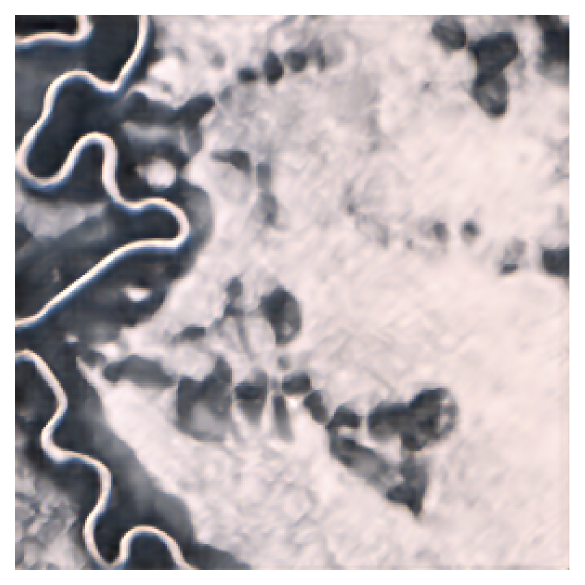}
        \end{subfigure}
        \begin{subfigure}[b]{0.24\textwidth}
            \centering
            \includegraphics[width=\textwidth]{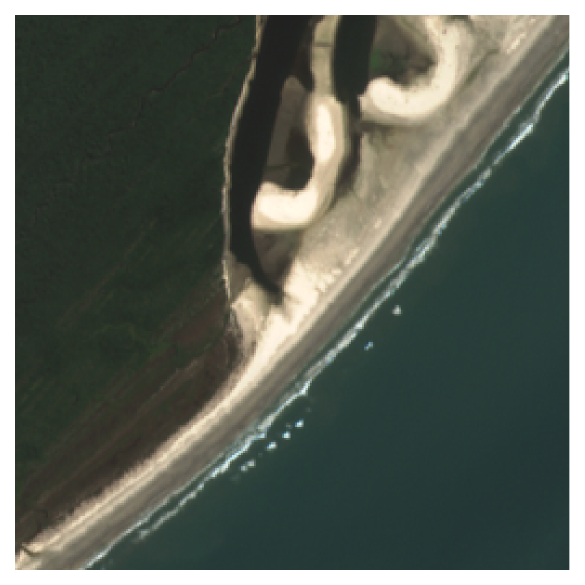}
        \end{subfigure}
        \begin{subfigure}[b]{0.24\textwidth}
            \centering
            \includegraphics[width=\textwidth]{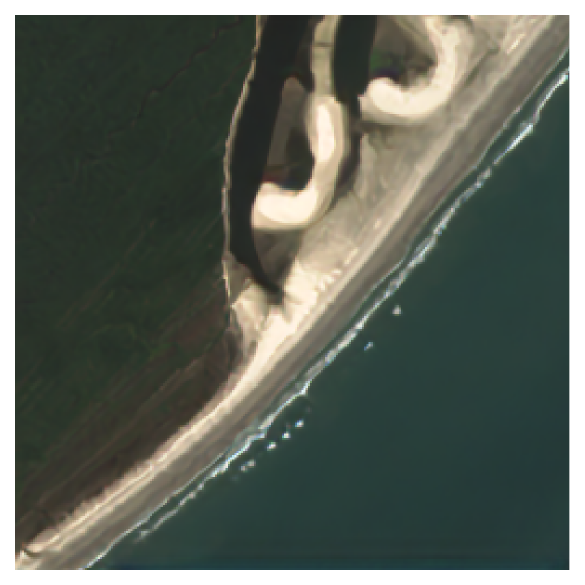}
        \end{subfigure}
        \par\vspace{0.5em}     
    \caption{Original (left) vs Reconstruction (right) for the Sentinel-2 RGB subset. Locations 1-10 are shown in order left-to-right, top-to-bottom. }
    \label{fig:s2l1c global reconstructions}
\end{figure}

In Figure~\ref{fig:S2L1C diverse sample locations}, we provide an overview of the globally sampled locations from which we reconstruct satellite images in Figure~\ref{fig:s2l1c global reconstructions}. Note that all samples in Figure~\ref{fig:s2l1c global reconstructions} are vastly outside of the training distribution of Phaedra.

\clearpage
\subsection{ERA5 Samples}
\label{sec: era5 figures samples}

\begin{figure}[h]
    \centering
        \includegraphics[trim={0pt 1510pt 0pt 0pt},clip,height=7in]{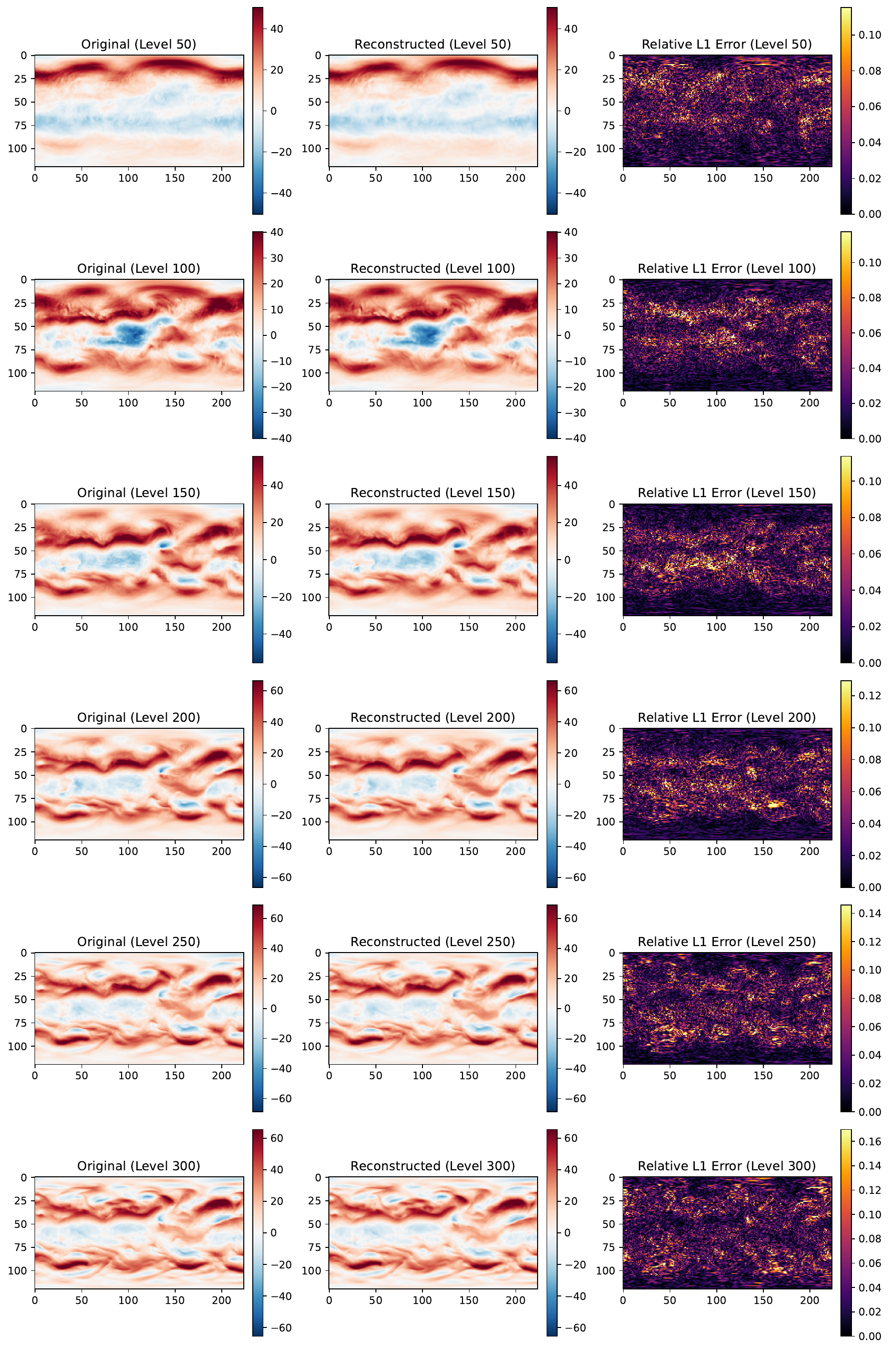}
        \caption{Zonal winds across levels  50-300 (U-component)}
        \label{fig:era5_v10m_horizonal}
\end{figure}
\begin{figure}[h]
    \centering
        \includegraphics[trim={0pt 0pt 0pt 1510pt},clip,height=7in]{images/EO/u_component_of_wind_rot_left.pdf}
        \caption{Zonal winds across levels  500-1000 (U-component)}
        \label{fig:era5_v10m_horizontal 2}
\end{figure}

\begin{figure}[ht]
    \centering
        \includegraphics[trim={0pt 1510pt 0pt 0pt},clip,height=7in]{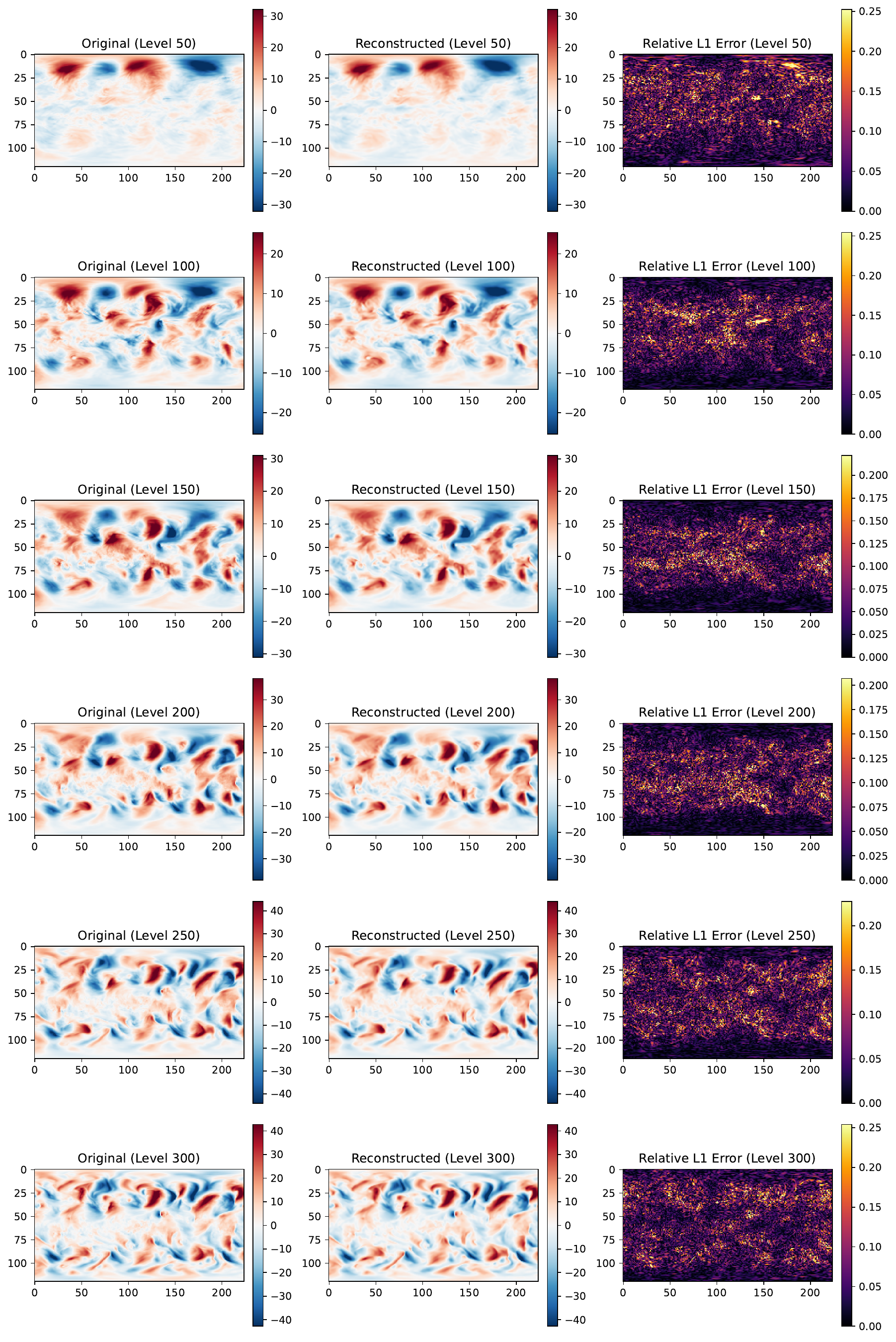}
        \caption{Meridional winds across levels  50-300 (V-component)}
        \label{fig:era5_v10m_vert}
\end{figure}
\begin{figure}[ht]
    \centering
        \includegraphics[trim={0pt 0pt 0pt 1510pt},clip,height=7in]{images/EO/v_component_of_wind_rot_left.pdf}
        \caption{Meridional winds across levels  500-1000 (V-component)}
        \label{fig:era5_v10m_vert 2}
\end{figure}

\clearpage
\subsection{ILSVRC 2012 (ImageNet) Samples}
\label{sec: imagenet ilsvrc figures samples}
\begin{figure}[h]
    \centering
    \begin{subfigure}{0.65\linewidth}
    \includegraphics[width=\linewidth]{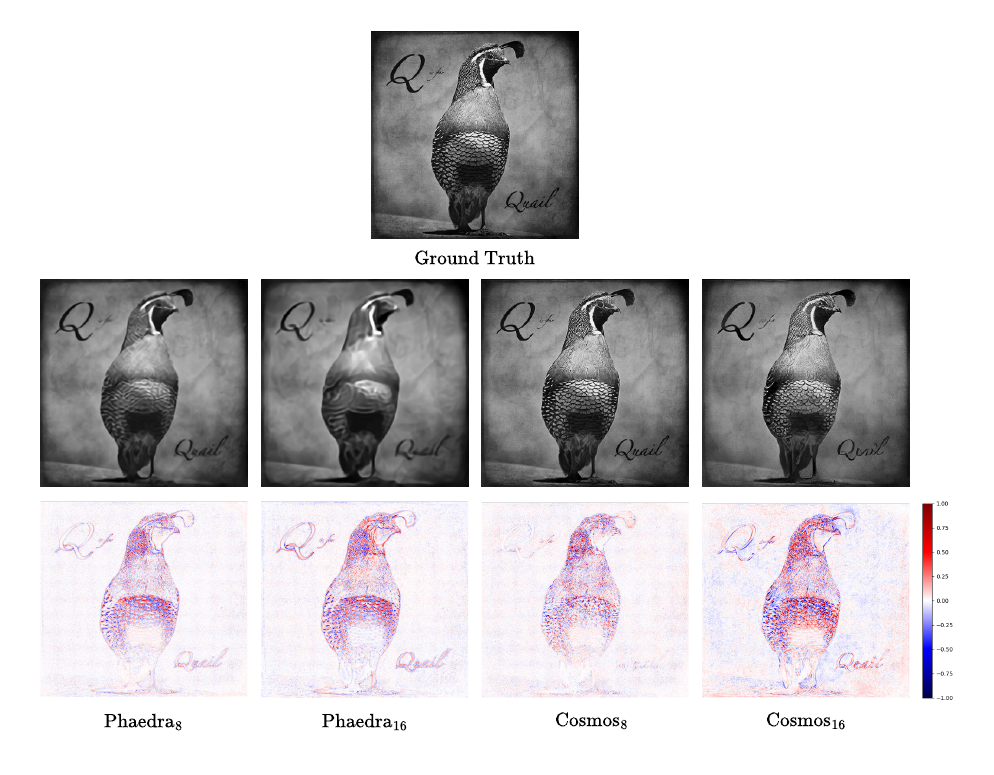}
    \caption{Sample 1 of the ImageNet dataset. }
    \label{fig: imagenet quail figures}
    \end{subfigure}
    \begin{subfigure}{0.65\linewidth}
    \includegraphics[width=\linewidth]{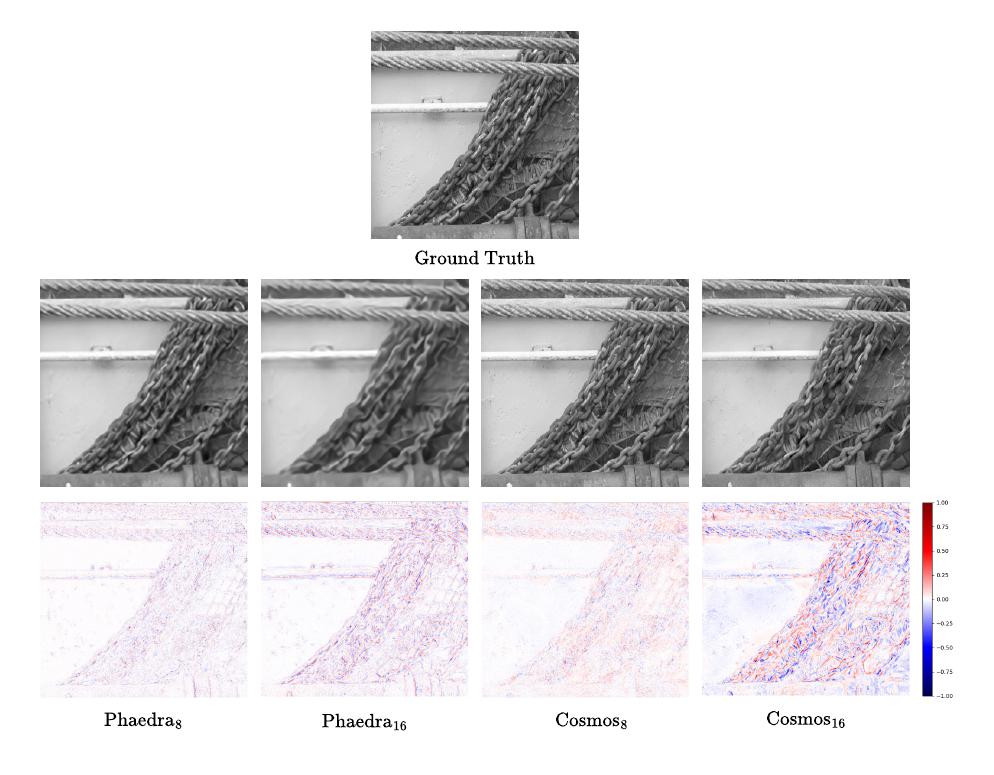}
    \caption{Sample 4 of the ImageNet dataset. }
    \label{fig: imagenet chains figures}
    \end{subfigure}
    \caption{Examples of image reconstructions by Phaedra and Cosmos. Phaedra exhibits smoothing, especially visible under the $16^2$ downsampling. This is primarily a result of the training process, as natural images have a frequency spectra which decays much slower than many PDEs. Alternatively, Cosmos$_{16}$ reconstructs sharper images, but exhibits failure modes such as misplacing sharp transitions (as visible in the chains and feathers) or completely removing details (e.g. the eye of the quail). }
    \label{fig: imagenet reconstructions}
\end{figure}

\clearpage
\section{Extended Downstream Task Results \& Experimental Setup}
\label{sec: downstream task results}

\subsection{Operator Learning}
\textbf{Experimental Setup.} The operator learning models are trained on the entire training dataset utilizing the ``all2all'' training scheme introduced in Poseidon. Specifically, the models map even input timesteps ($t \in [0, 12]$) to all future timesteps ($t \in [2, 14]$) for 28 possible combinations. During testing, only the initial conditions ($t=0$) are provided as input. We evaluate 2-step, 6-6-2 step, and direct 14-step rollout predictions, reporting the best performing rollout strategy for each model. Error metrics are computed exclusively with respect to the final target timestep ($t=14$). Processing speeds for 50 samples show FNO evaluates at $0.0126$ seconds per sample ($79.1$ samples/sec), while Phaedra evaluates at $0.0398$ seconds per sample ($25.12$ samples/sec). 

\textbf{Model Configurations.} The Phaedra (Seq2Seq) transformer utilizes an encoder-decoder architecture with dual decoding heads for amplitude and morphology tokens. The FNO and CNO baselines utilize standard hyperparameter scales. The specific parameter settings are as follows:
\begin{itemize}
    \item \textbf{Phaedra-transformer (38.06M parameters):} \texttt{embed\_dim: 288}, \texttt{encoder\_depth: 19}, \texttt{amp\_decoder\_depth: 5}, \texttt{morph\_decoder\_depth: 5}, \texttt{num\_heads: 8}, \texttt{mlp\_ratio: 4}, \texttt{layerscale\_init: 1.0e-5}; vocabularies 1024 (amplitude) and 8640 (morphology)
    \item \textbf{FSQ-transformer (37.30M parameters):} \texttt{embed\_dim: 288}, \texttt{encoder\_depth: 19}, \texttt{decoder\_depth: 10}, \texttt{num\_heads: 8}, \texttt{mlp\_ratio: 4}, \texttt{layerscale\_init: 1.0e-5}; single token stream, vocabulary 8640
    \item \textbf{VQ-VAE-2-transformer (44.30M parameters):} identical backbone to the Phaedra-transformer (\texttt{embed\_dim: 288}, \texttt{encoder\_depth: 19}, \texttt{amp\_decoder\_depth: 5}, \texttt{morph\_decoder\_depth: 5}, \texttt{num\_heads: 8}, \texttt{mlp\_ratio: 4}, \texttt{layerscale\_init: 1.0e-5}); the top ($16\times16$, vocabulary 4096) and bottom ($32\times32$, vocabulary 16384) VQ-VAE-2 codes take the place of the amplitude and morphology streams. The 32.48M-parameter body is identical to the Phaedra-transformer; the additional parameters are entirely the larger token embeddings and output heads.
    \item \textbf{Continuous-latent transformer (38.24M parameters):} \texttt{latent\_channels: 8}, \texttt{embed\_dim: 320}, \texttt{encoder\_depth: 19}, \texttt{decoder\_depth: 9}, \texttt{num\_heads: 8}, \texttt{mlp\_ratio: 4}; the same encoder-decoder skeleton with the token embeddings and cross-entropy heads replaced by linear input/output projections of the continuous autoencoder latents, a single (non-bifurcated) decoder, and an $L_1$ loss in latent space
    \item \textbf{ViT (37.92M parameters):} \texttt{embed\_dim: 512}, \texttt{depth: 12}, \texttt{num\_heads: 8}, \texttt{mlp\_ratio: 4}, \texttt{patch\_size: 4}; operates directly on the $128\times128$ physical fields
    \item \textbf{FNO (37.83M parameters):} \texttt{width: 96}, \texttt{depth: 8}, \texttt{modes\_x: 16}, \texttt{modes\_y: 16}, \texttt{padding: 8}
    \item \textbf{CNO (38.78M parameters):} \texttt{width: 76}, \texttt{levels: 4}, \texttt{level blocks: 2}, \texttt{bottleneck blocks: 2}, \texttt{channel multiplier: 2}
\end{itemize}

The FNO and CNO are trained on 4 RTX 4090 GPUs in parallel, while the transformer-based models are trained on 8 RTX Pro 6000 GPUs to support larger batch sizes, due to the increased computational load of large transformers. Each model is trained for 50 epochs. This takes approximately 8 hours for the FNO or CNO and initially requiring approximately 80 hours for the transformers; however, this was cut down to approximately 20 hours on 8 GPUs after several optimizations. 

\begin{table}[h]
    \centering
    \caption{Full per-variable operator learning relative $L_1$ errors at final time step ($t=0.7$). Three rollout time-stepping strategies with short (2-step), medium (6-6-2 step), and long (direct) steps are tested, and the best results for each model are reported.}
    
 \resizebox{\linewidth}{!}{
    \begin{tabular}{llccccccc}
        \toprule
        Model & Dataset  & $\rho$ & $u$ & $v$ & $p$ & Average & Strategy & CI$_{95}$ \\
        \midrule
        FNO & KH & 6.24 & 11.18 & 32.23 & 0.54 & 12.55 & 6-6-2 & $\pm0.36$ \\
        FNO & RC & 26.31 & 53.62 & 53.42 & 9.16 & 35.63 & 6-6-2 & $\pm0.69$ \\
        FNO & RKH & 10.08 & 26.36 & 25.67 & 4.54 & 16.66 & 6-6-2 & $\pm0.91$ \\
        \midrule
        CNO & KH & 5.06 & 9.10 & 25.93 & 0.51 & 10.15 & 2-step & $\pm0.32$ \\
        CNO & RC & 22.65 & 44.44 & 44.66 & 7.94 & 29.92 & 2-step & $\pm0.74$ \\
        CNO & RKH & 6.79 & 18.83 & 18.50 & 3.17 & 11.82 & 6-6-2 & $\pm0.82$ \\
        \midrule
        ViT & KH & 4.64 & 8.25 & 23.33 & 0.51 & \textbf{9.18} & 6-6-2 & $\pm0.27$ \\
        ViT & RC & 28.76 & 60.17 & 59.51 & 10.28 & 39.68 & direct & $\pm0.58$ \\
        ViT & RKH & 9.09 & 20.86 & 19.66 & 3.18 & 13.20 & 6-6-2 & $\pm0.79$ \\
        \midrule
        Continuous Transformer & KH & 4.64 & 8.34 & 23.53 & 0.61 & 9.28 & 6-6-2 & $\pm0.30$ \\
        Continuous Transformer & RC & 77.95 & 197.31 & 156.93 & 44.00 & 119.05 & direct & $\pm2.96$ \\
        Continuous Transformer & RKH & 7.09 & 18.09 & 17.87 & 3.18 & 11.56 & 6-6-2 & $\pm0.70$ \\
        \midrule
        VQ-VAE-2 & KH & 7.95 & 13.90 & 38.04 & 0.68 & 15.14 & 6-6-2 & $\pm0.48$ \\
        VQ-VAE-2 & RC & 29.54 & 63.17 & 64.39 & 12.48 & 42.40 & direct & $\pm0.63$ \\
        VQ-VAE-2 & RKH & 14.73 & 38.09 & 37.29 & 7.97 & 24.52 & 6-6-2 & $\pm1.15$ \\
        \midrule
        FSQ & KH & 5.30 & 10.91 & 28.27 & 0.62 & 11.28 & 6-6-2 & $\pm0.34$ \\
        FSQ & RC & 28.12 & 57.08 & 56.77 & 10.21 & 38.05 & 6-6-2 & $\pm1.05$ \\
        FSQ & RKH & 10.37 & 27.73 & 28.78 & 5.08 & 17.99 & 6-6-2 & $\pm1.37$ \\
        \midrule
        Phaedra & KH & 4.78 & 8.43 & 24.29 & 0.51 & 9.50 & direct & $\pm0.34$ \\
        Phaedra & RC & 20.66 & 40.73 & 40.42 & 7.02 & \textbf{27.21} & 6-6-2 & $\pm0.71$ \\
        Phaedra & RKH & 6.52 & 16.01 & 15.64 & 2.74 & \textbf{10.23} & 6-6-2 & $\pm0.80$ \\
        \bottomrule
    \end{tabular}}
    \label{tab:op_learning_full}
\end{table}

\begin{table}[h]

    \centering
    \caption{Operator learning relative $L_1$ errors under the direct-step strategy (no rollout) across timesteps.}
    
 \resizebox{\linewidth}{!}{
 \begin{tabular}{llccccccc}
    \toprule
    Model & Dataset & $t=0.1$ & $t=0.2$ & $t=0.3$ & $t=0.4$ & $t=0.5$ & $t=0.6$ & $t=0.7$ \\
    \midrule
    FNO & KH & 6.63 & 7.43 & 8.23 & 9.20 & 10.34 & 11.60 & 13.10 \\
    CNO & KH & \textbf{5.15} & 5.97 & 6.73 & 7.70 & 8.80 & 10.05 & 11.72 \\
    ViT & KH & 5.20 & \textbf{5.53} & \textbf{6.06} & \textbf{6.81} & 7.67 & 8.53 & 9.71 \\
    Continuous Transformer & KH & 6.15 & 6.08 & 6.32 & 6.86 & \textbf{7.59} & \textbf{8.40} & \textbf{9.42} \\
    VQ-VAE-2 & KH & 12.19 & 11.11 & 10.80 & 11.54 & 12.48 & 13.76 & 15.28 \\
    FSQ & KH & 6.78 & 7.64 & 8.24 & 8.98 & 10.09 & 11.09 & 11.99 \\
    Phaedra & KH & 5.64 & 5.96 & 6.30 & 6.85 & 7.62 & 8.42 & 9.50 \\
    \midrule
    FNO & RC & 17.29 & 22.05 & 26.90 & 31.47 & 35.18 & 38.74 & 42.80 \\
    CNO & RC & 7.61 & 13.19 & 19.72 & 26.17 & 31.63 & 36.38 & 41.23 \\
    ViT & RC & 16.53 & 21.38 & 26.48 & 31.21 & 34.14 & 36.99 & 39.68 \\
    Continuous Transformer & RC & 113.76 & 118.12 & 119.40 & 125.10 & 129.61 & 122.10 & 119.05 \\
    VQ-VAE-2 & RC & 19.73 & 23.84 & 29.21 & 33.91 & 36.95 & 39.37 & 42.40 \\
    FSQ & RC & 10.96 & 16.21 & 22.74 & 28.77 & 33.75 & 38.31 & 42.21 \\
    Phaedra & RC & \textbf{7.14} & \textbf{9.83} & \textbf{13.65} & \textbf{18.01} & \textbf{21.93} & \textbf{25.35} & \textbf{28.87} \\
    \midrule
    FNO & RKH & 5.43 & 6.76 & 9.00 & 11.53 & 13.09 & 15.42 & 18.59 \\
    CNO & RKH & \textbf{2.50} & 3.52 & 5.31 & 7.56 & 9.47 & 12.36 & 15.84 \\
    ViT & RKH & 4.60 & 5.74 & 7.89 & 9.97 & 11.26 & 13.38 & 15.89 \\
    Continuous Transformer & RKH & 3.49 & 3.92 & 5.41 & 7.05 & 7.98 & \textbf{9.55} & \textbf{11.85} \\
    VQ-VAE-2 & RKH & 9.12 & 10.59 & 13.86 & 17.72 & 19.97 & 22.80 & 26.00 \\
    FSQ & RKH & 4.71 & 5.63 & 7.85 & 10.29 & 12.30 & 15.13 & 18.55 \\
    Phaedra & RKH & 2.83 & \textbf{3.32} & \textbf{4.69} & \textbf{6.36} & \textbf{7.66} & 9.67 & 12.27 \\
    \bottomrule
    \end{tabular}}
    \label{tab: operator learning direct across timesteps}
\end{table}

\paragraph{Scaling Experiment.}
We present the relative $L_1$ error per timestep, averaged over 4 variables, at 5M / 38M / 200M parameters for the RKH OOD dataset. The learning rate is held fixed for this experiment. Absolute errors are therefore higher than elsewhere, which follow a cosine learning rate schedule. The results show monotone decay across all timesteps without saturation. 

\begin{table}[htbp]
\centering
 \caption{Performance across timesteps and model scales for the CEU RKH dataset.}
 \resizebox{\linewidth}{!}{
\begin{tabular}{l ccccccc}
\toprule
\textbf{Model Size} & \textbf{$t=0.1$} & \textbf{$t=0.2$} & \textbf{$t=0.3$} & \textbf{$t=0.4$} & \textbf{$t=0.5$} & \textbf{$t=0.6$} & \textbf{$t=0.7$} \\
\midrule
5M   & $16.94 \pm 3.61$ & $19.59 \pm 4.86$ & $24.19 \pm 5.99$ & $28.68 \pm 6.74$ & $30.88 \pm 7.10$ & $33.51 \pm 9.22$ & $36.16 \pm 9.38$ \\
38M  & $5.38 \pm 1.58$  & $6.42 \pm 1.89$  & $8.77 \pm 2.97$  & $11.53 \pm 4.15$ & $13.39 \pm 4.81$ & $15.98 \pm 6.92$ & $18.99 \pm 7.93$ \\
200M & $3.72 \pm 1.23$  & $4.34 \pm 1.45$  & $6.03 \pm 2.38$  & $8.10 \pm 3.38$  & $9.88 \pm 4.06$  & $12.10 \pm 6.02$ & $14.89 \pm 7.45$ \\
\bottomrule
\end{tabular}}
\label{tab:model scaling across timesteps}
\end{table}

\subsection{Masked Autoencoding (MAE)}
\textbf{Experimental Setup.} The MAE models are pretrained on all three PDEs and subsequently fine-tuned for several epochs on each specific PDE dataset. During training and evaluation, tokens from $75\%$ of the spatial locations are masked. This spatial masking mask is consistent across all physical variables (channels). Unlike the operator learning task, reconstruction error for the MAE task is averaged across all timesteps.

\textbf{Model Configurations.} To ensure a direct comparison with FSQ, the Phaedra MAE model replaces the dual-head decoder with a single decoder handling both amplitude and morphology. The configuration for both Phaedra and FSQ is:
\texttt{embed\_dim: 512}, \texttt{encoder\_depth: 8}, \texttt{decoder\_depth: 4}, \texttt{num\_heads: 8}, \texttt{mlp\_ratio: 4}, \texttt{layerscale\_init: 1.0e-5}.

\begin{table}[h]
    \centering
    \caption{Full masked autoencoding results. Average across all 4 physical channels. Inputs are $75\%$ masked. Wasserstein-1 (W1) metrics are scaled by $10^3$.}
    \begin{tabular}{llcc}
        \toprule
        Model & Dataset & Relative $L_1$ Error & W1 \\
        \midrule
        FSQ & KH & 5.84 & 1.71 \\
        FSQ & RC & 29.16 & 21.71 \\
        FSQ & RKH & 9.45 & 8.19 \\
        \midrule
        Phaedra & KH & 4.75 & 1.21 \\
        Phaedra & RC & 22.17 & 12.50 \\
        Phaedra & RKH & 7.27 & 5.52 \\
        \bottomrule
    \end{tabular}
    \label{tab:mae_full}
\end{table}

\clearpage
\subsection{Operator Learning Visualizations}
\begin{figure}[h]
    \centering
    \begin{subfigure}[b]{0.7\textwidth}
        \centering
        \includegraphics[width=\linewidth]{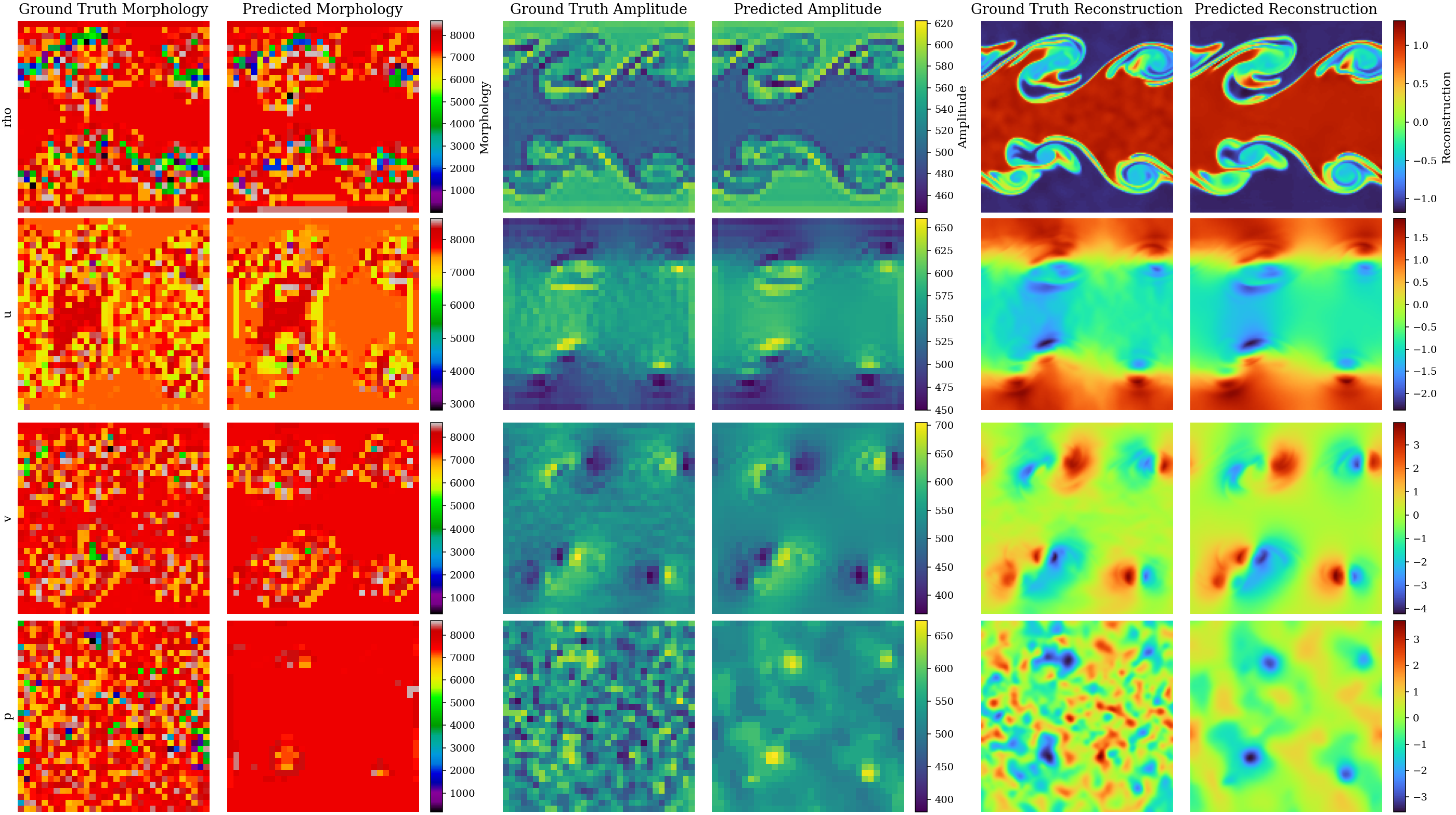}
        \caption{Kelvin-Helmholtz (KH)}
        \label{fig:tokens_kh}
    \end{subfigure}
    
    \begin{subfigure}[b]{0.7\textwidth}
        \centering
        \includegraphics[width=\linewidth]{images/DownstreamOperatorLearning/Phaedra/RC_tokens.png}
        \caption{Curved Riemann (RC)}
        \label{fig:tokens_rc}
    \end{subfigure}

    \begin{subfigure}[b]{0.7\textwidth}
        \centering
        \includegraphics[width=\linewidth]{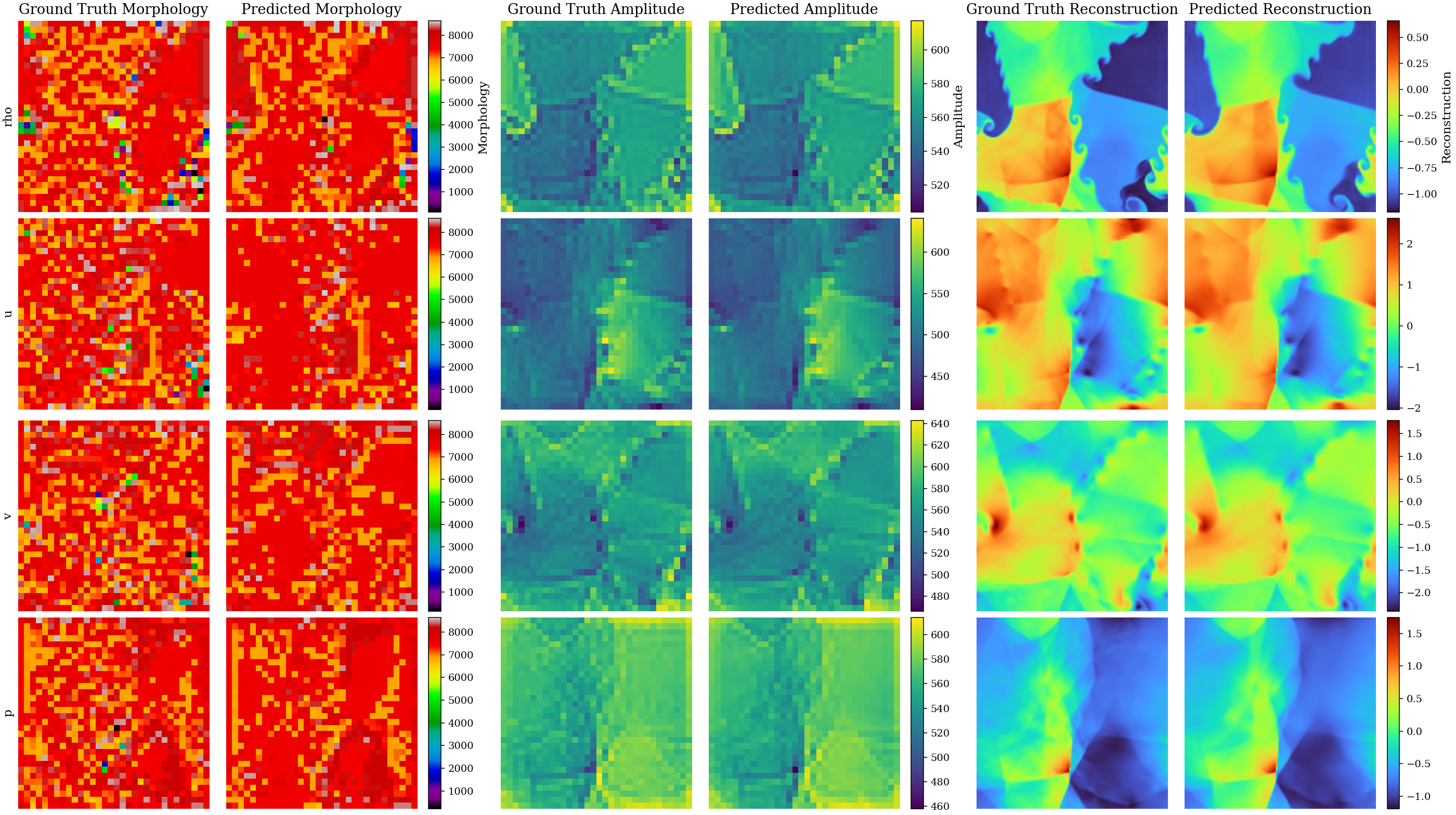}
        \caption{Riemann Kelvin-Helmholtz (RKH)}
        \label{fig:tokens_rkh}
    \end{subfigure}
    \caption{Visualizations of the discrete Phaedra tokens predicted by the transformer model across the three evaluation tasks. Each figure corresponds to the first test sample from the test set. All predictions are produced by a rollout from the initial conditions.}
    \label{fig:op_learning_tokens}
\end{figure}

\begin{figure}[htbp]
    \centering 
    \includegraphics[width=0.8\linewidth, trim={0 0 0 30},clip]{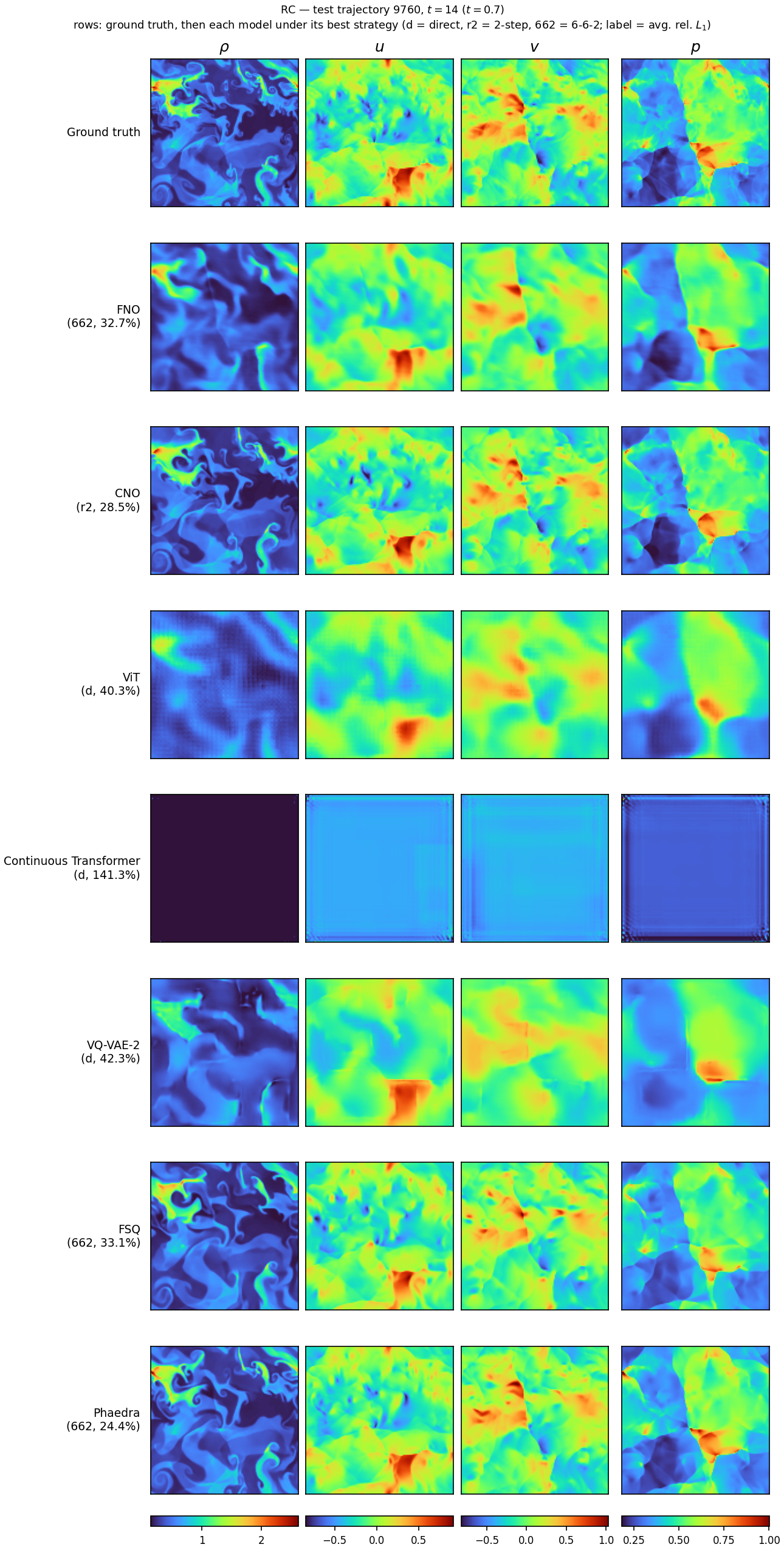}
    \caption{Comparison of the predicted physical fields against the ground truth simulations for the CEU RC operator learning task for the first test sample. Each model is presented under its best strategy (d=direct, r2=2-step, 662=6-6-2 lead times; label = avg. rel $L_1$).}
    \label{fig:operator learning downstream task ceu rc}
\end{figure}

\begin{figure}[htbp]
    \centering 
    \includegraphics[width=0.8\linewidth, trim={0 0 0 30},clip]{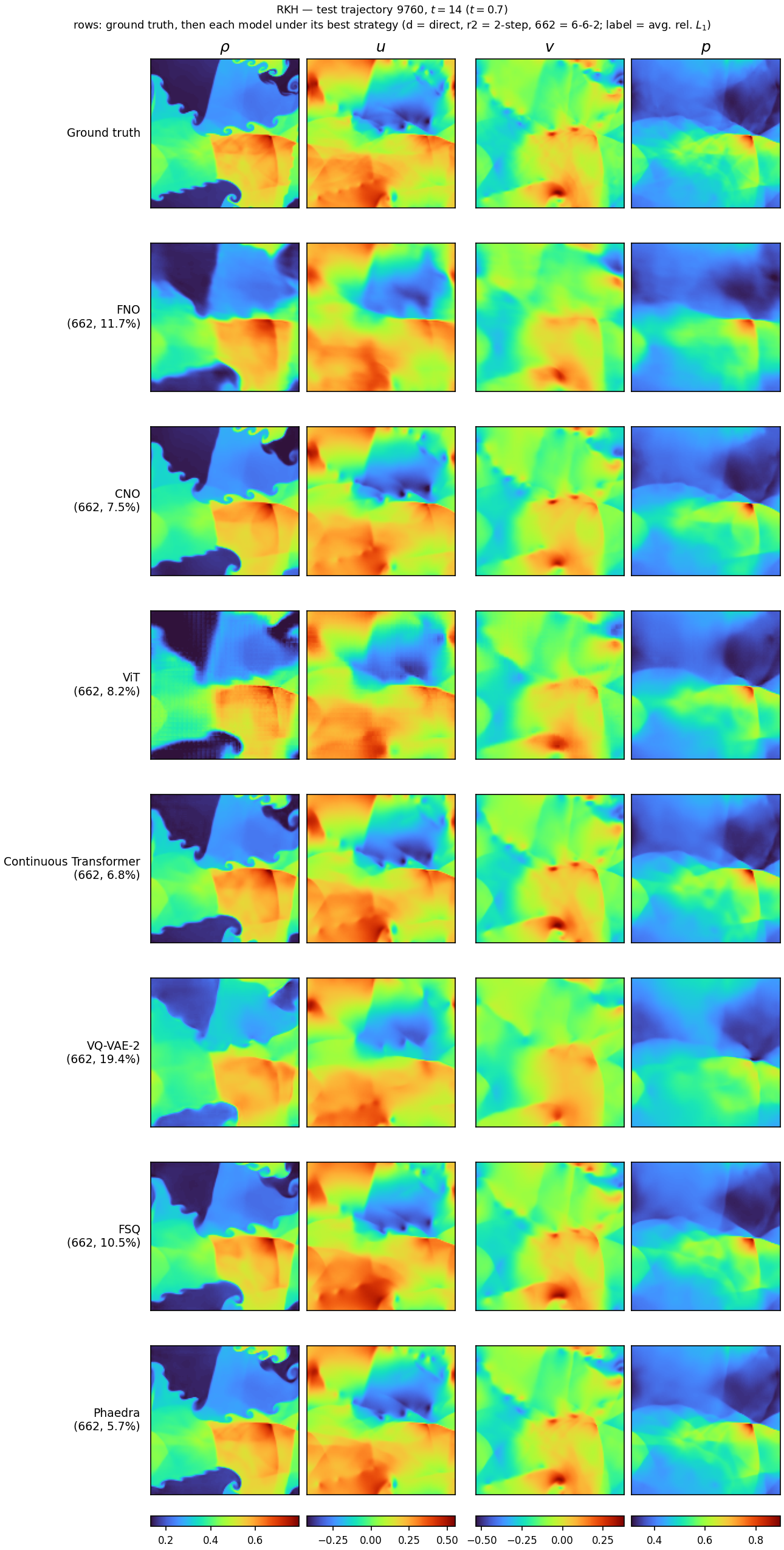}
    \caption{Comparison of the predicted physical fields against the ground truth simulations for the CEU RKH operator learning task for the first test sample. Each model is presented under its best strategy (d=direct, r2=2-step, 662=6-6-2 lead times; label = avg. rel $L_1$).}
    \label{fig:operator learning downstream task ceu rkh}
\end{figure}

\clearpage
\subsection{Masked Autoencoding (MAE) Visualizations}
\begin{figure}[htbp]
    \centering
    \begin{subfigure}[b]{0.8\textwidth}
        \centering
        \includegraphics[width=\linewidth]{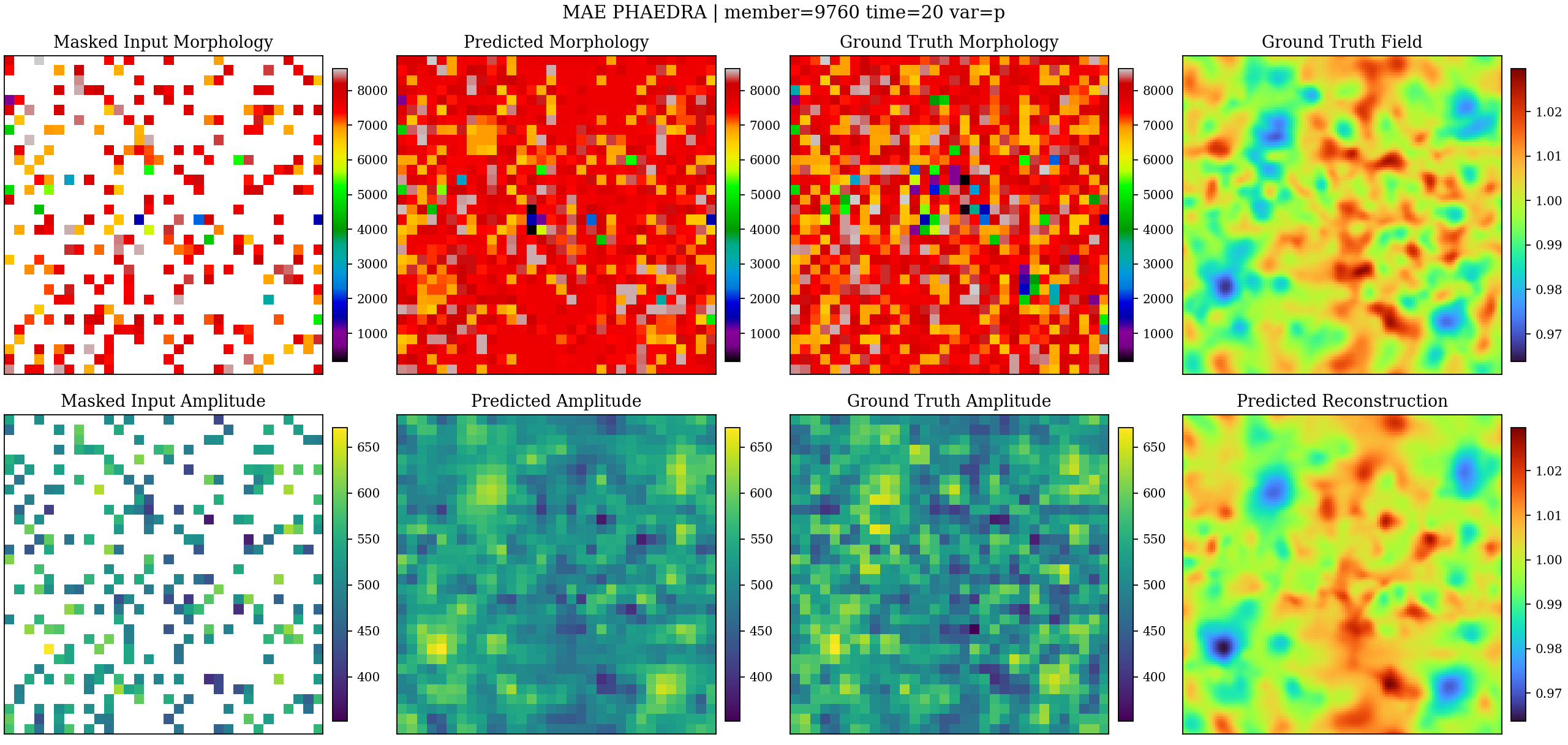}
        \caption{Kelvin-Helmholtz, $p$ (KH)}
        \label{fig:mae_kh}
    \end{subfigure}

    \begin{subfigure}[b]{0.8\textwidth}
        \centering
        \includegraphics[width=\linewidth]{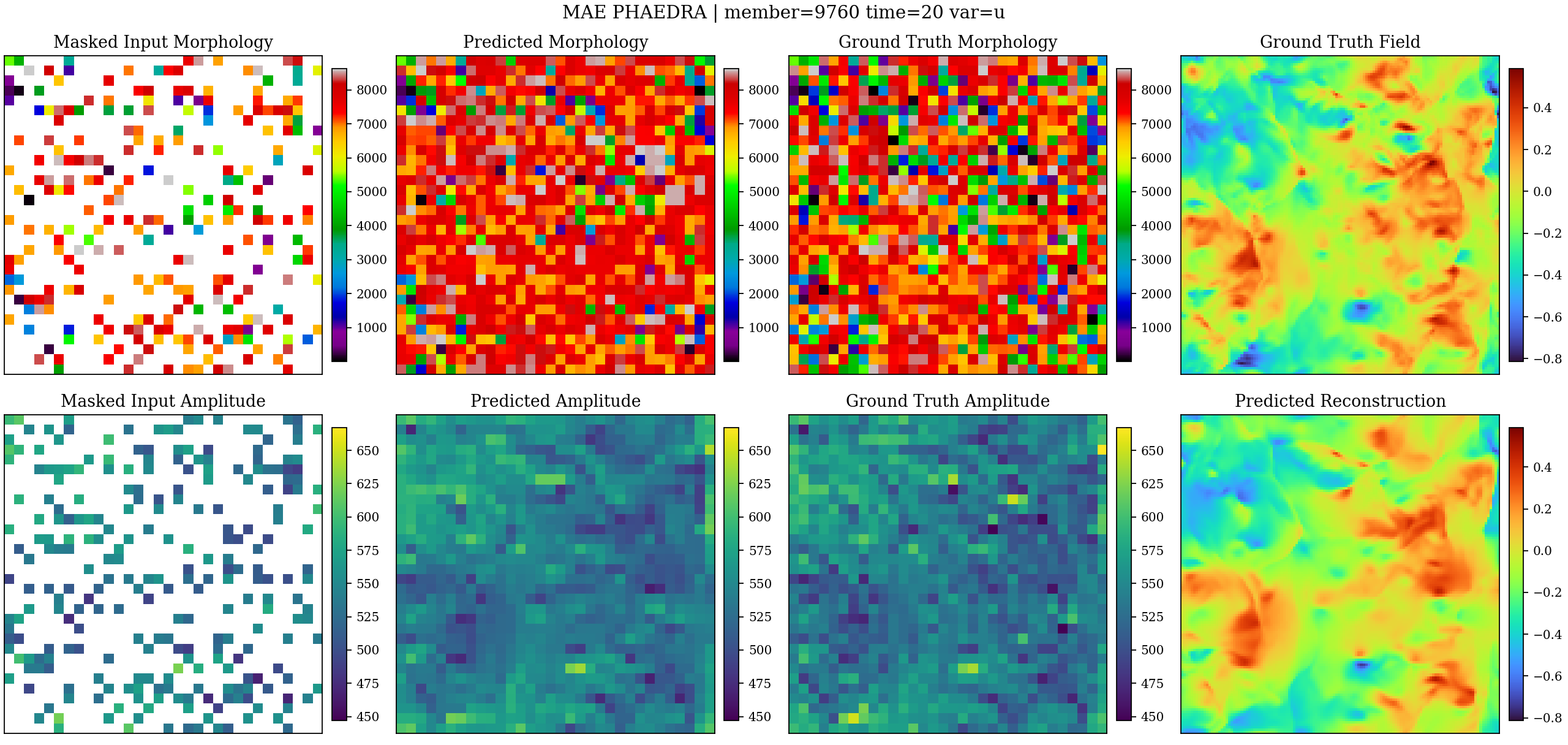}
        \caption{Curved Riemann, $u$ (RC)}
        \label{fig:mae_rc}
    \end{subfigure}

    \begin{subfigure}[b]{0.8\textwidth}
        \centering
        \includegraphics[width=\linewidth]{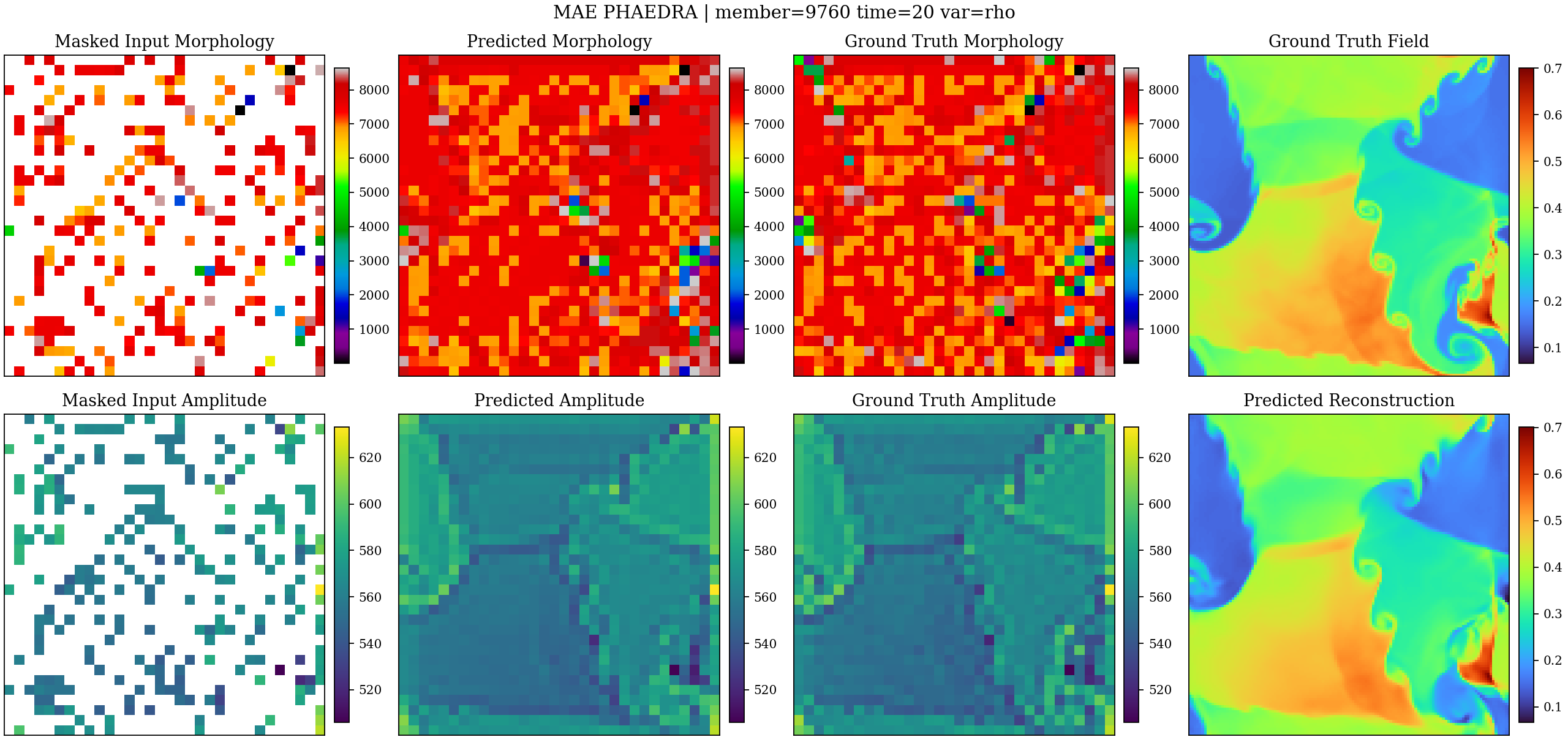}
        \caption{Riemann Kelvin-Helmholtz, $\rho$ (RKH)}
        \label{fig:mae_rkh}
    \end{subfigure}
    \caption{Masked Autoencoding (MAE) reconstructions for selected fields under $75\%$ spatial masking. Each panel contrasts the masked input provided to the model with the predicted reconstruction and the unmasked ground truth.}
    \label{fig:mae_density_fields}
\end{figure}

\newpage

\end{document}